\RequirePackage{fix-cm}
\documentclass[smallextended]{svjour3}       
\smartqed  
\usepackage{graphicx}
\usepackage{caption}
\usepackage{subcaption}
\usepackage{epstopdf}
\usepackage{amsmath,amsfonts}      
\usepackage{natbib}
\usepackage{booktabs}
\usepackage{xifthen,bm,tikz,tikz-cd}
\usetikzlibrary{positioning}
\usepackage{pgfplots}
\usepackage{float}
\journalname{Machine Learning}
\begin{document}
	
	\title{Adversarial Concept Drift Detection under Poisoning Attacks for Robust Data Stream Mining
	}
	
	\titlerunning{Adversarial Concept Drift Detection under Poisoning Attacks}        
	
	\author{\L{}ukasz Korycki         \and
		Bartosz Krawczyk$^{\dagger}$ 
	}
	
	
	\institute{\L{}. Korycki and B. Krawczyk  \at
		Department of Computer Science, Virginia Commonwealth University, Richmond VA, USA \\
		\email{\{koryckil,bkrawczyk\}@vcu.edu}           \\
		$^{\dagger}$Corresponding author
	}
	
	\date{Received: date / Accepted: date}

	\maketitle
	
	\begin{abstract}
		Continuous learning from streaming data is among the most challenging topics in the contemporary machine learning. In this domain, learning algorithms must not only be able to handle massive volumes of rapidly arriving data, but also adapt themselves to potential emerging changes. The phenomenon of the evolving nature of data streams is known as concept drift. While there is a plethora of methods designed for detecting its occurrence, all of them assume that the drift is connected with underlying changes in the source of data. However, one must consider the possibility of a malicious injection of false data that simulates a concept drift. This adversarial setting assumes a poisoning attack that may be conducted in order to damage the underlying classification system by forcing adaptation to false data. Existing drift detectors are not capable of differentiating between real and adversarial concept drift. In this paper, we propose a framework for robust concept drift detection in the presence of adversarial and poisoning attacks. We introduce the taxonomy for two types of adversarial concept drifts, as well as a robust trainable drift detector. It is based on the augmented Restricted Boltzmann Machine with improved gradient computation and energy function. We also introduce Relative Loss of Robustness -- a novel measure for evaluating the performance of concept drift detectors under poisoning attacks. Extensive computational experiments, conducted on both fully and sparsely labeled data streams, prove the high robustness and efficacy of the proposed drift detection framework in adversarial scenarios.
		
	\end{abstract}
	
	\section{Introduction}
	\label{sec:int}
	
	Modern machine learning algorithms should not assume that they will deal with finite, closed collections of data. Contemporary data sources generate new information constantly and at a high speed. This combination of velocity and volume gave birth to the notion of data streams that constantly expand and flood the computing system \citep{Bifet:2019}. Data stream cannot be stored in memory and must be analyzed on the fly, without latency that may reduce the responsiveness and lead to a bottleneck. Such characteristics pose new challenges for machine learning algorithms. They need not only to display a high predictive accuracy, but also be capable of fast incorporation of new information, being computationally lightweight and responsive. The recent COVID-19 pandemic is an example of such a scenario, where models needed to be updated daily, whenever new information become available \citep{Chatterjee:2020a}. Due to the novelty of this disease and scarce access to ground truth, any new confirmed data was extremely valuable and updating the predictive model was of crucial importance. Efficient learning from data streams calls for such algorithms that can incorporate new data without a need for being retrained from scratch. Furthermore, algorithms dedicated to data stream mining problems must take into account the possibility of dealing with dynamic and non-stationary distributions \citep{Ditzler:2015}. Properties of data may change over time, making previously trained model outdated and forcing constant adaptation to changes. The ever-changing nature of data streams is known as concept drift.
	
	Concept drift can originate from the changes in underlying data generators (e.g., class distributions) or simply from having a lack of access to truly stable and well-defined collections of instances \citep{Lu:2019}. The former is the most common case, where the problem under consideration is subject to non-stationary changes and shifts in its nature. Recommendation systems are excellent examples of such naturally occurring drifts, as the tastes of users may change over time. An example of more rapid changes would include a sensor network, where one of the sensors becomes suddenly damaged and the entire system needs to adapt to the new situation \citep{Liu:2017}. The latter case is strongly connected with the problem of limited access to ground truth in data streams \citep{Masud:2011}. As we deal with constantly arriving instances, it may be impossible to provide a label for every one of them. Therefore, this becomes a problem of managing budget (how many instances can we afford to analyze and label) and time (how quickly are we able to label selected instances) \citep{Lughofer:2017}. Here, concept drift can be seen as a byproduct of the exploration-exploitation trade-off, where as we learn more about the underlying distributions, we need to update the previous models accordingly \citep{Korycki:2017}.
	
	While there is a plethora of methods dedicated to explicit or implicit concept drift detection, they all assume that the occurrence of the drift originates purely in the underlying changes in data sources \citep{Sethi:2018}. But what would happen if we considered the potential presence of a malicious party, aiming at attacking our data stream mining system \citep{Biggio:2018}? Adversarial learning in the presence of malicious data and poisoning attacks recently gained a significant attention \citep{Miller:2020}. However, this area of research focuses on dealing with corrupted training/testing sets \citep{Biggio:2012,Xiao:2015} and handling potentially dangerous instances present there \citep{Umer:2019}. Creating artificial adversarial instances \citep{Gao:2020} or using evasion \citep{Mahloujifar:2019} are considered as the most efficient approaches. The adversarial learning set-up has rarely been discussed in the data stream context, where adversarial instances may be injected into the stream at any point. One should consider the dangerous potential of introducing adversarial concept drift into the stream. Such a fake change may either lead to a premature and unnecessary adaptation to false changes, or slow down the adaptation to the real concept drift. Analyzing the presence of adversarial data and poisoning attack in the streaming setting is a challenging, yet important direction towards making modern machine learning systems truly robust. 
	
	\smallskip
	\noindent \textbf{Goal of the paper.} To develop concept drift detector capable of efficient change detection, while being robust to adversarial and poisoning attacks that inject fake concept drifts.
	
	\smallskip
	\noindent \textbf{Summary of the content.} In this paper, we propose a holistic analysis of the adversarial concept drift problem, together with a robust framework for handling poisoning attacks coming from data streams. We discuss the nature of adversarial concept drift and introduce a taxonomy of different types of possible attacks -- based on poisoning instances or entire concepts. This allows us to understand the nature of the problem and gives a foundation for formulating a robust concept drift detector. We achieve this by introducing a novel and trainable drift detector based on Restricted Boltzmann Machine. It is capable of learning the compressed properties of the current state of stream and using a reconstruction error to detect the presence of concept drifts. In order to make it robust to adversarial and corrupted instances, we use a robust online gradient descent approach, together with a dedicated energy function used in our neural network. Finally, we introduce Relative Loss of Robustness -- a new measure for evaluating the robustness of concept drift detectors to varying levels of adversarial instances injected into the data stream. 
	
	\smallskip
	\noindent \textbf{Main contributions.} This paper offers the following contributions to the field of learning from drifting data streams:
	
	\begin{itemize}
		
		\item \textbf{Taxonomy of adversarial concept drifts:} we discuss the potential nature of poisoning attacks that may result in adversarial concept drift occurrence and their influence on drift detectors. We also formulate scenarios that can be used to evaluate the robustness of an algorithm dedicated to data stream mining. 
		
		\smallskip
		\item \textbf{Robust concept drift detector:} we introduce a novel drift detector based on Restricted Boltzmann Machine that is augmented with robust online gradient procedure and dedicated energy function to alleviate the influence of poisoned instances on the detector.
		
		\smallskip
		\item \textbf{Measure of robustness to adversarial drift:} we propose Relative Loss of Robustness, an aggregated measure used to analyze the effects of various levels of adversarial drifts injected into the stream.
		
		\smallskip
		\item \textbf{Extensive experimental study:} we evaluate the robustness of the proposed and state-of-the-art drift detectors, using a carefully designed experimental test bed that involves fully and sparsely labeled data stream benchmarks.
		
	\end{itemize} 
	
	
	\section{Data stream mining}
	\label{sec:dsm}
	
	Data stream is defined as a sequence ${<S_1, S_2, ..., S_n,...>}$, where each element $S_j$ is a new instance. In this paper, we assume the (partially) supervised learning scenario with classification task and thus we define each instance as $S_j \sim p_j(x^1,\cdots,x^d,y) = p_j(\mathbf{x},y)$, where $p_j(\mathbf{x},y)$ is a joint distribution of the $j$-th instance, defined by a $d$-dimensional feature space and assigned to class $y$. Each instance is independent and drawn randomly from a probability distribution $\Psi_j (\mathbf{x},y)$.
	
	\smallskip
	\noindent \textbf{Concept drift.} When all instances come from the same distribution, we deal with a stationary data stream. In real-world applications data very rarely falls under stationary assumptions \citep{Masegosa:2020}. It is more likely to evolve over time and form temporary concepts, being subject to concept drift \citep{Lu:2019}. This phenomenon affects various aspects of a data stream and thus can be analyzed from multiple perspectives. One cannot simply claim that a stream is subject to the drift. It needs to be analyzed and understood in order to be handled adequately to specific changes that occur \citep{Goldenberg:2019,Goldenberg:2020}. More precise approaches may help us achieving faster and more accurate adaptation \citep{Shaker:2015}. Let us now discuss the major aspects of concept drift and its characteristics. 
	
	\smallskip
	\noindent \textbf{Influence on decision boundaries}. Firstly, we need to take into account how concept drift impacts the learned decision boundaries, distinguishing between real and virtual concept drifts \citep{Oliveira:2019}. The former influences previously learned decision rules or classification boundaries, decreasing their relevance for newly incoming instances. Real drift affects posterior probabilities $p_j(y|\mathbf{x})$ and additionally may impact unconditional probability density functions. It must be tackled as soon as it appears, since it impacts negatively the underlying classifier. Virtual concept drift affects only the distribution of features $\mathbf{x}$ over time:
	
	\begin{equation}
	\widehat{p}_j(\mathbf{x}) = \sum_{y \in Y} p_j(\mathbf{x},y),
	\label{eq:cd2}
	\end{equation}
	
	\noindent where $Y$ is a set of possible values taken by $S_j$. While it seems less dangerous than real concept drift, it cannot be ignored. Despite the fact that only the values of features change, it may trigger false alarms and thus force unnecessary and costly adaptations. 
	
	\smallskip
	\noindent \textbf{Locality of changes}. It is important to distinguish between global and local concept drifts~\citep{Gama:2006}. The former one affects the entire stream, while the latter one affects only certain parts of it (e.g., regions of the feature space, individual clusters of instances, or subsets of classes). Determining the locality of changes is of high importance, as rebuilding the entire classification model may not be necessary. Instead, one may update only certain parts of the model or sub-models, leading to a more efficient adaptation.
	
	\smallskip
	\noindent \textbf{Speed of changes}. Here we distinguish between sudden, gradual, and incremental concept drifts~\citep{Lu:2019}. 
	
	\begin{itemize}
		\item \textbf{Sudden concept drift} is a case when instance distribution abruptly changes with $t$-th example arriving from the stream:
		
		\begin{equation}
		p_j(\mathbf{x},y) =
		\begin{cases}
		D_0 (\mathbf{x},y),       & \quad \text{if } j < t\\
		D_1 (\mathbf{x},y),  & \quad \text{if } j \geq t.
		\end{cases}
		\label{eq:cd3}
		\end{equation}
		
		\smallskip
		\item \textbf{Incremental concept drift} is a case when we have a continuous progression from one concept to another (thus consisting of multiple intermediate concepts in between), such that the distance from the old concept is increasing, while the distance to the new concept is decreasing:
		
		\begin{equation}
		p_j(\mathbf{x},y) =
		\begin{cases}
		D_0 (\mathbf{x},y),       &  \text{if } j < t_1\\
		(1 - \alpha_j) D_0 (\mathbf{x},y) + \alpha_j D_1 (\mathbf{x},y),       &\text{if } t_1 \leq j < t_2\\
		D_1 (\mathbf{x},y),  &  \text{if } t_2 \leq j
		\end{cases}
		\label{eq:cd4}
		\end{equation}
		
		\noindent where

		\begin{equation}
		\alpha_j = \frac{j - t_1}{t_2 - t_1}.
		\label{eq:cd5}
		\end{equation}
		
		\smallskip
		\item \textbf{Gradual concept drift} is a case where instances arriving from the stream oscillate between two distributions during the duration of the drift, with the old concept appearing with decreasing frequency:
		
		\begin{equation}
		p_j(\mathbf{x},y) =
		\begin{cases}
		D_0 (\mathbf{x},y),       &  \text{if } j < t_1\\
		D_0 (\mathbf{x},y),       &  \text{if } t_1 \leq j < t_2 \wedge \delta > \alpha_j\\
		D_1 (\mathbf{x},y),       &  \text{if } t_1 \leq j < t_2 \wedge \delta \leq \alpha_j\\
		D_1 (\mathbf{x},y),  &  \text{if } t_2 \leq j,
		\end{cases}
		\label{eq:cd4}
		\end{equation}
		\noindent where $\delta \in [0,1]$ is a random variable. 
	\end{itemize}
	
	\smallskip
	\noindent \textbf{Recurrence}. In many scenarios it is possible that a previously seen concept from $k$-th iteration may reappear D$_{j+1}$ = D$_{j-k}$ over time \citep{Sobolewski:2017}. One may store models specialized in previously seen concepts in order to speed up recovery rates after a known concept re-emerges \citep{Guzy:2020}. 
	
	\smallskip
	\noindent \textbf{Presence of noise}. Apart from concept drift, one may encounter other types of changes in data. They are connected with the potential appearance of incorrect information in the stream and known as blips or noise. The former stands for singular random changes in a stream that should be ignored and not mistaken for a concept drift. The latter stands for significant corruption in the feature values or class labels and must be filtered out in order to avoid feeding false \citep{Krawczyk:2018} or even adversarial information to the classifier \citep{Sethi:2018}. 
	
	\smallskip
	\noindent \textbf{Feature drift}. This is a type of change that happens when a subset of features becomes, or stops to be, relevant to the learning task \citep{Barddal:2017}. Additionally, new features may emerge (thus extending the feature space), while the old ones may cease to arrive. 
	
	\smallskip
	\noindent \textbf{Drift detectors.} In order to be able to adapt to evolving data streams, classifiers must either have explicit information on when to update their model, or use continuous learning to follow the progression of a stream. Concept drift detectors are external tools that can be paired with any classifier and used to monitor a state of the stream \citep{Barros:2018}. Usually, this is based on tracking the error of the classifier \citep{Pinage:2020} or measuring the statistical properties of data \citep{Korycki:2019}. Drift detectors emit two-level signals. The warning signal means that changes start to appear in the stream and recent instances should be stored in a dedicated buffer. This buffer is used to train a new classifier in the background. The drift signal means that changes are significant enough to require adaptation and the old classifier must be replaced with the new one trained on the most recent buffer of data. This reduces the cost of adaptation by lowering the number of times when we train the new classifier, but may be subject to costly false alarms or missing changes appearing locally or on a smaller magnitude.
	
	One of the first and most popular drift detectors is Drift Detection Method (DDM) \citep{Gama:2004} that analyzes the standard deviation of errors coming from the underlying classifier. DDM assumes that the increase in error rates directly corresponds to changes in incoming data stream and thus can be used to signal the presence of drift. This concept was extended by Early Drift Detection Method (EDDM) \citep{Garcia:2006} by replacing the standard error deviation with a distance between two consecutive errors. This makes EDDM more reactive to slower, gradual changes in the stream, at the cost of losing sensitivity to sudden drifts. Reactive Drift Detection Method (RDDM) \citep{Barros:2017} is an improvement upon DDM that allows detecting sudden and local changes under access to a reduced number of instances. RDDM offers better sensitivity than DDM by implementing a pruning mechanism for discarding outdated instances. Adaptive Windowing (ADWIN) \citep{Bifet:2007} is based on a dynamic sliding window that adjusts its size according to the size of the stable concepts in the stream. ADWIN stores two sub-windows for old and new concepts, detecting a drift when mean values in these sub-windows differ more than a given threshold. Statistical Test of Equal Proportions (STEPD) \citep{Nishida:2007} also keeps two sub-window, but uses a statistical test with continuity correction to determine if instances in both windows originate from similar or different distributions. Page-Hinkley Test (PHT) \citep{Sebastiao:2017} measures the current accuracy, as well as mean accuracy over a window of older instances. PHT computes cumulative and minimum differences between those two values and compares them to a predefined threshold, assuming that higher values of cumulative differences indicate increasing presence of concept drift. Exponentially Weighted Moving Average for Concept Drift Detection (ECDD) \citep{Ross:2012} detects changes in the mean of sequences of instances realized as random variables, without the need for a prior knowledge about their mean and standard deviation. Sequential Drift (SEQDRIFT) \citep{Pears:2014} can be seen as an improved version of ADWIN, where the sub-window of instances from the old concept is obtained by a reservoir sampling with a single-pass approach. Additionally, the comparison between two windows is done using the Bernstein bound, alleviating the need for a user-specified threshold. SEED Drift Detector (SEED) \citep{Huang:2014} is another example of ADWIN extension, where the Hoeffding's inequality with Bonferroni correction is used as the threshold bound for comparing two sub-windows. Furthermore, SEED compresses its sub-windows by eliminating redundant instances from homogeneous parts of its windows. Drift Detection Methods based on the Hoeffding's bounds (HDDM) \citep{Blanco:2015} uses the identical bound as SEED, but drops the idea of sub-windows and focuses on measuring both false positive and false negative rates. Fast Hoeffding Drift Detection Method (FHDDM) \citep{Pesaranghader:2016} is yet another drift detector utilizing the popular Hoeffding's inequality, but its novelty lies in measuring the probability of correct decisions returned by the underlying classifier. Fisher Test Drift Detector (FTDD) \citep{Cabral:2018} is an extension of STEPD that addressees the issue of sub-windows being of insufficient size or holding imbalanced distributions. Wilcoxon Rank Sum Test Drift Detector (WSTD) \citep{Barros:2018w} is another extension of STEPD that uses the Wilcoxon rank-sum statistical test for comparing distributions in sub-windows. Diversity Measure as Drift Detection Method (DMDDM) \citep{Mahdi:2020} uses a combination of a pairwise diversity measure and PHT to offer an aggregated measure of difference between two concepts. 
	
	Recently, we can see the emergence of the ensemble learning paradigm applied to drift detectors \citep{Krawczyk:2017}. Dynamic Classifier Selection with Local Accuracy and Drift Detector (DCS-LA+DDM) \citep{Pinage:2020} uses an ensemble of base classifiers together with a single drift detector. However, while based on ensemble idea, one cannot consider it a true ensemble of drift detectors. Drift Detection Ensemble (DDE) \citep{Maciel:2015} uses a combination of three independent drift detectors, which can be seen as a heterogeneous ensemble drift detector. Stacking Fast Hoeffding Drift Detection Method (FHDDMS) \citep{Pesaranghader:2018} uses a sequential combination of FHDDM detectors. Ensemble Drift Detection with Feature Subspaces (EDFS) \citep{Korycki:2019} uses a combination of incremental Kolmogorov-Smirnov detectors \citep{Reis:2016} deployed on a set of diverse feature subspaces. Every time drift is detected, the subspaces are reconstructed, allowing EDFS to capture the new data characteristics. EDFS can be seen as a homogeneous ensemble drift detector.
	
	\smallskip
	\noindent \textbf{Classifiers for drifting data streams.}  Alternative approaches assume using classifiers that are capable of learning in an incremental or online manner. Sliding windows storing only the most recent instances are very popular, allowing for natural forgetting of older instances \citep{Ramirez-Gallego:2017tsmc,Roseberry:2019}. The size of the window is an important parameter and adapting it over time seems to yield the best results \citep{Bifet:2007}. Online learners are capable of learning instance by instance, discarding data after it passed the training procedure. They are very efficient on their own, but need to be equipped with a forgetting mechanism in order not to endlessly grow their complexity \citep{Yu:2019}. Adaptive Hoeffding Trees \citep{Bifet:2009} and gradient-based methods \citep{Jothimurugesan:2018} are among the most popular solutions. 
	
	\smallskip
	\noindent \textbf{Ensemble approaches.} Combining multiple classifiers is a very popular and powerful approach for standard learning problems \citep{Wozniak:2014}. The technique transferred seamlessly to data stream mining scenarios, where ensemble approaches have displayed a great efficacy \citep{Krawczyk:2017}. They not only offer improved predictive power, robustness, and reduction of variance, but also can easily handle concept drift and use it as a natural way of maintaining diversity. By encapsulating new knowledge in the ensemble pool and removing outdated models, one can assure that the base classifiers are continuously mutually complementary, while adapting to changes in the stream. Popular solutions are based on the usage of online versions of bagging \citep{Bifet:2010}, boosting \citep{Oza:2001}, Random Forest \citep{Gomes:2017}, or instance-based clustering \citep{Korycki:2018}, as well as dedicated architectures such as Accuracy Updated Ensemble \citep{Brzezinski:2014} or Kappa Updated Ensemble \citep{Cano:2020}.
	
	\section{Adversarial concept drift and poisoning attacks in streaming scenarios}
	\label{sec:apa}
	
	In this section, we will discuss: the unique characteristic of adversarial learning scenario in the context of drifting data streams, the inadequacy of existing detectors for handling adversarial instances, why adversarial drift cannot be simply disregarded as a noise, how it impacts the learning from data streams, and what type of poisoning attacks we may expect. 
	
	\smallskip
	\noindent \textbf{Adversarial learning for static data.} With the advent of deep learning and numerous success stories of its application in real-life problems, researchers started to notice that deep models can be easily affected by corrupted or noisy information \citep{Elsayed:2018}. Multiple studies reported that even small perturbations in the training data may have devastating effects on the efficacy of the neural model \citep{Su:2019}. While learning in the presence of noisy instances (either in a form of feature noise \citep{Adeli:2019} or label noise \citep{Frenay:2014}) has been studied thoroughly in machine learning, the specifics of deep learning gave a rise to novel challenges \citep{Wang:2020}. As deep learning usually deals with complex representations of data, such as images \citep{Dong:2020} or text \citep{Wallace:2019}, noise can be introduced in various new forms and affect multiple types of outputs generated by deep models: predictions, extracted features, embeddings, or generated instances \citep{Choi:2018}. Adversarial learning is nowadays used in the broad context of preparing models to deal with noisy and corrupted data \citep{Zhang:2016}, ether by evasion \citep{Li:2020} or poisoning attack \citep{Bojchevski:2019} training schemes. It is important to note that in the literature the term adversarial learning is used for both scenarios with a malicious party attacking the model \citep{Madry:2018} and scenarios where corrupted information is the effect of environmental factors \citep{Kaneko:2020}. As for the effect on data, it is either assumed that the training set is already corrupted \citep{Cohen:2020}, or that the corruption may appear during the prediction phase and thus the training set must be enriched in order to prepare the model \citep{Xiao:2018}. 
	
	\smallskip
	\noindent \textbf{Unique characteristics of adversarial learning in data streams.} Works in adversarial learning assume the static nature of data and the fact that the true nature of classes is known beforehand. Therefore, one may generate instances that differ from the training data in order to predict the nature of noisy or corrupted instances. In this work, we propose to extend the concept of adversarial learning into the data stream scenario with concept drift presence. This poses massive new challenges for the learning algorithms that cannot be tackled by the existing adversarial models. Let us now discuss the unique challenges that are present in the adversarial set-up of learning from data streams.
	
	\begin{itemize}
		\smallskip
		\item \textbf{The true state of classes is subject to change over time.} Data streams are subject to change over time. Therefore, what could be considered an adversarial case, may become a valid instance from a drifted distribution. This prevents us from simply enhancing the training set with artificial adversarial instances, as we cannot know a priori if any changes in the stream originate from the actual drift presence or from an adversarial attack.
		
		\smallskip
		\item \textbf{Adversarial concept drift should be treated as malicious.} In order to inject an adversarial concept drift into data, an attacker must be aware of the nature of data. Through this work, we will consider a scenario with a malicious party providing corrupted, poisoned data with a clear aim of damaging and harming our learning system. Similarities to insider attacks \citep{Tuor:2017} can be drawn, as in both cases poisoning data is prepared in such a way that eludes clear early detection as outliers and damages the learning system over time. 
		
		\smallskip
		\item \textbf{Need to differentiate between valid and adversarial concept drift.} We must always assume that the analyzed stream may be subject to a valid, non-malicious concept drift. This poses a very interesting challenge - how can we differentiate between the changes that we want to follow and changes that may be of adversarial nature? Here, we should assume that concept drift will become more and more present as the stream progresses, while adversarial attacks may have a periodic nature and usually constitute only a small portion of the incoming instances. It is highly unlikely that we will have equal proportions of valid and adversarial instances in the stream.
		
		\smallskip
		\item \textbf{Robustness to adversarial data cannot hinder the adaptation process.} While designing robust machine learning algorithms for adversarial concept drift, we cannot follow the standard procedure that treats all data different from the training set as adversarial ones. A robust learner that treats all new data distributions as corrupted will not be able to adapt to new concepts. We need to design novel algorithms capable of differentiating between valid and adversarial drifts, which in turn will be used to make autonomous and on-the-fly decisions whether the learner should adapt to new data or not. Therefore, we find creating robust concept drift detectors particularly promising direction.
	\end{itemize}
	
	\noindent \textbf{Limitations of existing drift detectors.} State-of-the-art drift detectors, discussed in details in Section~\ref{sec:dsm}, share common principles. They all monitor some characteristics of newly arriving instances from the data stream and use various thresholds to decide if the new data are different enough to signal concept drift presence. It is easy to notice that the used measures and statistical tests are very sensitive to any perturbations in data and offer no robustness to poisoning attacks. Existing drift detectors concentrate on checking the level of difference between two distributions, without actually analyzing the content of the newly arriving instances. Furthermore, they are realized as simple thresholding modules, not being able to adapt themselves to data at hand. This calls for new drift detectors that have enhanced robustness and can learn properties of data, instead of just measuring some simple statistics. 
	
	\smallskip
	\noindent \textbf{Noisy data streams vs adversarial concept drift.} It is important to offer a clear distinction between noisy data streams and adversarial concept drift. The former cases assume that either some features or labels have been incorrectly measured or provided by annotators. They do not have any underlying characteristics and usually come from additive random distributions or human errors. They may misguide the training procedure, but there is no malicious intent associated with them. Adversarial concept drift and associated poisoning attacks assume that there is a malicious party actively working against our machine learning system. The injected poisoned data was designed and crafted in such a way that will purposefully damage the classifier while avoiding easy detection. Adversarial drift must be seen as a coordinated attempt to disturb or completely shut down the system under attack. Negative impacts of poisoning attacks will be discussed next. 
	
	\smallskip
	\noindent \textbf{Negative impact of adversarial concept drift.} In order to offer a complex and holistic view on the problem of adversarial concept drift, let us now discuss the major difficulties imposed by such a phenomenon on the underlying machine learning system:
	
	\begin{itemize}
		\item \textbf{Forcing false and unnecessary adaptation.} Data stream mining algorithms, primarily detectors or adaptive classifiers, actively monitor the characteristics of the stream and trigger changing the learning model when newly arriving data differ from the previous concepts. Therefore, by injecting an adversarial drift during the stable period of a data stream (when no changes in data distributions are taking place), a false alarm will be raised. The learning algorithm will be cheated into thinking that a drift is taking place and will adapt to poisoned instances provided by the attacker. This can be used by a malicious party to damage the accuracy of the system at will, e.g., when two competitors are analyzing the same stream of data. Unnecessary adaptation not only will reduce the accuracy of the classifier, but also will consume precious computational resources and time, slowing the system down \citep{Zliobaite:2015} or even paralyzing it completely during the recovery after such an attack \citep{Shaker:2015}. 
		
		\smallskip
		\item \textbf{Impairing the adaptation process.} Adversarial concept drift may also be injected during the occurrence of a valid concept drift. If poisoning instances are well-crafted, they will offer information contrary to the actual changes in data. This will confuse any adaptive learning system, forcing it to adapt to two completely opposing concepts. If the algorithm deals with a high number of poisoned instances, this may lead to a complete nullification of the adaptation process. Even under the assumption that the adversarial concept is much smaller than the valid concept, it will still significantly slow down the adaptation process. The speed of adaptation to new data is of crucial importance and any tampering with it may result in massive financial losses or staying behind the competition. 
		
		\smallskip
		\item \textbf{Hindering label query process in partially labeled streams.} In real-life scenarios, one does not have access to fully labeled streams. As it is impossible to obtain the ground truth for each newly arriving instance, active learning is being used to select the most useful instances for the label query \citep{Lughofer:2017}. Obtaining labels is connected with budget management (i.e., the monetary cost for paying the annotator or domain expert) and time (i.e., how quickly the expert can label instance under the consideration). Adversarial concept drift will produce a number of instances that will pretend to be useful for the classifier (as they seem to originate from a novel concept). Even if the domain expert will correctly identify them as adversarial instances, the budget and time have already been spent. Therefore, adversarial concept drift is particularly dangerous for partially labeled streams, where it may force misuse of already scarce resources \citep{Zliobaite:2015}. 
	\end{itemize}
	
	\noindent \textbf{Taxonomy of adversarial concept drift.} We have analyzed and understood the unique nature of adversarial concept drift, the challenges connected with its differentiation from a valid concept drift, and the negative impacts it may have on the learning algorithms. Let us now propose the first taxonomy of poisoning attacks in the streaming scenario:
		
			\begin{figure}[h]
			\centering
			\begin{subfigure}{0.31\linewidth}{
					\includegraphics[width=\linewidth,trim=2cm 2cm 2cm 2cm,clip]{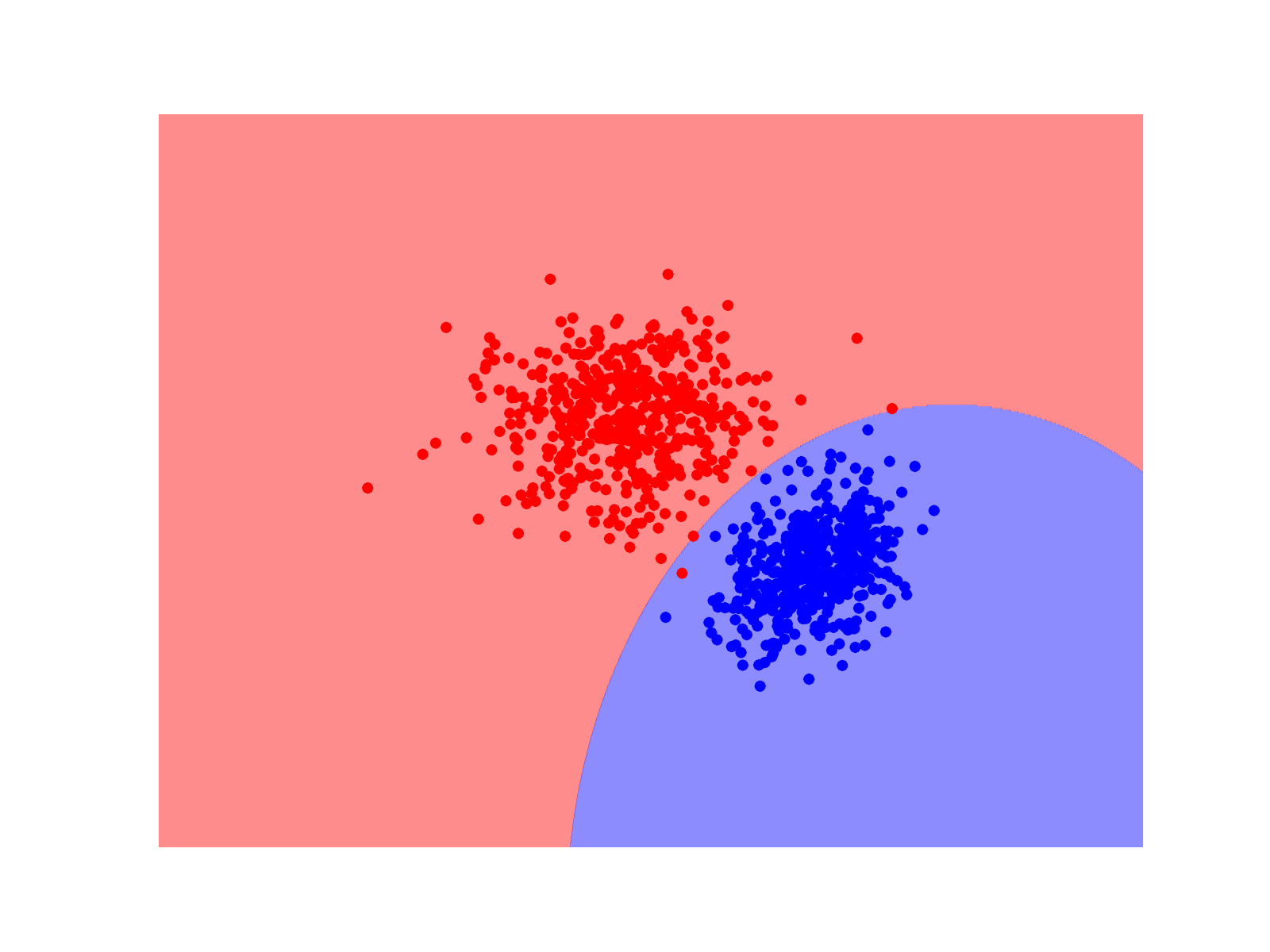}
					\subcaption{Before drift}}
			\end{subfigure}\hspace*{10pt}%
			\begin{subfigure}{0.31\linewidth}
				{\includegraphics[width=\linewidth,trim=2cm 2cm 2cm 2cm,clip]{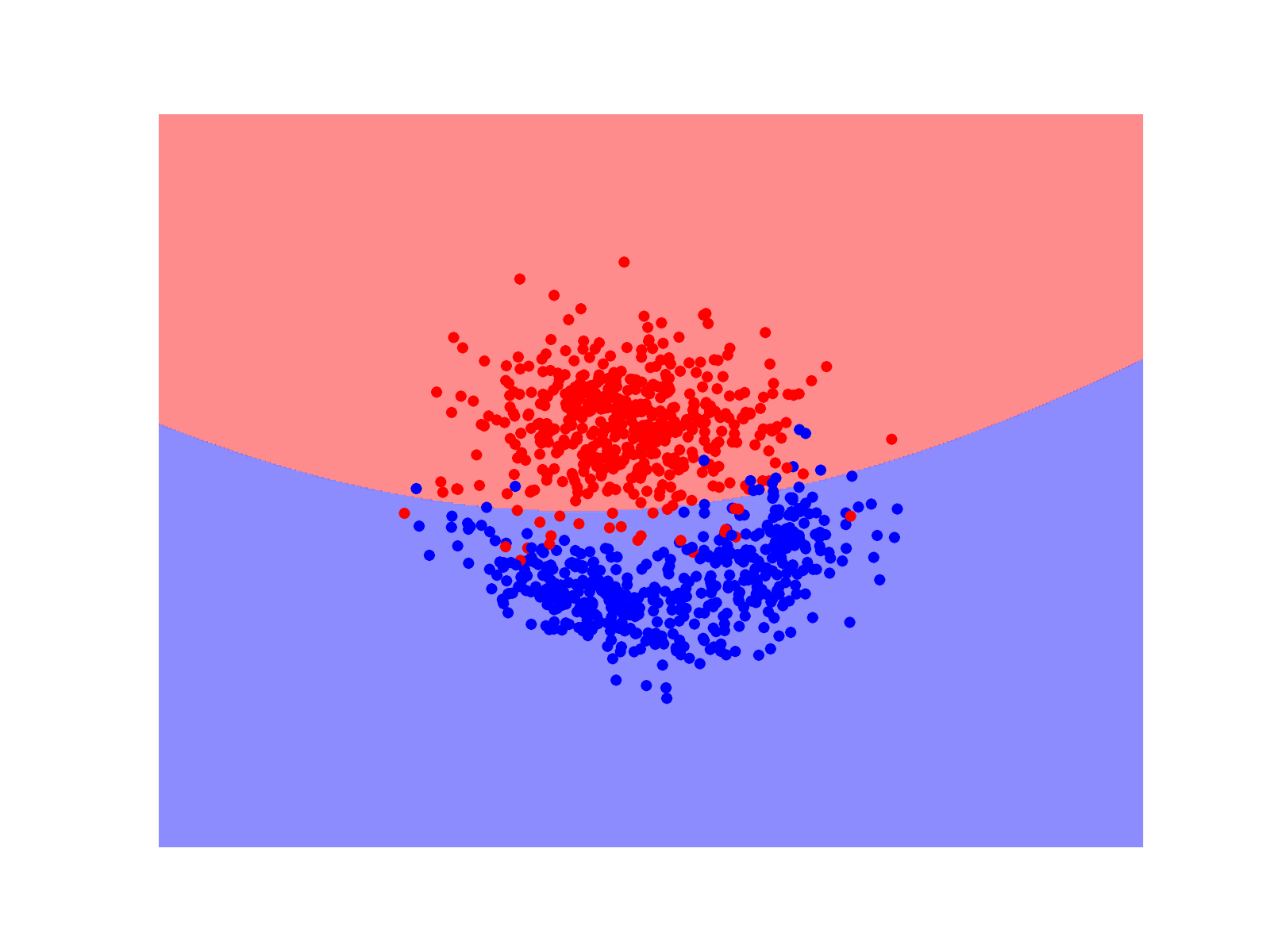}
					\subcaption{Valid drift}}
			\end{subfigure}\hspace*{10pt}%
			\begin{subfigure}{0.31\linewidth}
				{\includegraphics[width=\linewidth,trim=2cm 2cm 2cm 2cm,clip]{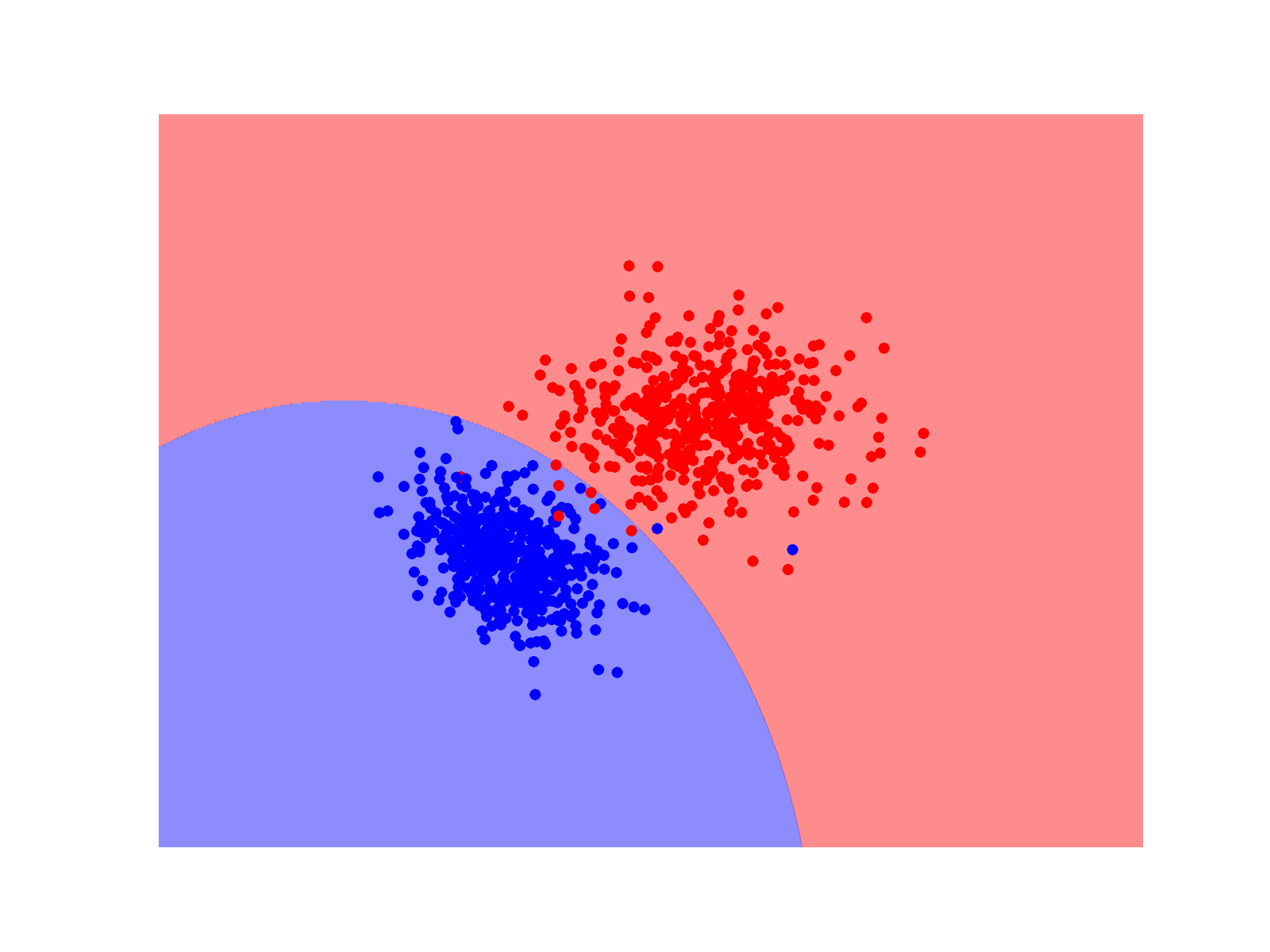}
					\subcaption{Second valid drift}}
			\end{subfigure}\vspace*{3pt}\\
			\caption{Accurate adaptation to valid concept drift.}
			\label{fig:dr1}
		\end{figure}
		
		\begin{figure}[h]
			\centering
			\begin{subfigure}{0.31\linewidth}{
					\includegraphics[width=\linewidth,trim=2cm 2cm 2cm 2cm,clip]{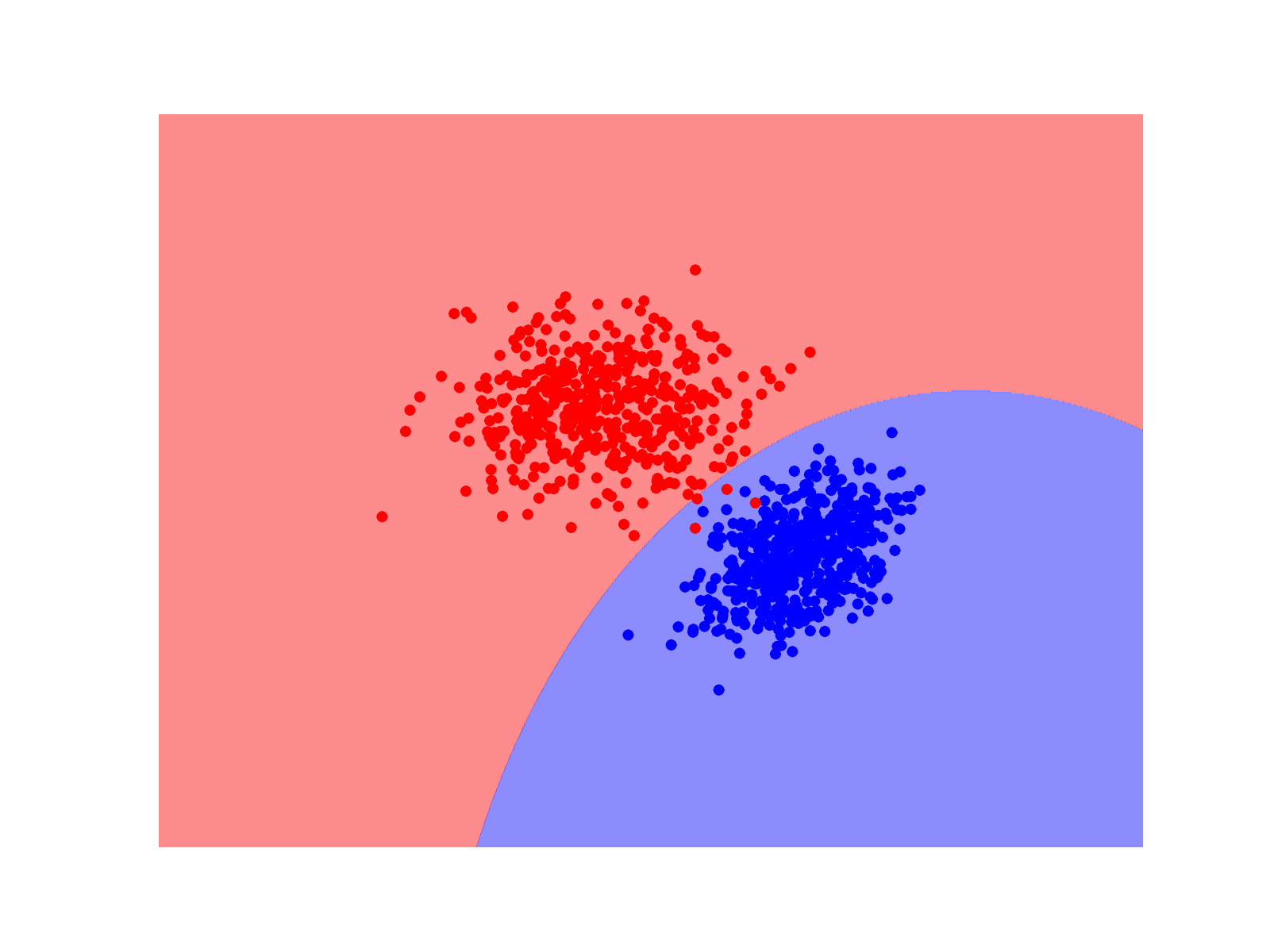}
					\subcaption{Before drift}}
			\end{subfigure}\hspace*{10pt}%
			\begin{subfigure}{0.31\linewidth}
				{\includegraphics[width=\linewidth,trim=2cm 2cm 2cm 2cm,clip]{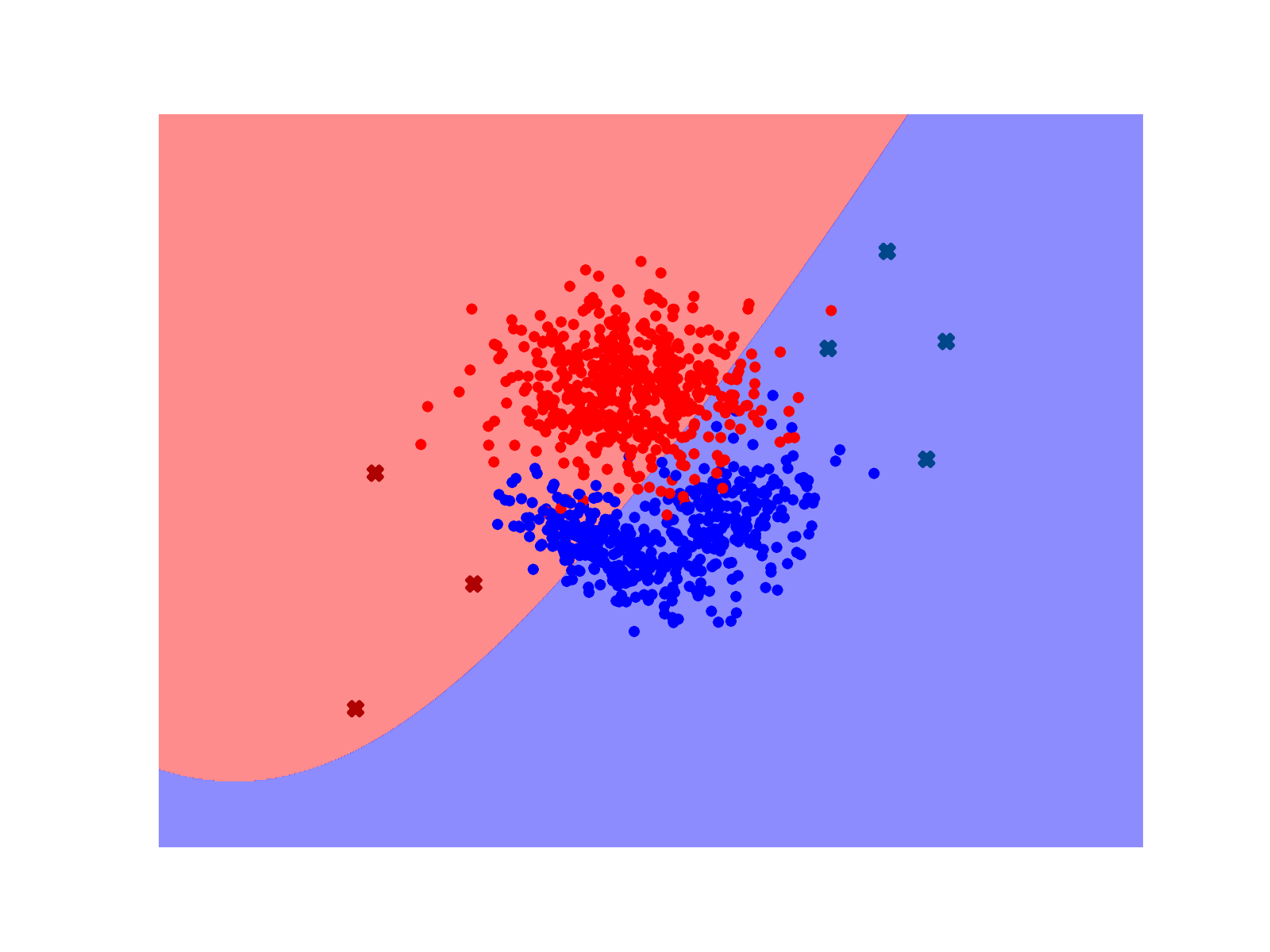}
					\subcaption{Hindered adaptation}}
			\end{subfigure}\hspace*{10pt}%
			\begin{subfigure}{0.31\linewidth}
				{\includegraphics[width=\linewidth,trim=2cm 2cm 2cm 2cm,clip]{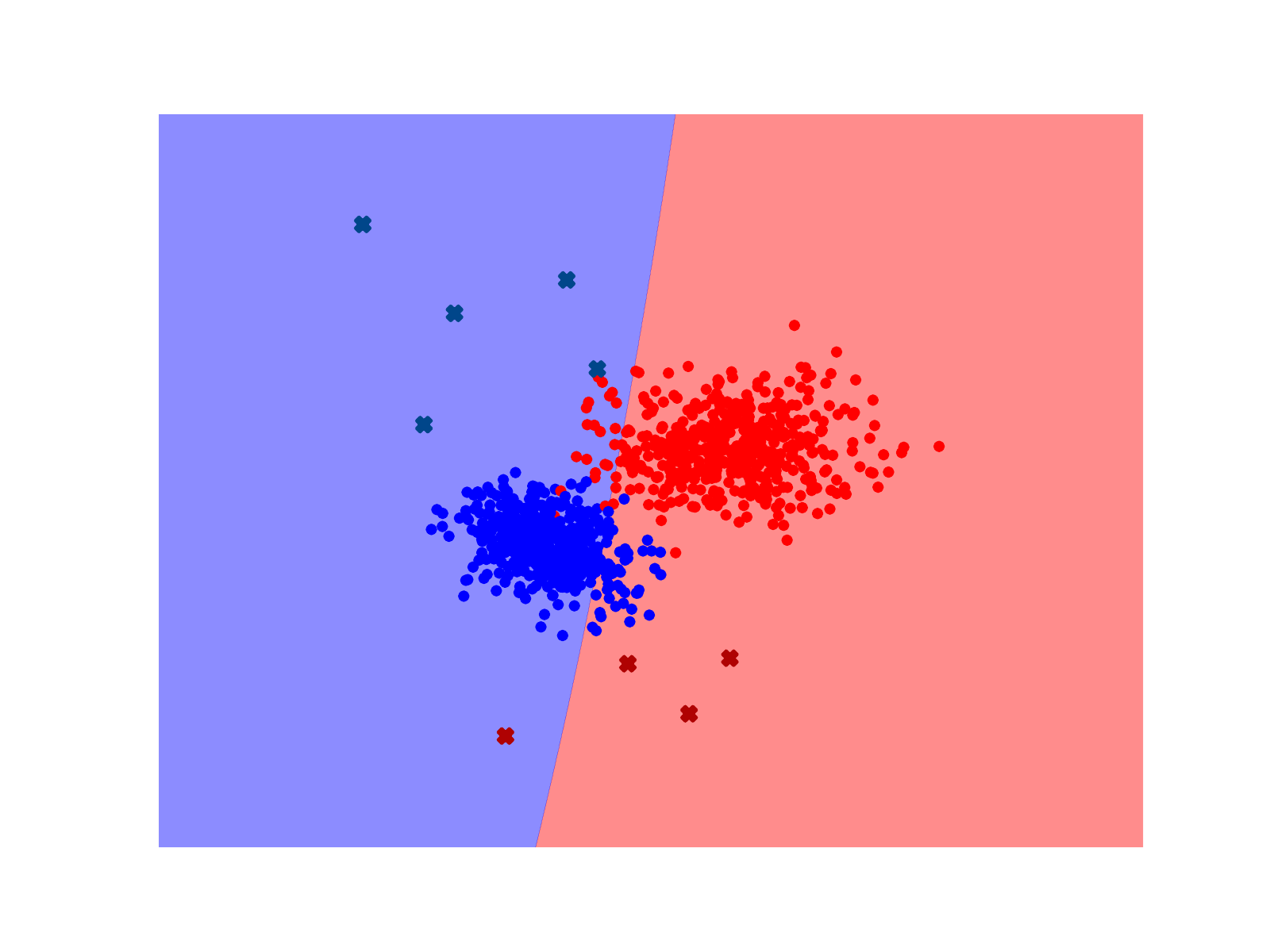}
					\subcaption{Overfitting}}
			\end{subfigure}\vspace*{3pt}\\
			\caption{Adversarial drift via instance-based poisoning attacks hinders (b) or exaggerates (c) the adaptation process.}
			\label{fig:dr2}
		\end{figure}
		
		\begin{figure}[h]
			\centering
			\begin{subfigure}{0.31\linewidth}{
					\includegraphics[width=\linewidth,trim=2cm 2cm 2cm 2cm,clip]{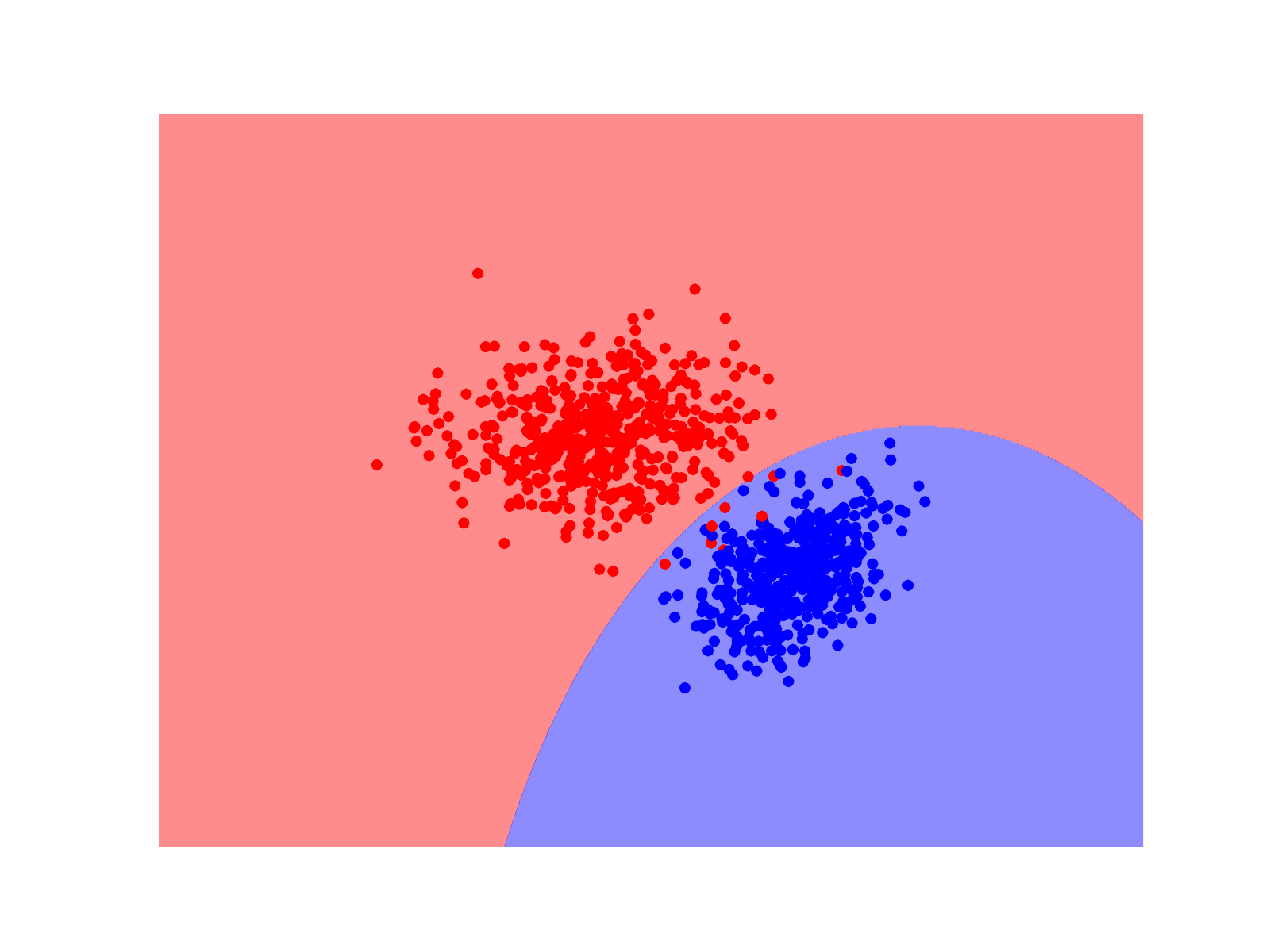}
					\subcaption{Before drift}}
			\end{subfigure}\hspace*{10pt}%
			\begin{subfigure}{0.31\linewidth}
				{\includegraphics[width=\linewidth,trim=2cm 2cm 2cm 2cm,clip]{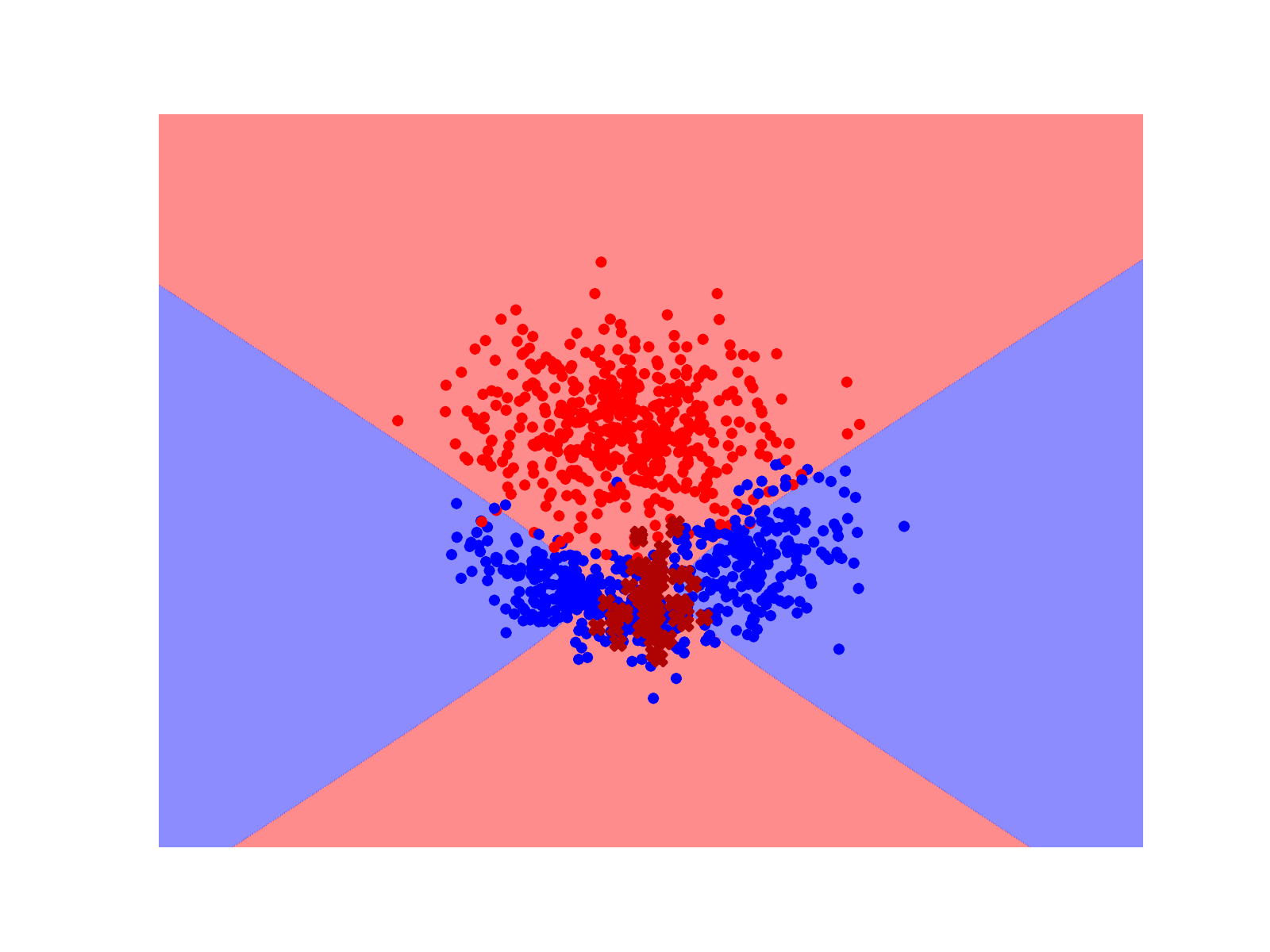}
					\subcaption{False adaptation}}
			\end{subfigure}\hspace*{10pt}%
			\begin{subfigure}{0.31\linewidth}
				{\includegraphics[width=\linewidth,trim=2cm 2cm 2cm 2cm,clip]{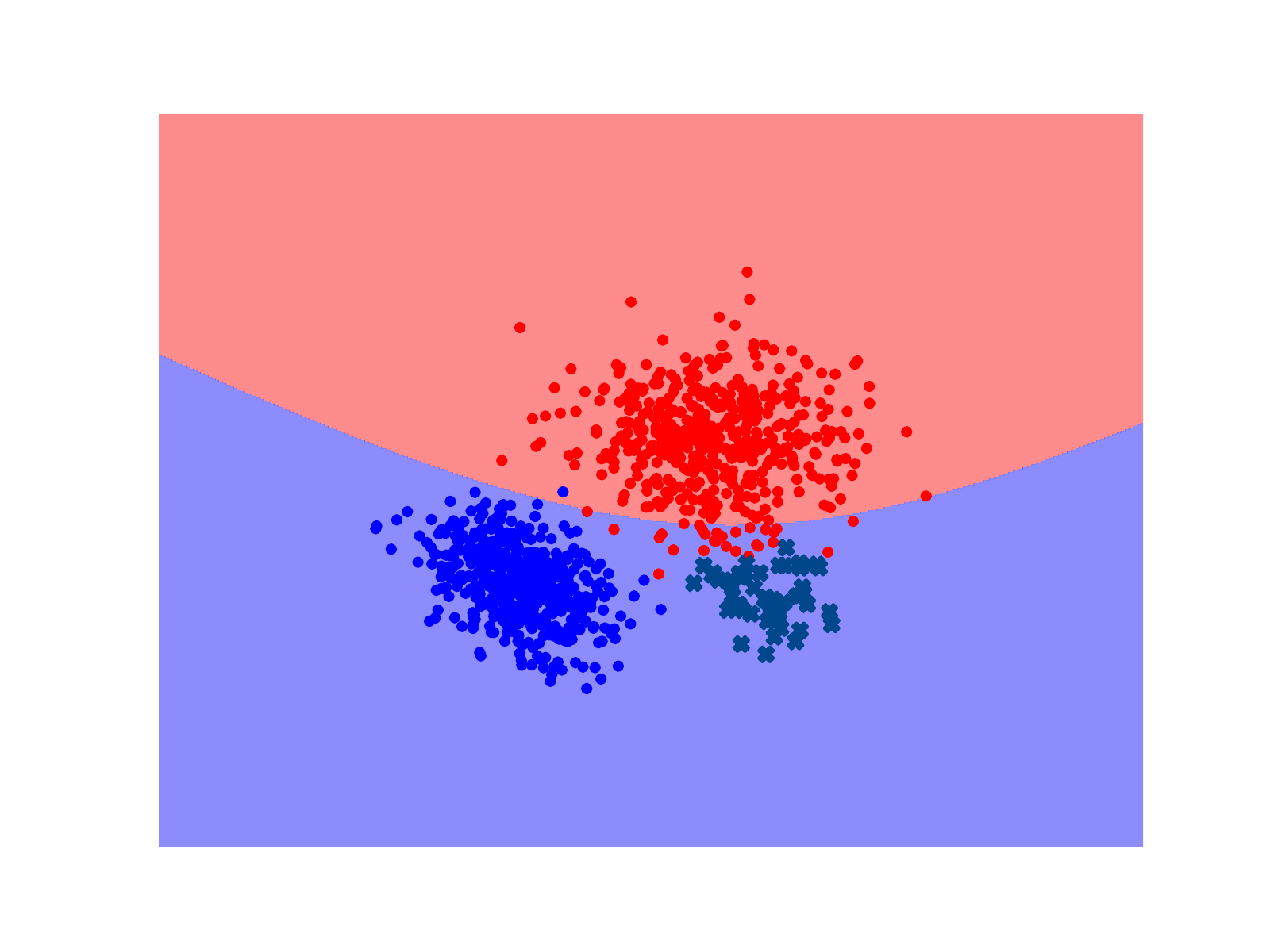}
					\subcaption{Lack of adaptation}}
			\end{subfigure}\vspace*{3pt}%
			\caption{Adversarial drift via concept-based poisoning attacks critically misguides (b) or completely nullifies (c) the adaptation process.}
			\label{fig:dr3}
		\end{figure}

	\begin{itemize}
	
	\smallskip
	\item \textbf{Adversarial concept drift with instance-based poisoning attacks.} The first type assumes that the malicious party injects singular corrupted instances into the stream. They may be corrupted original instances with flipped labels or with modified feature values. Instance-based attacks are common when the attacker wants to test the robustness of the system and still needs to learn about the distribution of data in the stream. Additionally, supplying independent poisoned instances, especially when done from multiple sources, may be harder to detect that injecting a large number of instances at once. Such attacks, at lower rates, may be picked up by noise/outlier detection techniques. However, these attacks will appear more frequently than natural anomalies and will be crafted with malicious intent, requiring dedicated methods to filter them out. Instance-based poisoning attacks will not cause a false drift detection, but may significantly impair the adaptation to the actual concept drift. Figure~\ref{fig:dr1} depicts the correct adaptation to a valid concept drift, while Figure~\ref{fig:dr2} depicts the same scenario with a hindered or exaggerated adaptation affected by the instance-based poisoning attacks.
	
	\smallskip
	\item \textbf{Adversarial concept drift with concept-based poisoning attacks.} The second type assumes that the malicious party have crafted poisoned instances that form a coherent concept. This can be seen as injecting an adversarial distribution of data that fulfills the cluster and smoothness assumptions. Therefore, now we must handle a difficult attack that will elude any outlier/noise/novelty detection methods. With the concept-based poisoning attack, we may assume that the malicious party poses a significant knowledge about the real data distributions and is able to craft such concepts that are going to directly cause false alarms and conflicts with valid concept drift. The effects of concept-based poisoning attacks are much more significant that its instance-based counterparts and may result, if undetected, in significant harm to the learning system and increased recovery times for rebuilding the model. Such attacks can both cause false drift detection and hinder, critically misguide or even completely nullify, the adaptation of the learning algorithm. Figure~\ref{fig:dr1} depicts the correct adaptation to a valid concept drift, while Figure~\ref{fig:dr2} depicts the same scenario with an incorrect adaptation thwarted by the concept-based poisoning attacks.
	
	\end{itemize}

	\section{Robust Restricted Boltzmann Machine for adversarial drift detection}
	\label{sec:rrb}
	
	In this section, we describe in detail the proposed concept drift detector, realized as a trainable Restricted Boltzmann Machine with enhanced robustness to adversarial instances. 
	
	\smallskip
	\noindent \textbf{Overview of the proposed method.} We introduce a novel concept drift detector that is characterized by an increased robustness to adversarial concept drift, while maintaining high sensitivity to valid concept drift. It is realized as a Restricted Boltzmann Machine with leveraged robustness via improved online gradient calculation and extended energy function. It is a fully trainable drift detector, capable of autonomous adaptation to the current state of the stream and not relying on user-defined thresholds or statistical tests. 
	
	\subsection{Restricted Boltzmann Machine}
	\label{sec:rbm}
	
	\noindent \textbf{Neural network architecture.}  Restricted Boltzmann Machines (RBMs) are a popular family of generative two-layered neural networks that are constructed using  the $\mathbf{v}$ layer of $V$ visible neurons and the $\mathbf{h}$ layer of $H$ hidden neurons:
	\begin{equation}
	\begin{split}
	\mathbf{v} = [v_1,\cdots, v_V] \in \{0,1\}^V,\\
	\mathbf{h} = [h_1,\cdots, h_H] \in \{0,1\}^H
	\end{split} 
	\label{eq:rbm1}
	\end{equation}
	
	As we deal with the task of supervised learning from data streams (as defined in Section~\ref{sec:dsm}), we need to extend this two-layer RBM architecture with the third, final $\mathbf{z}$ layer for class representation. It is implemented as a "one-hot" encoding, meaning that only a single neuron in $\mathbf{z}$ may activate at a given moment (i.e., have value set to 1, while every other one has value set to 0). By $\mathbf{1}_z$ we denote the vector of RBM outputs with $1$ returned by $z$-th neuron and $0$ returned by all other ones. This allows to define $\mathbf{z}$, known also as the class layer or the softmax layer:
	\begin{equation}
	\mathbf{z} = [z_1, \cdots, z_Z] \in {\mathbf{1}_1,\cdots, \mathbf{1}_Z}.
	\end{equation}
	
	\noindent This final, class layer uses the softmax function to estimate the probabilities of activation of each neuron in $\mathbf{z}$. 
	
	RBM assumes no connection between units in the same layer, which holds for $\mathbf{v}$, $\mathbf{h}$, and $\mathbf{z}$. Neurons in the visible layer $\mathbf{v}$ are connected with neurons in the hidden layer $\mathbf{h}$, and neurons in $\mathbf{h}$ are connected with those in the class layer $\mathbf{z}$. The weight assigned to a connection between the $i$-th visible neuron $v_i$ and the $j$-th hidden neuron $h_j$ is denoted as $w_{ij}$, while the weight assigned to a connection between the $j$-th hidden neuron $h_j$ and the $k$-th class neuron $z_k$ is denoted as $u_{jk}$. This allows us to define the RBM energy function:
	\begin{equation}
	E(\mathbf{v},\mathbf{h},\mathbf{z}) = - \sum_{i=1}^V v_i a_i - \sum_{j=1}^H h_j b_j - \sum_{k=1}^Z z_k c_k - \sum_{i=1}^V \sum_{j=1}^H v_i h_j w_{ij} - \sum_{j=1}^H \sum_{k=1}^Z  h_j z_k u_{jk},
	\label{eq:rbm2}
	\end{equation}
	
	\noindent where $a_i, b_j,$ and $c_k$ are biases introduced to $\mathbf{v}, \mathbf{h}$, and $\mathbf{z}$ respectively. Energy formula $E(\cdot)$ for state $[\mathbf{v},\mathbf{h},\mathbf{z}]$ can be used to calculate the probability of RBM of being in this state, using the Boltzmann distribution:
	
	\begin{equation}
	P(\mathbf{v},\mathbf{h},\mathbf{z}) = \frac{\exp \left( -E(\mathbf{v},\mathbf{h},\mathbf{z})\right)}{F},
	\label{eq:rbm3}
	\end{equation}
	
	\noindent where $F$ is a partition function allowing to normalize the probability $P(\mathbf{v},\mathbf{h},\mathbf{z})$ to 1.
	
	RBM assumes that its hidden neurons in $\mathbf{h}$ are independent and work with variables (features in the case of supervised learning) given by the visible layer $\mathbf{v}$. The activation probability of the $j$-th given neuron $h_j$ can be calculated as follows:
	\begin{equation}
	\begin{split}
	P(h_j |\mathbf{v},\mathbf{z}) = \frac{1}{1 + \exp\left(-b_j - \sum_{i=1}^V v_i w_{ij} - \sum_{k=1}^Z z_k u_{jk}\right)} \\ 
	= \sigma \left( b_j + \sum_{i=1}^V v_i w_{ij} + \sum_{k=1}^Z z_k u_{jk}\right),
	\end{split} 
	\label{eq:rbm4}
	\end{equation}
	
	\noindent where $\sigma(\cdot) = 1/(1+\exp(- \cdot))$ stands for a sigmoid function.
	
	The same assumption may be made for neurons in the visible layer $\mathbf{v}$, when values of neurons in the hidden layer $\mathbf{h}$ are known. This allows us to calculate the activation probability of the $i$-th visible neuron as:
	\begin{equation}
	P(v_i |\mathbf{h}) = \frac{1}{1 + \exp\left(-a_i - \sum_{j=1}^H h_j w_{ij}\right)} = \sigma \left( a_i + \sum_{j=1}^H h_j w_{ij} \right),
	\label{eq:rbm5}
	\end{equation}
	
	\noindent where one must note that given $\mathbf{h}$, the activation probability of neurons in $\mathbf{v}$ does not depend on $\mathbf{z}$. The activation probability of class layer (i.e., decision which class the object should be assigned to) is calculated using the softmax function:
	\begin{equation}
	P(\mathbf{z} = \mathbf{1}_k |\mathbf{h}) = \frac{\exp \left( - c_k - \sum_{j=1}^H h_j u_{jk}\right)}{\sum_{l=1}^Z \exp\left( - c_l - \sum_{j=1}^H h_j u_{jl}\right)},
	\label{eq:rbm6}
	\end{equation}
	
	\noindent where $k \in [1, \cdots, Z]$ and $k \neq l$.
	
	\smallskip
	\noindent \textbf{RBM training procedure.} As RBM is a neural network model, we may train it using a cost function $C(\cdot)$ minimization with a selected gradient descent method. RBM most commonly uses the negative log-likelihood of both external layers $\mathbf{v}$ and $\mathbf{z}$:
	\begin{equation}
	C(\mathbf{v},\mathbf{z}) = - \log\left( P(\mathbf{v},\mathbf{z}) \right).
	\label{eq:rbm7}
	\end{equation}
	
	\noindent By taking each independent weight $w_{ij}$, we may calculate now its gradient of the cost function:
	\begin{equation}
	\nabla C(w_{ij}) = \frac{\delta C(\mathbf{v},\mathbf{z})}{\delta w_{ij}} = \sum_{\mathbf{v},\mathbf{h},\mathbf{z}} P(\mathbf{v},\mathbf{h},\mathbf{z}) v_i h_j - \sum_{\mathbf{h}} P(\mathbf{h}|\mathbf{v},\mathbf{z})  v_i h_j.
	\label{eq:rbm8}
	\end{equation}
	
	\noindent This equation allows us to calculate the cost function gradient for a single instance. However, concept drift cannot be detected by analyzing individual instances independently. If we would base our change detection on variations induced by a single new instance, we would be highly sensitive to even the smallest noise ratio. Therefore, the properties of the stream should be analyzed over a batch of most recent instances. We propose to define RBM model for learning on mini-batches of instances. For a mini-batch of $n$ instances arriving in $t$ time $\mathbf{M}_t = {x_1^t,\cdots, x_n^t}$, we can rewrite the gradient from Eq.~\ref{eq:rbm8} using expected values:
	\begin{equation}
	\frac{\delta C(\mathbf{M}_t)}{\delta w_{ij}} = E_{\text{model}}[v_i h_j] - E_{\text{data}}[v_i h_j],
	\label{eq:rbm9}
	\end{equation}
	
	\noindent where $E_{\text{data}}$ is the expected value over the current mini-batch of instances and $E_{\text{model}}$ is the expected value from the current state of RBM. Of course, we cannot trace directly the value of $E_{\text{model}}$, therefore we must approximate it using Contrastive Divergence with $k$ Gibbs sampling steps to reconstruct the input data (CD-$k$):
	\begin{equation}
	\frac{\delta C(\mathbf{M}_t)}{\delta w_{ij}} \approx E_{\text{recon}}[v_i h_j] - E_{\text{data}}[v_i h_j].
	\label{eq:rbm10}
	\end{equation}
	
	After processing the $t$-th mini-batch $\mathbf{M}_t$, we can update the wights in RBM using any gradient descent method as follows:	
	\begin{equation}
	w_{ij}^{t+1} = w_{ij}^{t} - \eta \left( E_{\text{recon}}[v_i h_j] - E_{\text{data}}[v_i h_j]\right),
	\label{eq:rbm11}
	\end{equation}
	
	\noindent where $\eta$ stands for the learning rate of the RBM neural network. The way to update the $a_i$, $b_j$, and $c_k$ biases, as well as weights $u_{jk}$ is analogous to Eq.~\ref{eq:rbm11} and can be expressed as:
	
	\begin{equation}
	a_{i}^{t+1} = a_{i}^{t} - \eta \left( E_{\text{recon}}[v_i] - E_{\text{data}}[v_i]\right),
	\label{eq:rbm12}
	\end{equation}
	
	\begin{equation}
	b_{j}^{t+1} = b_{j}^{t} - \eta \left( E_{\text{recon}}[h_j] - E_{\text{data}}[h_j]\right),
	\label{eq:rbm13}
	\end{equation}
	
	\begin{equation}
	c_{k}^{t+1} = c_{k}^{t} - \eta \left( E_{\text{recon}}[z_k] - E_{\text{data}}[z_k]\right),
	\label{eq:rbm14}
	\end{equation}
	
	\begin{equation}
	u_{jk}^{t+1} = u_{jk}^{t} - \eta \left( E_{\text{recon}}[h_j z_k] - E_{\text{data}}[h_j z_k]\right).
	\label{eq:rbm15}
	\end{equation}
	
	\subsection{Drift detection with Robust RBM}
	\label{sec:drd}
	
	While RBM is a generative neural network model, we can use it as an explicit drift detector. The RBM model store compressed characteristics of the distribution of data it was trained on. Therefore, by using any similarity measure between the information stored in RBM and properties of newly arrived instances, one may evaluate if there are any changes in the distribution. This allows us to use RBM as a drift detector, introducing our Restricted Boltzmann Machine for Drift Detection (RBM-DD). Our RBM-DD model uses its similarity measure for monitoring the state of the stream and the level to which the newly arrived instances differ from the previous concepts. One should notice that RBM-DD is a fully trainable drift detector, capable not only of capturing the trends in a single evaluation measure (like most of drift detectors do), but also of learning and adapting to the current state of the stream. This makes it a highly attractive approach for handling difficult and rapidly changing streams where non-trainable drift detectors tend to fail. 
	
	\smallskip
	\noindent \textbf{Measuring data similarity}. In order to evaluate the similarity of newly arrived instances to old concepts stored in RBM-DD, we will use the reconstruction error. It is calculated for every instance independently in an online fashion, by inputting a newly arrived $d$-dimensional instance $S_n = [x_1^n, \cdots, x_d^n, y^n]$ to the $\mathbf{v}$ layer of RBM. Then values of neurons in $\mathbf{v}$ are calculated to reconstruct the feature values. Finally, class layer $\mathbf{z}$ is activated and used to reconstruct the class label. We can denote the reconstructed vector as:
	\begin{equation}
	\tilde{S}_n = [\tilde{x}_1^n, \cdots, \tilde{x}_d^n, \tilde{y}_1^n, \cdots, \tilde{y}_Z^n],
	\label{eq:rbm16}
	\end{equation}
	
	\noindent where the reconstructed vector features and labels are taken from probabilities calculated using the hidden layer:
	\begin{equation}
	\tilde{x}_i^n = P (v_i|h),
	\label{eq:rbm17}
	\end{equation}
	\begin{equation}
	\tilde{y}_k^n = P (z_k|h).
	\label{eq:rbm18}
	\end{equation}
	
	\noindent The $\mathbf{h}$ layer is taken from the conditional probability, in which the $\mathbf{v}$ layer is identical to the input instance:
	\begin{equation}
	\mathbf{h} \sim P(\mathbf{h}|\mathbf{v} = x^n, \mathbf{z} = \mathbf{1}_{y_n}).
	\label{eq:rbm19}
	\end{equation}
	
	\noindent This allows us to write the reconstruction error in a form of the mean squared error between the original and reconstructed instance:
	
	\begin{equation}
	R(S_n) = \sqrt{\sum_{i=1}^d (x_i^n - \tilde{x}_i^n)^2 + \sum_{k=1}^Z(\mathbf{1}^{y_n}_k - \tilde{y}_k^n)^2}.
	\label{eq:rbm19}
	\end{equation}
	\smallskip
	
	For the purpose of a stable concept drift detector, we do not look for a change in distribution over a single instance, but in a change of distribution over the newly arriving batch of instances. Therefore, we need to calculate the average reconstruction error over the recent mini-batch of data:
	
	\begin{equation}
	R(\mathbf{M}_t) = \frac{1}{n} \sum_{m=1}^n R(x_m^t).
	\label{eq:rbm20}
	\end{equation}
	
	\smallskip
	\noindent \textbf{Adapting reconstruction error to drift detection.} In order to make the reconstruction error a practical measure for detecting the presence of concept drift, we propose to measure the evolution of this measure (i.e., its trends) over arriving mini-batches of instances. We achieve this by using the well-known sliding window technique that will move over the arriving mini-batches. Let us denote the trend of reconstruction error over time as $Q_r(t)$ and calculate it using the following equation:
	
	\begin{equation}
	Q_r(t) = \frac{\bar{n}_t \bar{TR}_t - \bar{T}_t \bar{R}_t}{\bar{n}_t \bar{T^2}_t - (\bar{T}_t)^2 }.
	\label{eq:rbm21}
	\end{equation}
	\smallskip
	
	\noindent The trend over time can be computed using a simple linear regression, with the terms in Eq.~\ref{eq:rbm21} being simply sums over time as follows:
	
	\begin{equation}
	\bar{TR}_t = \bar{TR}_{t-1} + t R(\mathbf{M}_t),
	\label{eq:rbm22}
	\end{equation}
	
	\begin{equation}
	\bar{T}_t = \bar{T}_{t-1} + t,
	\label{eq:rbm23}
	\end{equation}
	
	\begin{equation}
	\bar{R}_t = \bar{R}_{t-1} + R(\mathbf{M}_t),
	\label{eq:rbm24}
	\end{equation}
	
	\begin{equation}
	\bar{T^2}_t = \bar{T^2}_{t-1} + t^2,
	\label{eq:rbm25}
	\end{equation}
	\smallskip
	
	\noindent where $\bar{TR}_0 = 0$, $\bar{T}_0 = 0$, $\bar{R}_0 = 0$, and $\bar{T^2}_0 = 0$. We capture those statistics using a sliding window of size $W$. Instead of using a manually set size, which is inefficient for drifting data streams, we propose to use a self-adaptive window size \citep{Bifet:2007}. To allow flexible learning from various sizes of mini-batches, we must consider a case where $t > W$. Here, we must compute the terms for the trend regression using the following equations:
	
	\begin{equation}
	\bar{TR}_t = \bar{TR}_{t-1} + t R(\mathbf{M}_t) - (t - w)R(\mathbf{M}_{t-w}),
	\label{eq:rbm26}
	\end{equation}
	
	\begin{equation}
	\bar{T}_t = \bar{T}_{t-1} + t - (t - w),
	\label{eq:rbm27}
	\end{equation}
	
	\begin{equation}
	\bar{R}_t = \bar{R}_{t-1} + R(\mathbf{M}_t) - R(\mathbf{M}_{t-w}),
	\label{eq:rbm28}
	\end{equation}

	\begin{equation}
	\bar{T^2}_t = \bar{T^2}_{t-1} + t^2 - (t - w)^2.
	\label{eq:rbm29}
	\end{equation}
	\smallskip
	
	\noindent The required number of instances $\bar{n}_t$ to compute the trend of $Q_r(t)$ as time $t$ is given as follows:
	\begin{equation}
	\bar{n}_t = 
	\begin{cases}
	t & \quad \text{if } t \leq w\\
	w & \quad \text{if } t > w 
	\end{cases}	
	\label{eq:rbm30}
	\end{equation}
	\smallskip
	
\smallskip
\noindent \textbf{Drift detection.} The above Eq.~\ref{eq:rbm21} allows us to compute the trends for every analyzed mini-batch of data. In order to detect the presence of drift we need to have capability of checking if the new mini-batch differs significantly from the previous one. Our RBM-DD achieves this by using Granger causality test \citep{Sun:2008} on trends from subsequent mini-batches of data $Q_r(\mathbf{M}_t)$ and $Q_r(\mathbf{M}_{t+1})$. This is a statistical test that determines whether one trend is useful in forecasting another. As we deal with non-stationary processes we perform the variation of Granger causality test based on first differences \citep{Mahjoub:2020}. Accepted hypothesis means that it is assumed that there exist Granger causality relationship between $Q_r(\mathbf{M}_t)$ and $Q_r(\mathbf{M}_{t+1})$, which means there is no concept drift. If the hypothesis is rejected, RBM-DD signals the presence of concept drift. 
	
	\subsection{Introducing robustness to RBM}
	\label{sec:rob}
	
	The previous section introduced our RBM-DD -- a drift detection method based on Restricted Boltzmann Machine. While this novel trainable drift detector may offer a significant drift detection capabilities and adaptation to the current state of the data stream, it offers no robustness to adversarial concept drift. In this section, we discuss how to introduce robustness into RBM-DD, leading to Robust Restricted Boltzmann Machine for Drift Detection (RRBM-DD) that is capable of handling adversarial concept drift, while still displaying sensitiveness to the emergence of valid drifts.
	
	\smallskip
	\noindent \textbf{Robust gradient descent.} The first weak point of the RBM-DD in the adversarial scenario lies in its training phase. Adversarial instances may affect the update of our trainable drift detector and thus make it less robust to poisoning attacks over time. To avoid this, we propose to create RRBM-DD by replacing the original online cost gradient calculation in RBM-DD (see Eq.~\ref{eq:rbm8}) used to update weights (see Eq.~\ref{eq:rbm11}) by the robust gradient descent introduced in \citep{Holland:2019}. It postulates to rescale the instance values by using a soft truncation of potentially adversarial instances. It achieves this by using a class of $M$-estimators of location and scale. It introduces a robust truncation factor $\hat{\theta}$ to control the influence of adversarial instances during the weight update:
	\begin{equation}
	w_{ij}^{t+1} = w_{ij}^{t} - \eta \left( E_{\text{recon}}[v_i h_j] - \hat{\theta_i} E_{\text{data}}[v_i h_j]\right),
	\label{eq:rgd1}
	\end{equation}
	
	\noindent where $\hat{\theta_i}$ stands for the truncation factor for the $i$-th neuron in $\mathbf{v}$ (and thus for the $i$-th input feature) that is calculated as follows:
	
	\begin{equation}
	\hat{\theta_i} \in \arg \min_{\theta \in \mathbb{R}} \sum_{a=1}^n \rho \left( \frac{L_i(y_a;\mathbf{z}) - \theta}{s_i} \right),
	\label{eq:rgd2}
	\end{equation}
	\smallskip
		
	\noindent where $\rho$ is a convex, even function and $L_i(y_a;\mathbf{z})$ is a loss function between true and predicted class labels (0-1 loss function is commonly used here). The authors of the robust gradient descent \citep{Holland:2019} postulate that for $\rho(\cdot) = \cdot^2$ the estimated truncation factor $\hat{\theta_i}$ is reduced to the sample mean of the loss function, thus alleviating the impact of extreme (in our case adversarial) instances. Therefore, they recommend to take $\rho(\cdot) = o(\cdot^2)$ for the function argument $\rightarrow \pm \infty$.
	
	The parameter $s_i$ is a scaling factor used to ensure that consistent estimates take place irrespective of the order of magnitude of the observations. It is calculated as:
	\begin{equation}
	s_i = \hat{\sigma}_i \sqrt{n/\log(2\delta^{-1})},
	\label{eq:rgd3}
	\end{equation}
	\smallskip
	
	\noindent where $\delta \in (0,1)$ is the confidence level and $\hat{\sigma}_i$ stands for an dispersion estimate of the instances in the mini-batch:
	
	\begin{equation}
	\hat{\sigma}_i \in \left\{ \sigma > 0: \sum_{a=1}^n \chi \left( \frac{L_i(y_a;\mathbf{z}) - \gamma_i}{\sigma} \right) = 0  \right\},
	\label{eq:rgd4}
	\end{equation}
	\smallskip
	
	\noindent where $\chi: \mathbb{R} \rightarrow \mathbb{R}$ is an even function that satisfies $\chi(0) < 0$ and $\chi(\cdot) > 0$ for the function argument $\rightarrow \pm \infty$. This ensures that $\hat{\sigma}_i$ is an adequate measure of dispersion of the loss function about a pivot point $\gamma_i = \sum_{a=1}^n L_i(y_a;\mathbf{z}) / n$ \citep{Holland:2019}.
	
	Analogously to Eq.~\ref{eq:rgd1}, we use the robust truncation factor $\hat{\theta}$ to update remaining weights and parameters of RRBM-DD given by Eq.~\ref{eq:rbm12}-\ref{eq:rbm15}. 
	
	\smallskip
	\noindent \textbf{Robust energy function.} Another weak spot of the RBM-DD lies in its energy function (see Eq.~\ref{eq:rbm2}). It is a crucial part for calculating the probability of RBM-DD entering a given state and is used as the basis for the RBM-DD training procedure. In our case, as we use RBM as a trainable drift detector, energy function plays a crucial role in updating the detector with new instances and adapting it to the current state of the stream. RBM-DD energy function assumes working on clean (i.e., non-adversarial) data and thus can easily be corrupted by any poisoning attacks. Therefore, we propose to improve it for our RRBM-DD to alleviate its sensitivity to corrupted instances. We achieve this by adding a gating step to the visible layer $\mathbf{v}$, allowing for switching on and off neurons that may be strongly affected by noise (i.e., coming from adversarial sources). We add two new sets of variables: $\mathbf{\tilde{v}} = [\tilde{v}_1,\cdots, \tilde{v}_V]$ that stands for a Gaussian noise model for each neuron in the visible layer and $\mathbf{g} = [g_1,\cdots, g_V]$ that stands for a binary gating function for each neuron. This allows us to use gating to switch off neurons in  $\mathbf{v}$ that have a chance probability of being affected by noisy instances. As we train our RRBM-DD over mini-batches of data, this approach will effectively switch off the RRBM-DD training procedure for every new instance that is denoted as adversarial by our underlying Gaussian noise model. The robust energy function for RRBM-DD is expressed as follows:
	
	\begin{equation}
	\begin{split}
	E_R(\mathbf{v},\mathbf{\tilde{v}},\mathbf{h},\mathbf{z},\mathbf{g}) = \frac{1}{2}\sum_{i=1}^V g_i (v_i - \tilde{v_i})^2 - \sum_{i=1}^V v_i a_i - \sum_{j=1}^H h_j b_j - \sum_{k=1}^Z z_k c_k \\
	- \sum_{i=1}^V\sum_{j=1}^H v_i h_j w_{ij} - \sum_{j=1}^H \sum_{k=1}^Z  h_j z_k u_{jk} + \frac{1}{2}\sum_{i=1}^V \frac{(\tilde{v_i} - \tilde{b_i})^2}{\tilde{\sigma}_i^2},
	\label{eq:ref1}
	\end{split}
	\end{equation}
	\smallskip
	
	\noindent where the first (added) term stands for the gating interactions between $v_i$ and $\tilde{v}_i$, the original energy function models the clean data, the last (added) term stands for the noise model, and $\tilde{b_i}$ and $\tilde{\sigma}_i^2$ stands for mean and variance of the noise. If RRBM-DD assumes that the $i$-th neuron in $\mathbf{v}$ is corrupted by the adversarial instance ($g_i = 0$) then $\tilde{v}_i \sim \mathcal{N}(\tilde{v}_i|\tilde{b_i};\tilde{\sigma}_i^2)$.
	
	\subsection{Metric for evaluating robustness to poisoning attacks in streaming scenarios}
	\label{sec:met}
	
	Concept drift detectors are commonly evaluated by the prequential accuracy \citep{Hidalgo:2019} or prequential AUC/G-mean \citep{Korycki:2019bd} of the underlying classifier. The reasoning behind it is that the better the detection of drift offered by the evaluated algorithm, the faster and more accurate classifier adaptation becomes. However, this set-up is not sufficient for evaluating drift detectors in the adversarial concept drift setting. As we want to evaluate the robustness of a drift detector to poisoning attacks, we should attack the data stream with various levels of intensity to determine the breaking point of each algorithm. However, simple evaluation of a performance metric over varying levels of adversarial concept drift intensity does not give us information on how much our algorithm deteriorates over increasingly poisoned data. 
	
	We assume that we evaluate the robustness of drift detectors in a controlled experimental environment, where we can inject a desired level of adversarial concept drift into any benchmark data stream. Let us denote by $\mathbf{L} = [l_1, \cdots, l_a]$ the set of adversarial poisoning levels, ordered in the ascending order $l_1 < l_a$ from the smallest to the highest level of adversarial drift injection. Following the taxonomy introduced in Section 3, we can understand those levels as:
	
	\begin{itemize}
		\item for instance-based poisoning attacks -- the percentage of instances in a mini-batch considered as adversarial in the data stream,	
		
		\smallskip	
		\item for concept-based poisoning attacks -- the number of adversarial concepts injected into the data stream.
	\end{itemize}
	
	Inspired by works in noisy data classification \citep{Saez:2016}, we introduce a Relative Loss of Robustness (RLR) that compares the performance of a given algorithm on a clean data stream against its performance on a data stream with $l$-th level of adversarial concept drift injected:
	
	\begin{equation}
	RLR_l = \frac{M_{0} - M_{l}}{M_{0}}.
	\end{equation}
	\smallskip
	
	\noindent This is a popular approach used to evaluate the impact of noise on classifiers and can be directly adapted to the setting of adversarial drift detection. While it offers a convenient measure over a single level of poisoning, it becomes increasingly difficult to use when we want to evaluate the robustness of an algorithm over multiple poisoning levels. To alleviate this drawback, we propose an aggregated version of RLR measure over all considered adversarial poisoning levels:
	\begin{equation}
	RLR = \frac{\sum_{l=1}^{\#\mathbf{L}} \omega_l RLR_l}{\#\mathbf{L}},
	\label{eq:rlr}
	\end{equation}
	\smallskip
	
	\noindent where $\omega_l$ is the importance weight associated to the performance on $l$-the level of adversarial information introduced to data and $\sum_{l=1}^{\#\mathbf{L}} \omega_l = 1$. The weighting part allows the end-user to adjust the measurements accordingly to the analyzed problem. If the user is interested only in the overall robustness of a given method, then all weights can be set to $1/\#L$. However, if it becomes crucial to select a drift detector that can perform the best even under the high intensity of attacks, the user may assign higher weights to higher levels of adversarial drift injection.  
	
	\section{Experimental study}
	\label{sec:exp}
	
	This experimental study was designed to evaluate the usefulness of the proposed RRBM--DD and answer the following research questions:
	
	\medskip
	\noindent \textbf{RQ1:} Does RRBM-DD algorithm offers improved robustness to adversarial concept drift realized instance-based poisoning attacks? 
	
	\smallskip
	\noindent  \textbf{RQ2:} Does RRBM-DD algorithm offers improved robustness to adversarial concept drift realized concept-based poisoning attacks? 
	
	\smallskip
	\noindent  \textbf{RQ3:} Is RRBM-DD capable of preserving its robust characteristics when dealing with sparsely labeled data streams?
	
	\smallskip
	\noindent  \textbf{RQ4:} What are the impacts of individual components of RRBM-DD on its robustness to adversarial concept drift?
	
	\subsection{Data stream benchmarks}
	\label{sec:dat}

	\begin{table}[h]
		\centering
		\caption{Properties of artificial and real data stream benchmarks.}
\scalebox{0.9}{
		\begin{tabular}{llrrrl}
			\toprule
			Abbr. & Stream & Instances & Features & Classes & Drift \\ 
			\midrule
			\multicolumn{6}{l}{\textbf{Artificial data streams}}\\
			HYP$_{I}$ & Hyperplane & 1 000 000 & 10 & 2 & incremental\\
			LED$_S$ & LED & 1 000 000 & 24 & 10 & sudden\\
			RBF$_G$ & RBF & 1 000 000 & 40 & 20 & gradual\\
			RBF$_S$ & RBF & 1 000 000 & 20 & 10 & sudden\\
			SEA$_G$ & SEA &3 000 000 & 3 & 4 & gradual\\
			TRE$_S$ & RandomTree & 2 000 000 & 10 & 6 & sudden\\
			\midrule
			\multicolumn{6}{l}{\textbf{Real data streams}}\\
			ecbdl14& Protein Structure Prediction &  9 600 000 & 631 & 2 & mixed\\
			higgs &High-energy Physics Classification & 4 954 752 & 28 & 2 & unknown\\
			IntelLab&Intel Lab Sensors & 2 313 153 & 6 & 58 & mixed\\
			iot& IoT Botnet Attacks& 7 062 606 & 115 & 11 & mixed\\
			kddcup & KDD Intrusion Detection & 3 107 709 & 41 & 24 & mixed\\
			susy &Supersymmetric Particle Detection& 2 305 347 & 18 & 2 & unknown\\
			\bottomrule
			\label{tab:data}
		\end{tabular}
}
	\end{table}
	
	For the purpose of evaluating the proposed RRBM-DD, we selected 12 benchmark data streams, six of which were generated artificially using MOA environment \citep{Bifet:2010moa} and other six being real-world data streams coming from various domains, such as security, physics, and proteomics. Such a diverse mix allowed us to evaluate the effectiveness of RRBM-DD over a plethora of scenarios. Using artificial data streams allows us to control the specific nature of drift and where it occurs, while real-world streams offer challenging problems that are characterized by a mix of different learning difficulties. Properties of used data stream benchmarks are given in Table~\ref{tab:data}.
	
	\subsection{Experimental setup}
	\label{sec:set}
	
	Here, we will present the details of the experimental study design.
	
	\smallskip
	\noindent \textbf{Injection of adversarial concept drift.} As there are no real-world benchmarks for adversarial concept drift, we will inject poisoning attacks into our 12 benchmark data streams with the following procedures:
	
	\begin{itemize}
		\item \textbf{instance-based poisoning attacks} are injected by corrupting a ratio $\mathbf{L}_{inst} \in \{0.05,0.10,0.15,0.20,0.25\}$ of randomly selected instances in the stream by flipping their class labels to a randomly chosen another class.
		
		\smallskip
		\item \textbf{concept-based poisoning attacks} are injected by generating a number $\mathbf{L}_{conc} \in \{10,30,50,70,100\}$ of small artificial concepts of $n = 250$ instances using corresponding MOA data generators.
	\end{itemize} 

	\smallskip
 	\noindent \textbf{Sparsely labeled data streams.} For the experiment investigating the robustness under limited access to ground truth, we investigate the 12 benchmark data streams with $\{5\%,10\%,15\%,20\%,25\%,30\%\}$ of labeled instances.
	
	\smallskip
	\noindent \textbf{Reference concept drift detectors.} As reference methods to the proposed RRBM-DD, we have selected five state of the art concept drift detectors: Early Drift Detection Method (EDDM) \citep{Garcia:2006}, Exponentially Weighted Moving Average for Concept Drift Detection (ECDD) \citep{Ross:2012}, Fast Hoeffding Drift Detection Method (FHDDM) \citep{Pesaranghader:2016}, Reactive Drift Detection Method (RDDM) \citep{Barros:2017},and Wilcoxon Rank Sum Test Drift Detector (WSTD) \citep{Barros:2018w}. Parameters of all six drift detectors are given in Table~\ref{tab:ddp}.
	
		\begin{table}[h!]
		\centering
		\caption{Examined drift detectors and their parameters.}
\scalebox{0.85}{
		\begin{tabular}{lll}
			\toprule
			Abbr. & Name & Parameters \\ 
			\midrule
			EDDM & Early Drift Detection        & warning threshold $\alpha_w \in \{0.90,0.92,0.95,0.98\}$\\
				 &                              & drift threshold $\alpha_d \in \{0.80,0.85,0.90.0.95\}$\\
				 &								& min. no. of errors $e \in \{10,30,50,70\}$\\
			ECDD & EWMA for Drift Detection     & differentiation weights $\lambda \in \{0.1,0.2,0.3,0.4\}$\\
				 & 								& min. no. of errors $n = \{10,30,50,70\}$\\
	    	FHDDM& Fast Hoeffding Drift Detection & sliding window size $\omega \in \{25,50,75,100\}$\\
	    		 &								  & allowed error $\delta \in \{0.000001,0.00001,0.0001,0.001\}$ \\ 
	    	RDDM & Reactive Drift Detection     & warning threshold $\alpha_w \in \{0.90,0.92,0.95,0.98\}$\\
	    	&                                   & drift threshold $\alpha_d \in \{0.80,0.85,0.90.0.95\}$\\
	    	&								    & min. no. of errors $e \in \{10,30,50,70\}$\\   
	    	& 							        & min. no. of instances $\min \in \{3000,5000,7000,9000\}$\\
	    	&                                   & max. no. of instances $\max \in \{10000,20000,30000,40000\}$\\
	    	&								    & warning limit $wL \in \{800,1000,1200,1400\}$\\   
	    	WSTD & Wilcoxon Rank Sum Test       & sliding window size $\omega \in \{25,50,75,100\}$\\
	    	&Drift Detection                                   & warning significance $\alpha_w \in \{0.01,0.03,0.05,0.07\}$\\
	    	&								    & drift significance $\alpha_d \in \{0.001,0.003,0.005,0.007\}$\\  
	    	& 							        & max. no of old instances $\min \in \{1000,2000,3000,4000\}$\\
			\midrule
			RRBM--DD & Robust RBM Drift Detection& mini--batch size $\mathbf{M} \in \{25,50,75,100\}$\\
			         &									  &	visible neurons $\mathbf{V} = $ no. of features\\ 
		           	&									  &	hidden neurons $\mathbf{H} \in \{0.25\mathbf{V},0.5\mathbf{V},0.75\mathbf{V},\mathbf{V}\}$ \\ 
			         &									  &	class neurons $\mathbf{Z} = $ no. of classes\\ 
			         &									  & learning rate $\eta \in \{0.01,0.03,0.05,0.07\}$\\
			         &                                    & Gibbs sampling steps $k \in \{1,2,3,4\}$\\
			         &                                    & robust gradient confidence $\delta \in \{0.90,0.92,0.95,0.98\}$\\
			\bottomrule
			\label{tab:ddp}
		\end{tabular}
}
	\end{table}

	\smallskip
	\noindent \textbf{Parameter tuning.} To offer a fair and thorough comparison, we perform parameter tuning for every drift detector and for every data stream benchmark. As we deal with a streaming scenario, we use self hyper-parameter tuning \citep{Veloso:2018} that is based on online Nelder \& Mead optimization.

	\smallskip
	\noindent \textbf{Ablation study.} In order to be able to answer \textbf{RQ4} we perform an ablation study to check the impact of individual introduced components on the robustness of the drift detector. We compare RRBM--DD with its simplified versions with only robust online gradient RBM--DD$_{RG}$, only robust energy function RBM--DD$_{RE}$, and basic RBM--DD with no explicit robustness mechanisms. 
	
	\smallskip
	\noindent \textbf{Base classifier.} To ensure fairness in comparison among examined drift detectors they all use Adaptive Hoeffding Decision Tree \citep{Bifet:2009} as a base classifier. 
	
	\smallskip
	\noindent \textbf{Evaluation metrics.} As we deal with drifting data streams, we evaluated examined algorithms using prequential accuracy \citep{Hidalgo:2019} and the proposed RLR metric (see Eq.~\ref{eq:rlr}). RLR assumes equal weights assigned to each level of adversarial drift injection.
	
	\smallskip
	\noindent\textbf{Windows.} We used a window size $W = 1000$ for window-based drift detectors and calculating the prequential metrics . 
	
	\smallskip
	\noindent\textbf{Statistical analysis.} We used the Friedman ranking test with Bonferroni-Dunn post-hoc for determining statistical significance over multiple comparisons with significance level $\alpha = 0.05$. 
	
	\subsection{Experiment 1: Evaluating robustness to instance-based poisoning attacks}
	\label{sec:exp1}

\begin{figure}[h]
			\centering
			\includegraphics[width=0.3\linewidth,trim=2cm 2cm 2cm 2cm,clip]{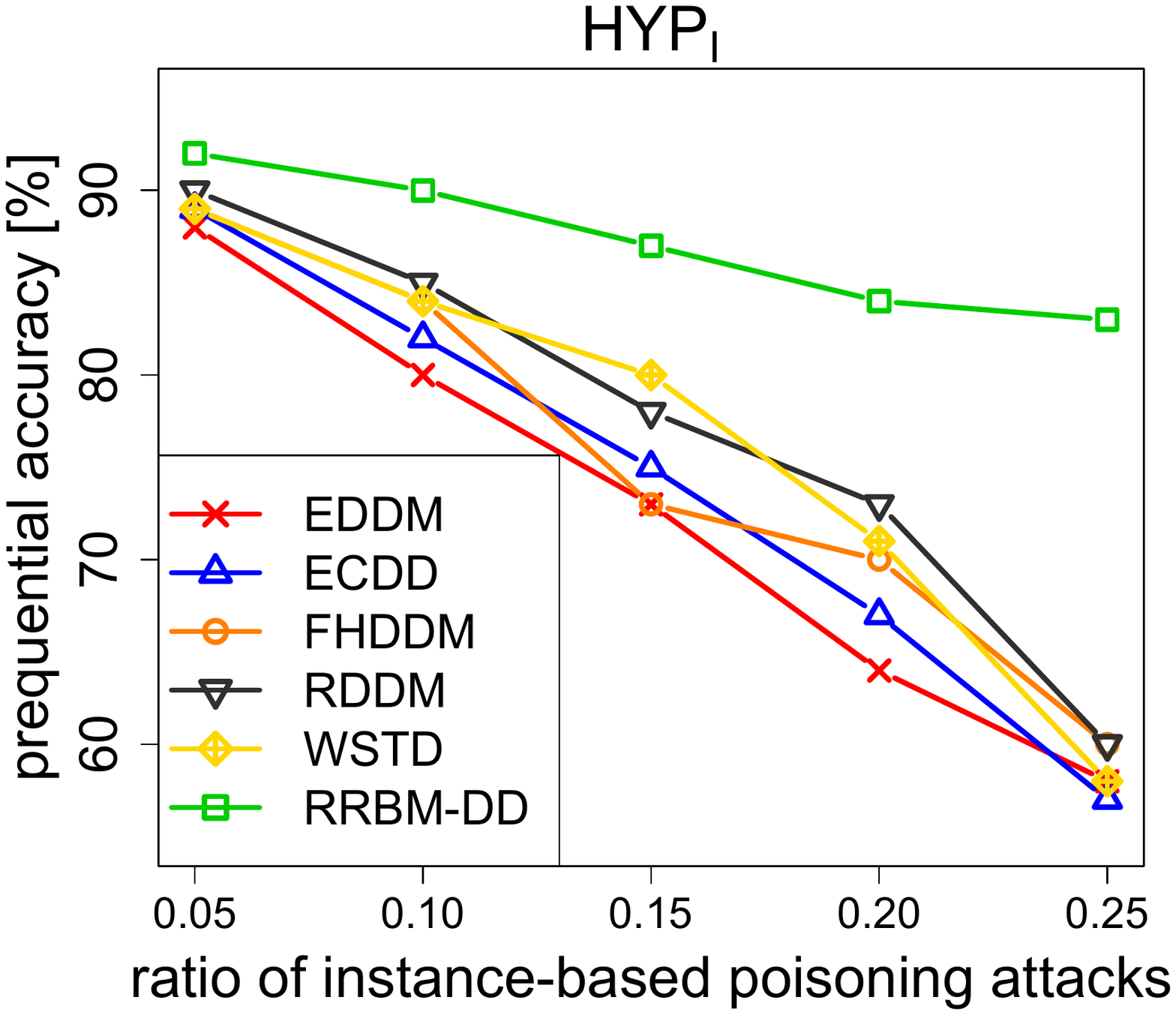}				    		   
			\includegraphics[width=0.3\linewidth,trim=2cm 2cm 2cm 2cm,clip]{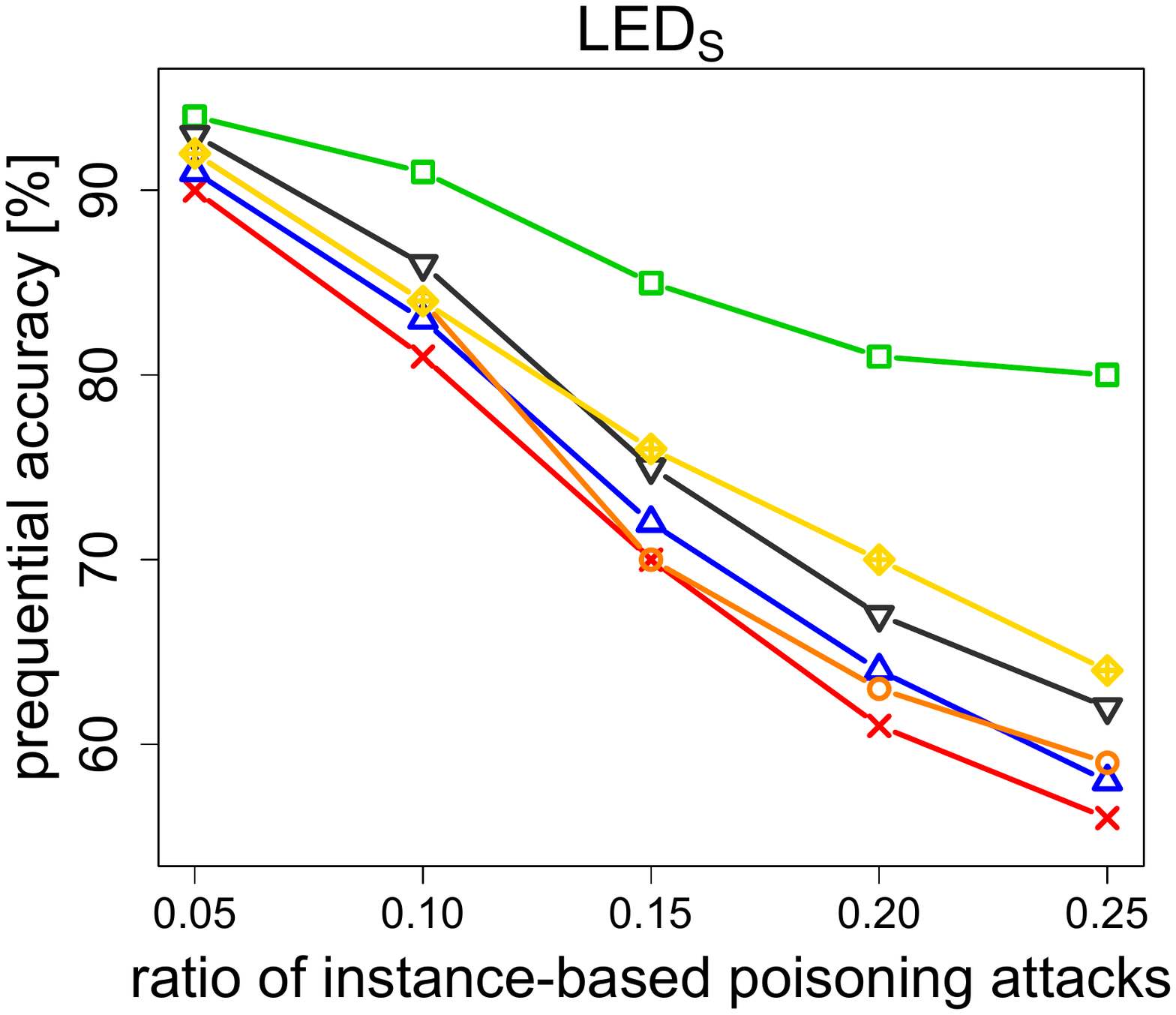}
			\includegraphics[width=0.3\linewidth,trim=2cm 2cm 2cm 2cm,clip]{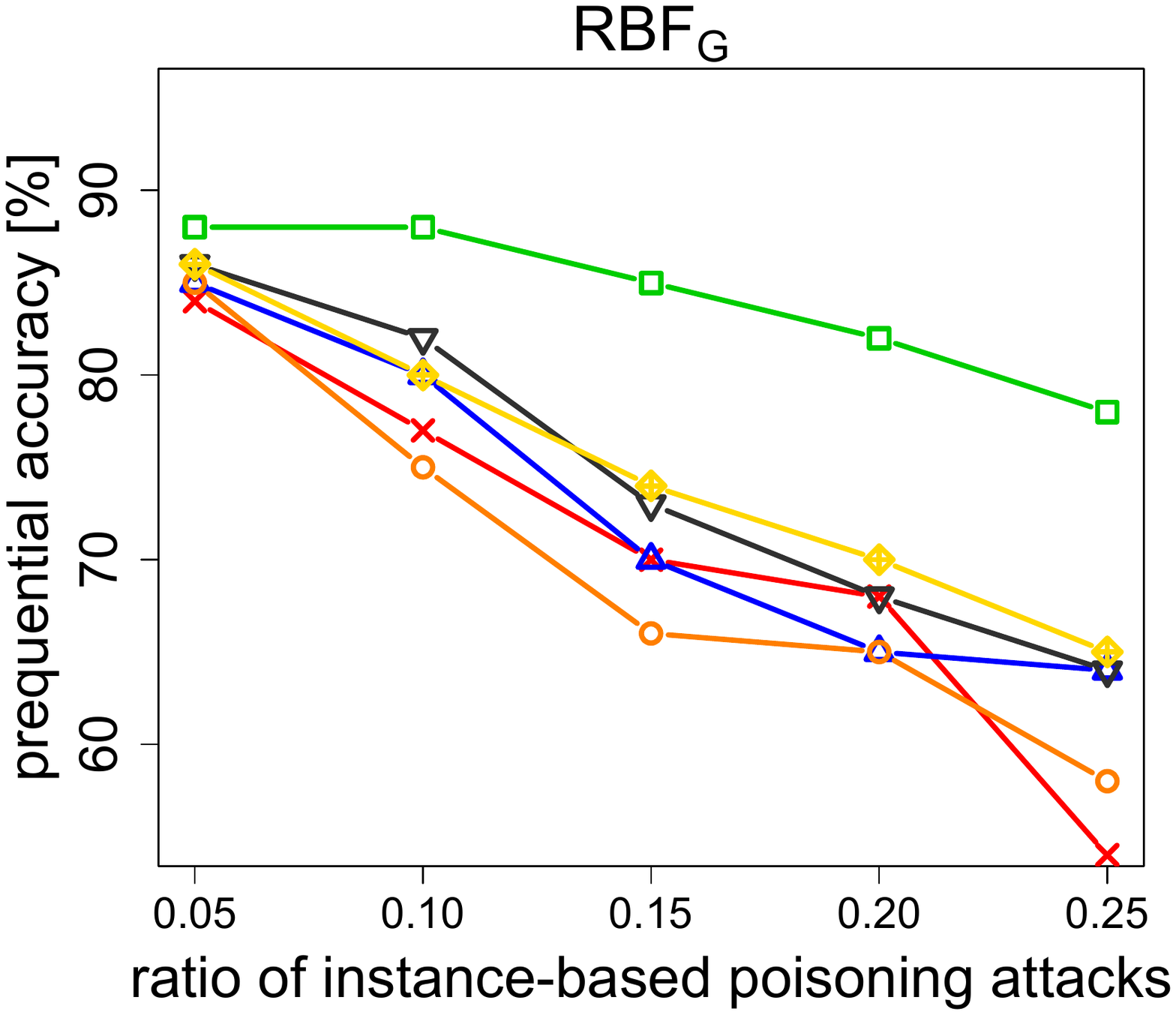}\vspace*{-1.5cm}
			\includegraphics[width=0.3\linewidth,trim=2cm 2cm 2cm 2cm,clip]{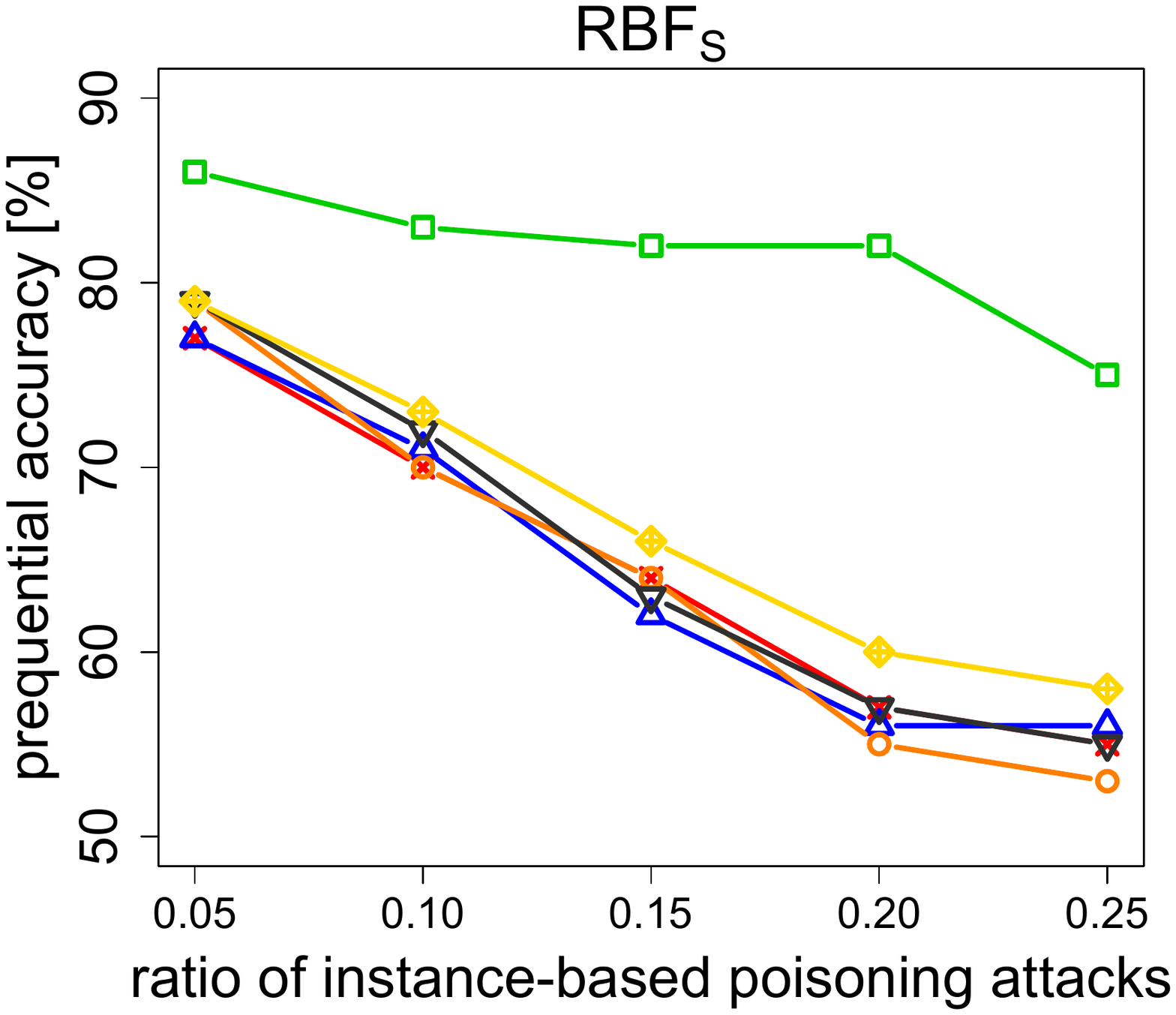}
			\includegraphics[width=0.3\linewidth,trim=2cm 2cm 2cm 2cm,clip]{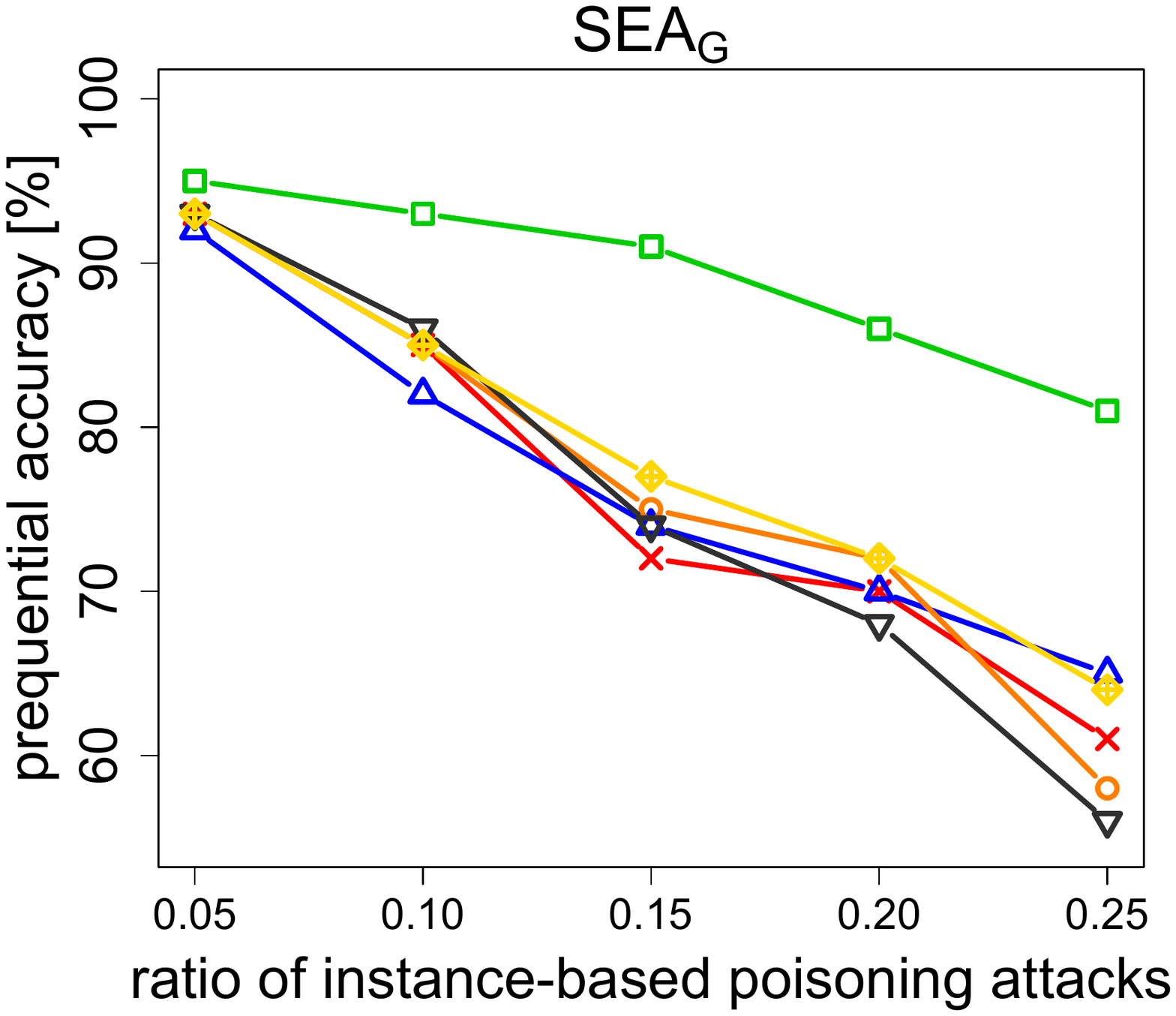}
			\includegraphics[width=0.3\linewidth,trim=2cm 2cm 2cm 2cm,clip]{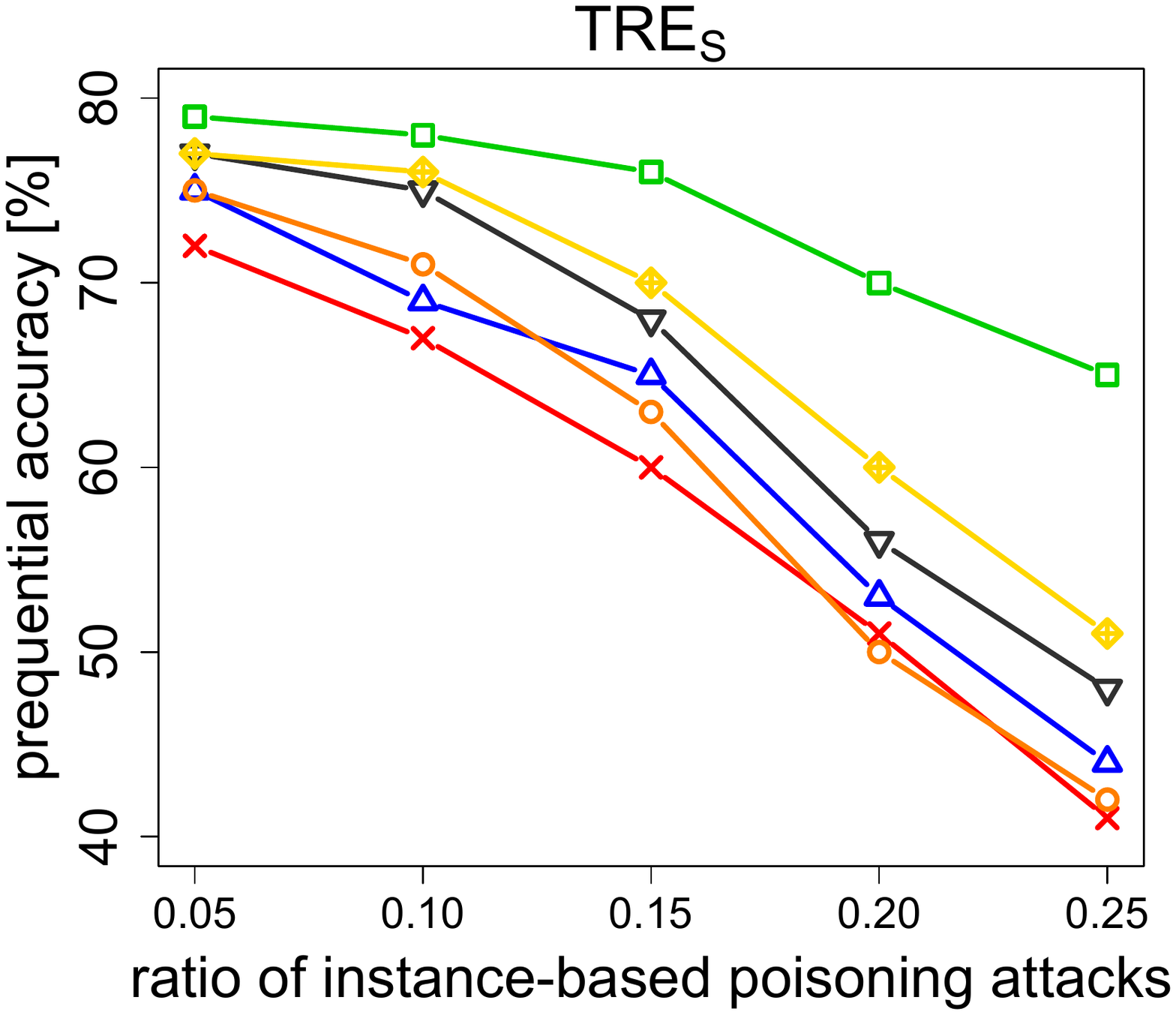}\vspace*{-1.5cm}
			\includegraphics[width=0.3\linewidth,trim=2cm 2cm 2cm 2cm,clip]{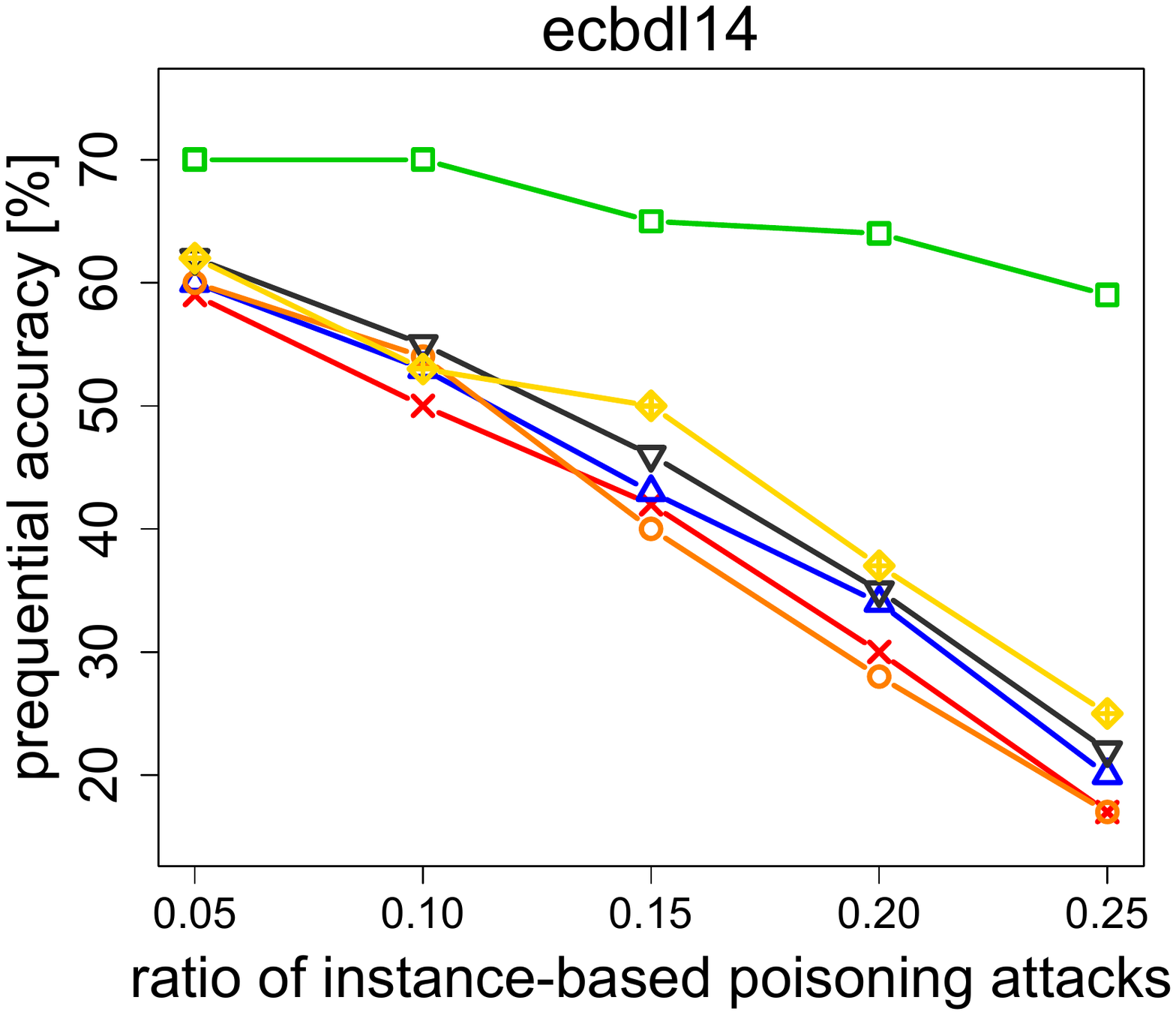}
			\includegraphics[width=0.3\linewidth,trim=2cm 2cm 2cm 2cm,clip]{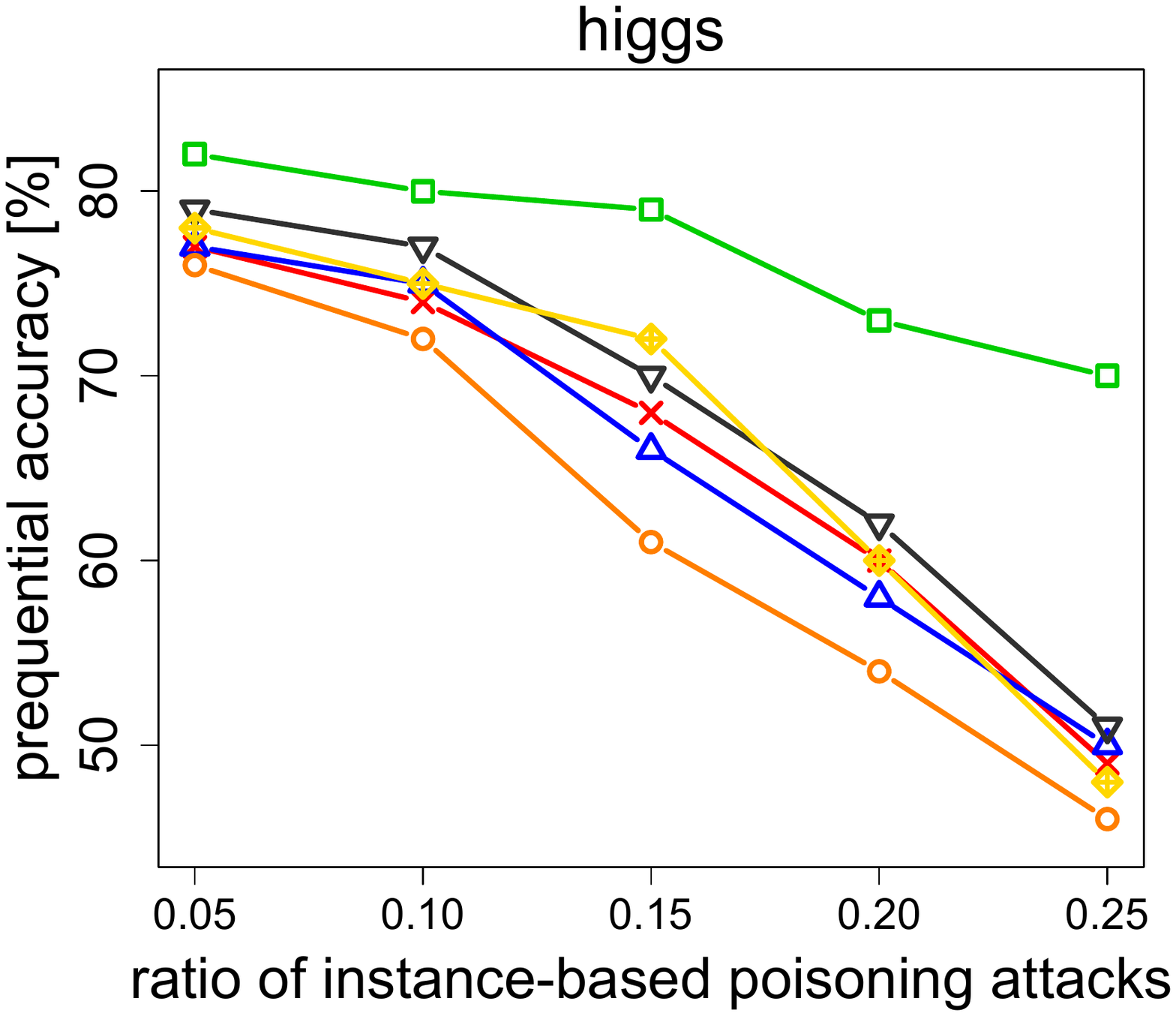}
			\includegraphics[width=0.3\linewidth,trim=2cm 2cm 2cm 2cm,clip]{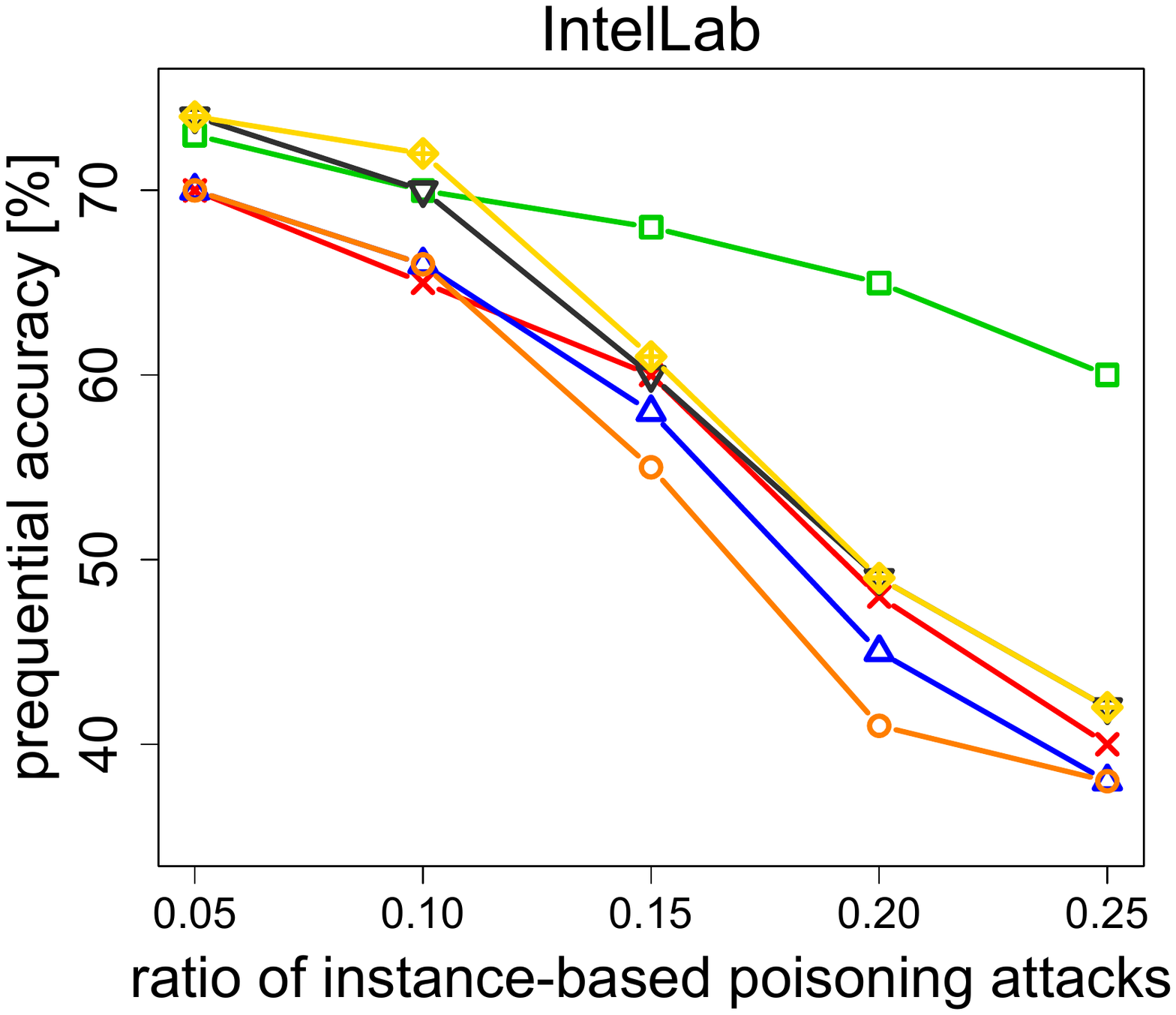}\vspace*{-1.5cm}
			\includegraphics[width=0.3\linewidth,trim=2cm 2cm 2cm 2cm,clip]{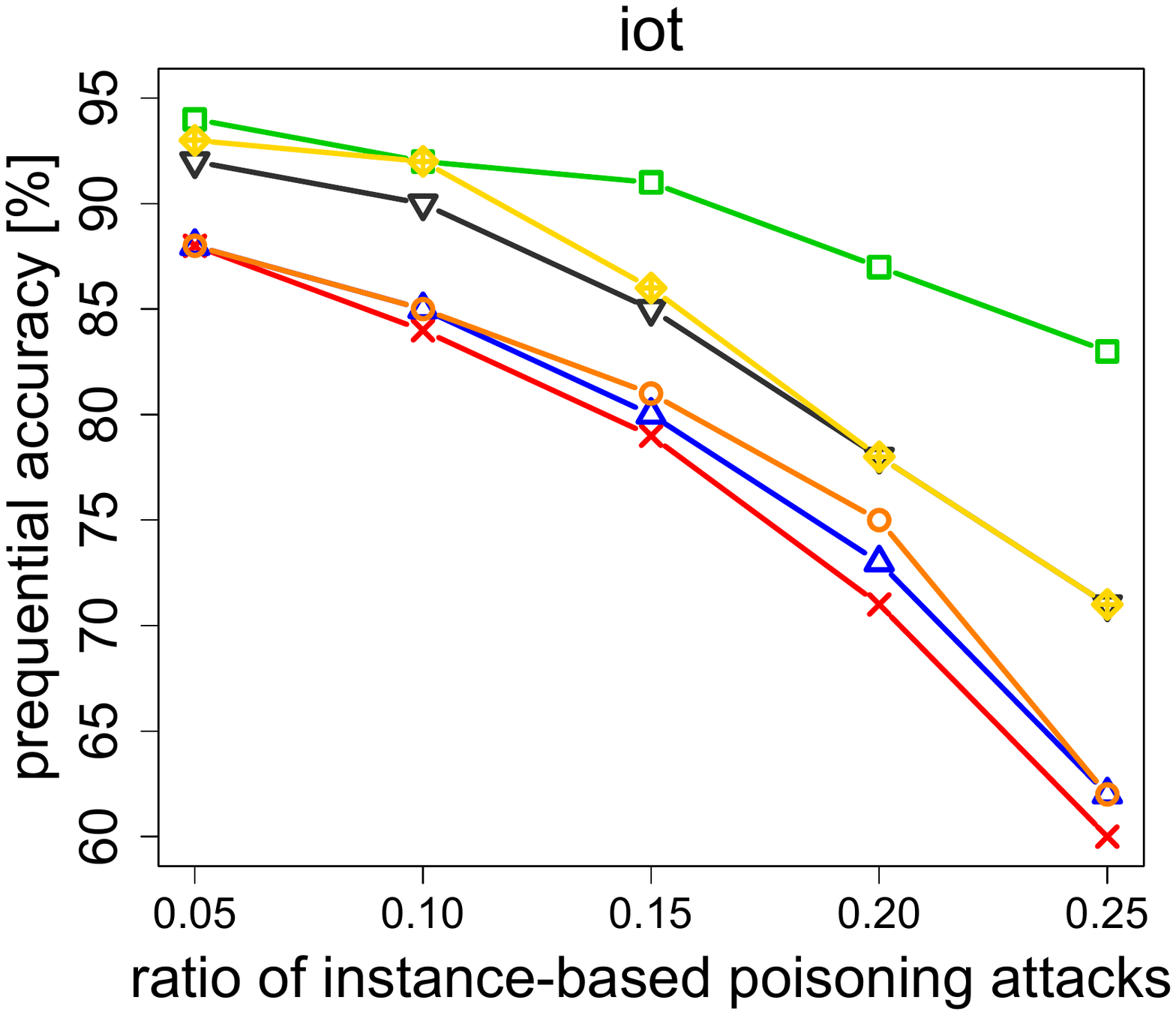}
			\includegraphics[width=0.3\linewidth,trim=2cm 2cm 2cm 2cm,clip]{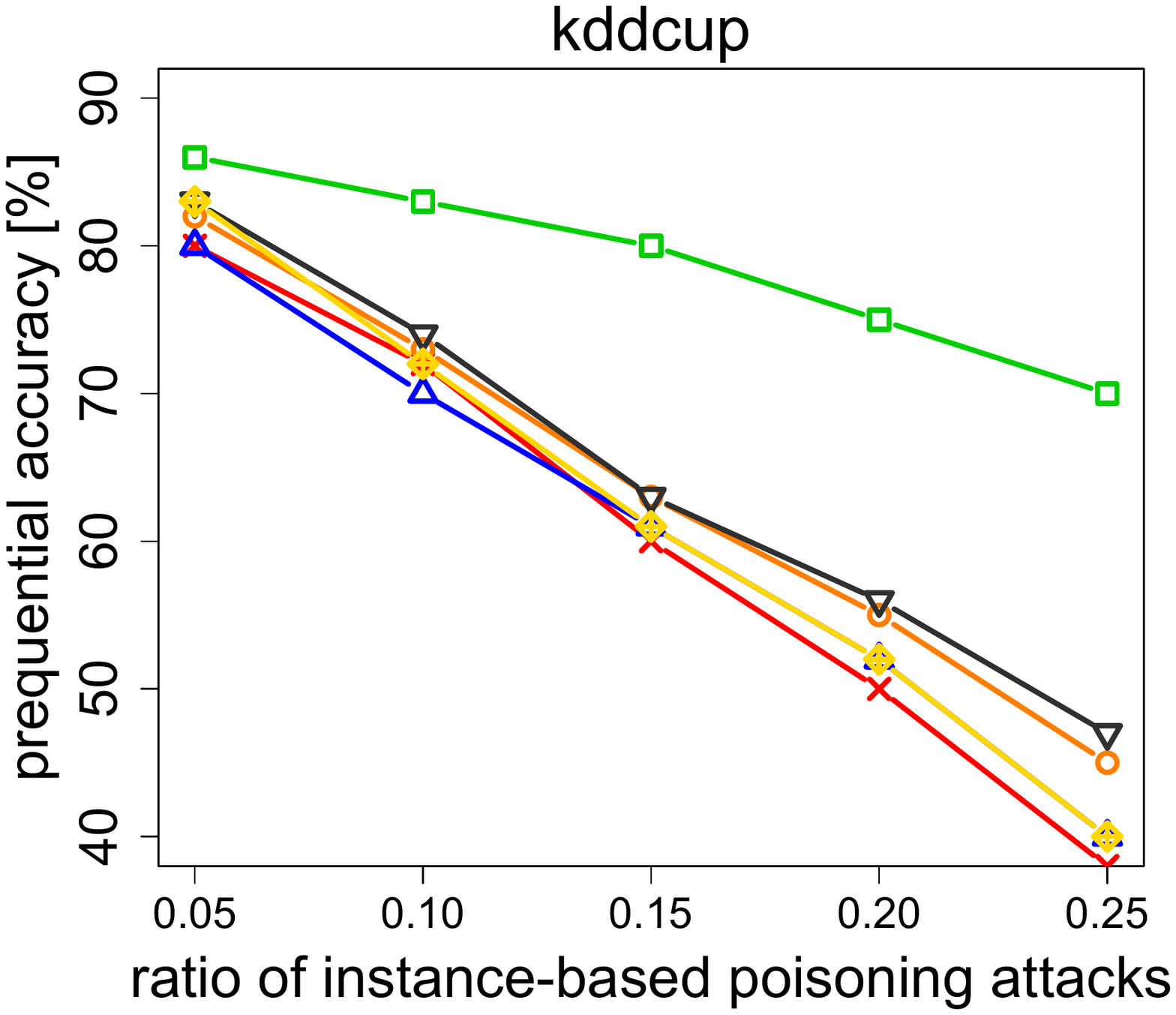}
			\includegraphics[width=0.3\linewidth,trim=2cm 2cm 2cm 2cm,clip]{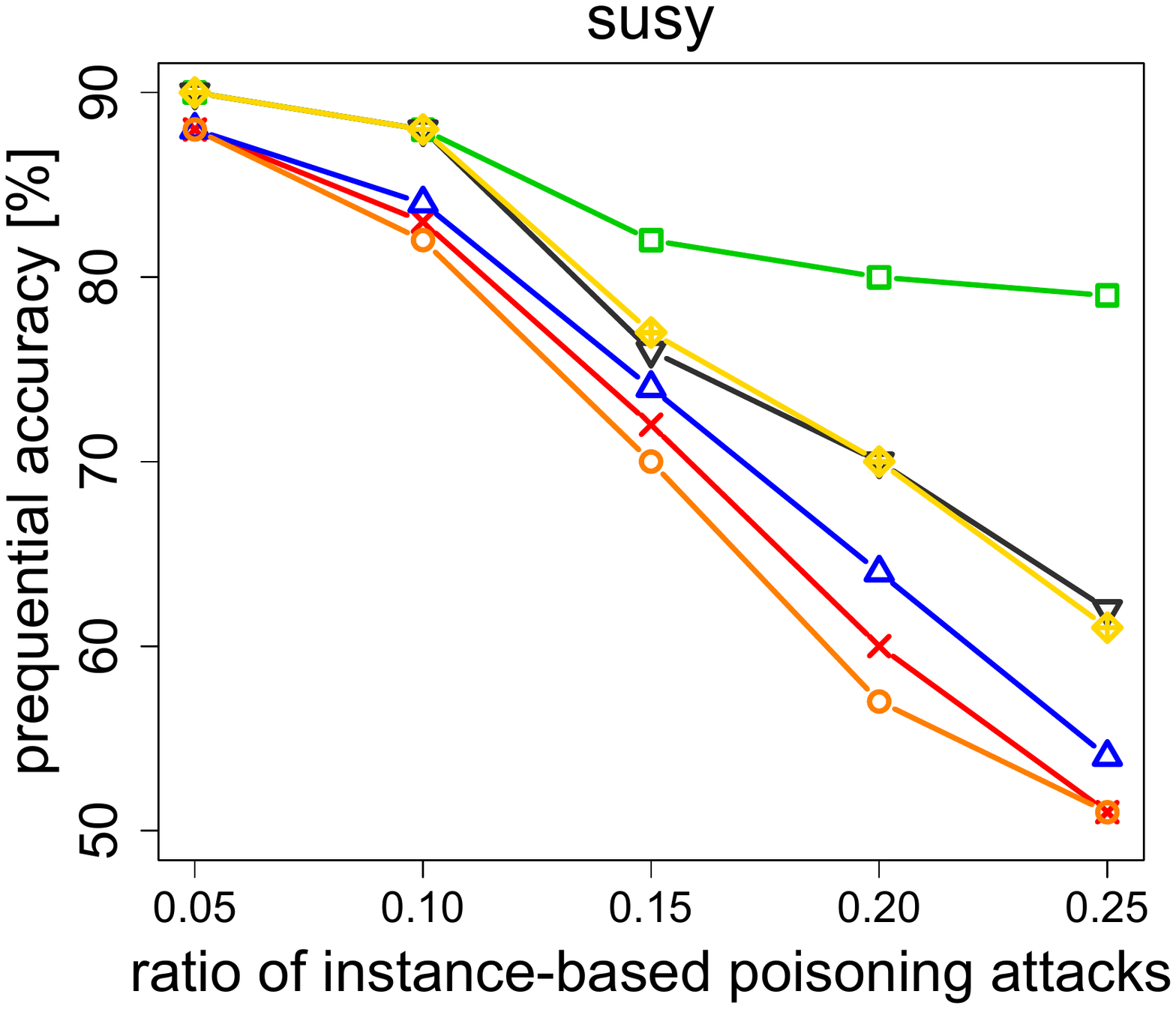}
			\caption{Relationship between prequential accuracy and ratio of injected adversarial concept drift via instance-based poisoning attacks.}
			\label{fig:exp1}
		\end{figure}

\begin{figure}[h]
	\centering
	\tiny
	\resizebox{\columnwidth}{!}{
		\begin{tikzpicture}
		\draw (1,0) -- (6,0);
		\foreach \x in {1,2,3,4,5,6} {
			\draw (\x, 0) -- ++(0,.1) node [below=0.15cm,scale=0.75] {\tiny \x};
			\ifthenelse{\x < 6}{\draw (\x+.5, 0) -- ++(0,.03);}{}
		}
		\coordinate (c0) at (1.00,0);
		\coordinate (c1) at (3.05,0);
		\coordinate (c2) at (3.22,0);
		\coordinate (c3) at (3.73,0);
		\coordinate (c4) at (4.12,0);
		\coordinate (c5) at (5.88,0);
		
		\node (l0) at (c0) [above right=.15cm and 0.1cm, align=center, scale=0.7] {\tiny RRBM-DD};
		\node (l1) at (c1) [above left=.15cm and 0.1cm, align=center, scale=0.7] {\tiny WSTD};
		\node (l2) at (c2) [above right=.45cm and 0.1cm, align=left, scale=0.7] {\tiny RDDM};
		\node (l3) at (c3) [above right=.30cm and 0.1cm, align=center, scale=0.7] {\tiny FHDDM};
		\node (l4) at (c4) [above right=.15cm and 0.1cm, align=center, scale=0.7] {\tiny ECDD};
		\node (l5) at (c5) [above left=.15cm and 0.1cm, align=left, scale=0.7] {\tiny EDDM};
		
		\fill[fill=gray,fill opacity=0.5] (1.00,-0.08) rectangle (2.05,0.08);
		
		\foreach \x in {0,...,5} {
			\draw (l\x) -| (c\x);
		};
		\end{tikzpicture}
	}
	\caption{The Bonferroni-Dunn test for comparison among drift detectors under instance-based poisoning attacks, based on RLR.}
	\label{fig:bon2}
\end{figure}
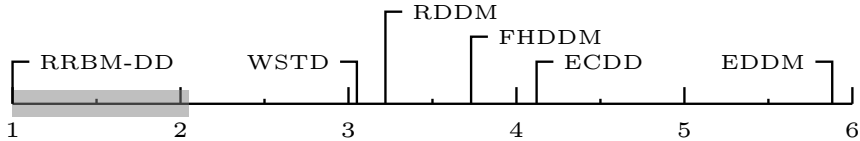

The first experiment was designed to evaluate the robustness of RRBM--DD and reference drift detectors to instance-based poisoning attacks. For this purpose, we injected five different ratios of poisoning attacks into 12 benchmark data streams. Figure~\ref{fig:exp1} depicts the effects of varying levels of adversarial concept drift on the prequential accuracy of the underlying classifier, while Table~\ref{tab:rlr1} presents the RLR metric results. Additionally, Figures~\ref{fig:bon1} and \ref{fig:bon2} show the visualizations of Friedman ranking test with Bonferroni-Dunn post-hoc on both used metrics. 

		\begin{table}[h]
		\centering
		\caption{RLR for RRBM--DD and reference drift detectors under instance-based poisoning attacks.}
		\begin{tabular}{lcccccc}
			\toprule
			Stream & EDDM & ECDD & FHDDM & RDDM & WSTD & RRBM--DD \\ 
			\midrule
			HYP$_{I}$ & 0.55 & 0.58 & 0.67 & 0.64 & 0.63 & 0.85 \\
			LED$_S$ &  0.61 & 0.62 & 0.71 & 0.71 & 0.74 & 0.90 \\
			RBF$_G$ &  0.54 & 0.55 & 0.58 & 0.56 & 0.61 & 0.77 \\
			RBF$_S$ &  0.49 & 0.47 & 0.50 & 0.54 & 0.52 & 0.73 \\
			SEA$_G$ &  0.67 & 0.71 & 0.70 & 0.74 & 0.77 & 0.86 \\
			TRE$_S$ &  0.44 & 0.43 & 0.48 & 0.49 & 0.51 & 0.72 \\
			ecbdl14&  0.40 & 0.45 & 0.52 & 0.56 & 0.53 & 0.69 \\
			higgs & 0.71 & 0.73 & 0.77 & 0.81 & 0.82 & 0.94 \\
			IntelLab&  0.41 & 0.44 & 0.46 & 0.48 & 0.49 & 0.69 \\
			iot&  0.73 & 0.77 & 0.81 & 0.84 & 0.87 & 0.95 \\
			kddcup &  0.63 & 0.60 & 0.67 & 0.65 & 0.69 & 0.83 \\
			susy &  0.65 & 0.71 & 0.74 & 0.76 & 0.73 & 0.85 \\
			\midrule 
			avg. rank & 5.88 & 4.12 & 3.73 & 3.22 &3.05 & 1.00 \\
			\bottomrule
			\label{tab:rlr1}
		\end{tabular}
	\end{table}

\begin{figure}[h]
	\centering
	\tiny
	\resizebox{\columnwidth}{!}{
		\begin{tikzpicture}
		\draw (1,0) -- (6,0);
		\foreach \x in {1,2,3,4,5,6} {
			\draw (\x, 0) -- ++(0,.1) node [below=0.15cm,scale=0.75] {\tiny \x};
			\ifthenelse{\x < 6}{\draw (\x+.5, 0) -- ++(0,.03);}{}
		}
		\coordinate (c0) at (1.26,0);
		\coordinate (c1) at (3.45,0);
		\coordinate (c2) at (3.99,0);
		\coordinate (c3) at (4.02,0);
		\coordinate (c4) at (4.68,0);
		\coordinate (c5) at (5.28,0);
		
		\node (l0) at (c0) [above right=.15cm and 0.1cm, align=center, scale=0.7] {\tiny RRBM-DD};
		\node (l1) at (c1) [above left=.15cm and 0.1cm, align=center, scale=0.7] {\tiny WSTD};
		\node (l2) at (c2) [above left=.40cm and 0.1cm, align=left, scale=0.7] {\tiny RDDM};
		\node (l3) at (c3) [above right=.40cm and 0.1cm, align=center, scale=0.7] {\tiny FHDDM};
		\node (l4) at (c4) [above right=.25cm and 0.1cm, align=center, scale=0.7] {\tiny ECDD};
		\node (l5) at (c5) [above right=.15cm and 0.1cm, align=left, scale=0.7] {\tiny EDDM};
		
		\fill[fill=gray,fill opacity=0.5] (1.26,-0.08) rectangle (2.05,0.08);
		
		\foreach \x in {0,...,5} {
			\draw (l\x) -| (c\x);
		};
		\end{tikzpicture}
	}
	\caption{The Bonferroni-Dunn test for comparison among drift detectors under instance-based poisoning attacks, based on prequential accuracy.}
	\label{fig:bon1}
\end{figure}

	We can see how instance-based poisoning attacks impact both drift detectors and the underlying classifier. For low attack ratios (0.05 and 0.10) we already can observe drops in performance, but all of the examined methods can still deliver acceptable performance. With the increase of poisoning attack ratios, all reference drift detectors start to fail. They cannot cope with such adversarial streams and are not able to select properly the moment for updating the Adaptive Hoeffding Tree classifier. In many cases, the performance starts to approach random decisions, which is a strong indicator of the negative effects that instance-based poisoning attacks have on detectors. This is especially visible in the case of EDDM and ECDD that are the weakest performing ones. RRBM--DD outperforms every single competitor, especially for RLR metric where it is the best algorithm over all 12 benchmarks. By analyzing Figure~\ref{fig:exp1} we can see that RRBM--DD offers significantly higher robustness to increasing levels of adversarial concept drift. For not a single data stream we can observe any sharp decline in performance, even when 25\% of instances in the stream are corrupted by the adversarial attacker. Such a situation can be explained by the way standard drift detectors compute the presence of a valid drift. They use statistics, such as mean values or errors, derived directly from data. This makes them highly susceptible to any adversarial attack, as even small corruption of the data being used to compute drift statistics will result in either incorrectly increased sensitivity to any variations in streams, or inhibits sensitivity to valid concept drift. RRBM--DD avoids this pitfall by being a fully trainable drift detector that uses two robustness inducing mechanisms that filter data before it is updated. This allows RRBM--DD to compute more accurate reconstruction error from mini-batches and use it to correctly react only to valid concept drift. 
	
	\smallskip
	\noindent \textbf{RQ1 answer.} Yes, RRBM--DD offers excellent robustness to instance-based poisoning attacks and is not significantly affected by various levels of such adversarial concept drift. 
	
	\subsection{Experiment 2: Evaluating robustness to concept-based poisoning attacks}
	\label{sec:exp2}

\begin{figure}[h]
			\centering
			\includegraphics[width=0.3\linewidth,trim=2cm 2cm 2cm 2cm,clip]{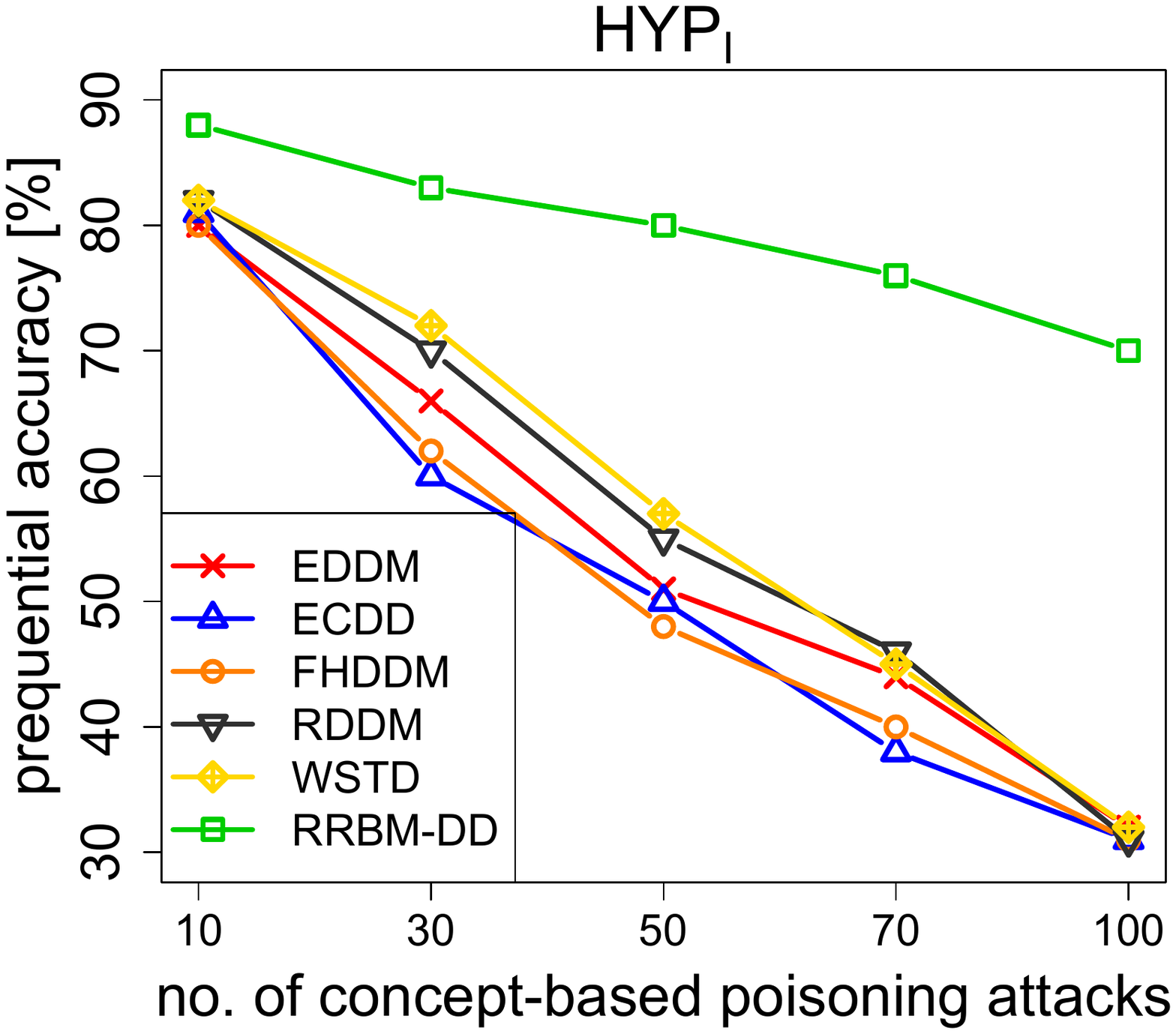}				    		   \includegraphics[width=0.3\linewidth,trim=2cm 2cm 2cm 2cm,clip]{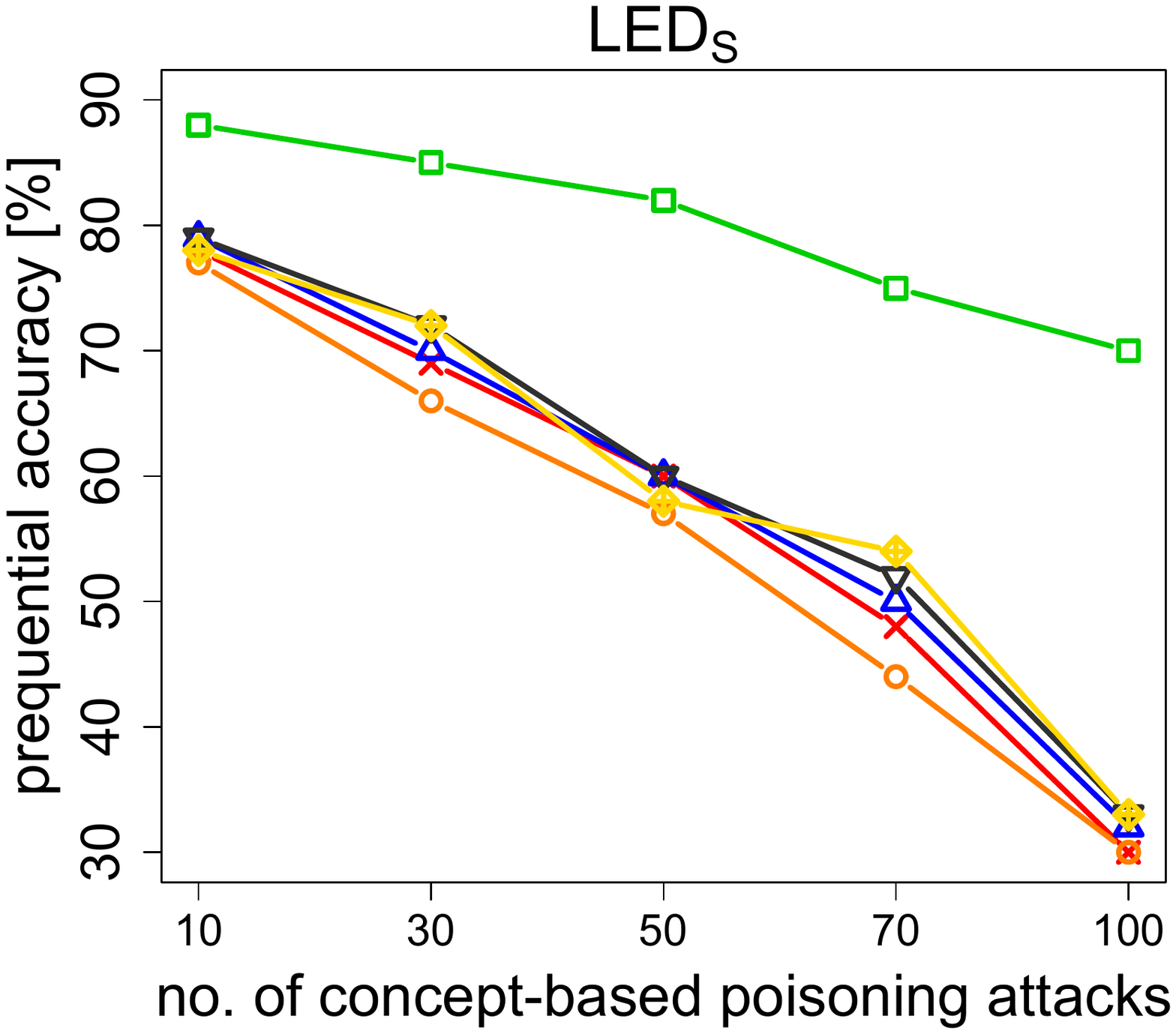}
			\includegraphics[width=0.3\linewidth,trim=2cm 2cm 2cm 2cm,clip]{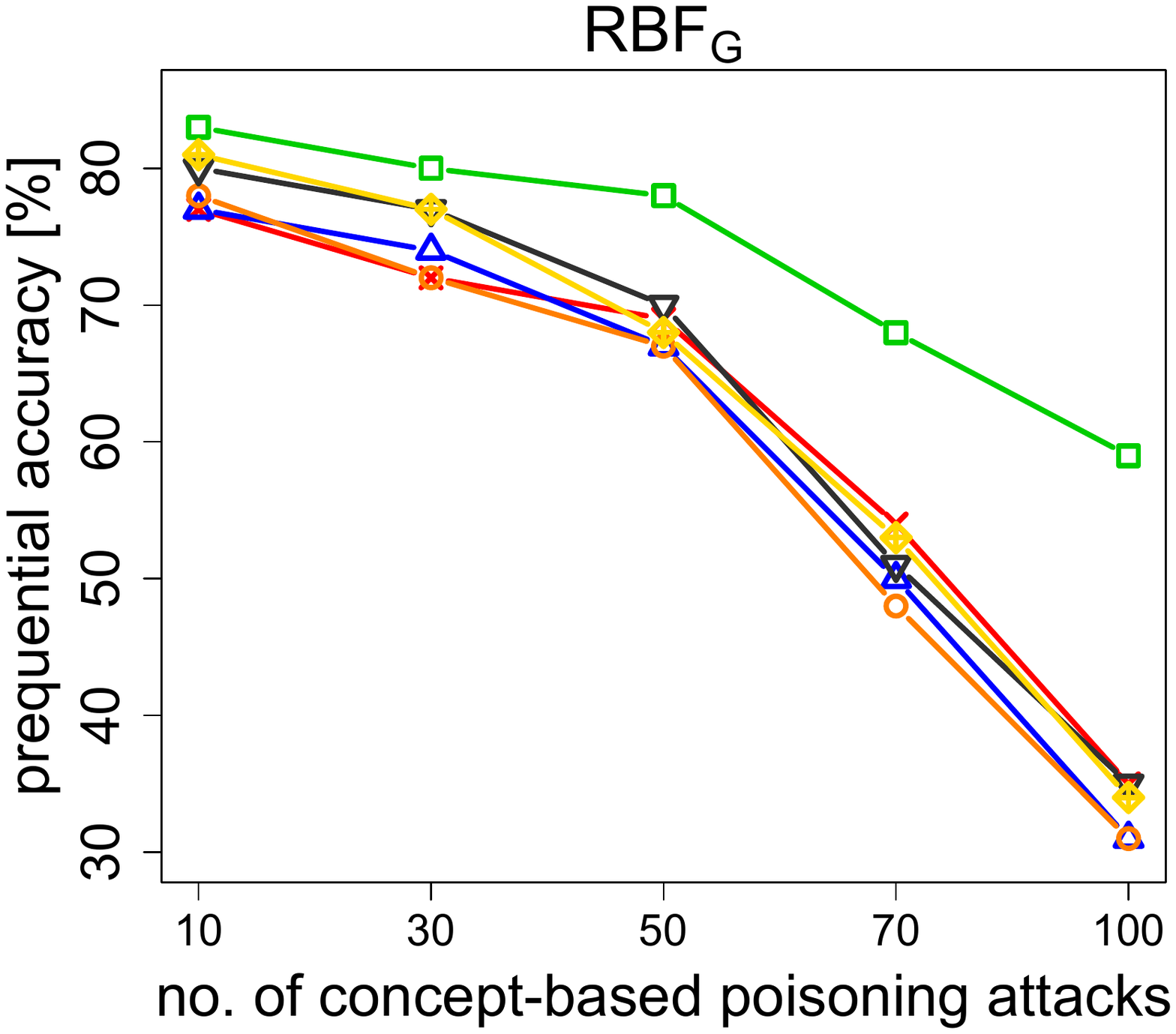}\vspace*{-1.5cm}
			\includegraphics[width=0.3\linewidth,trim=2cm 2cm 2cm 2cm,clip]{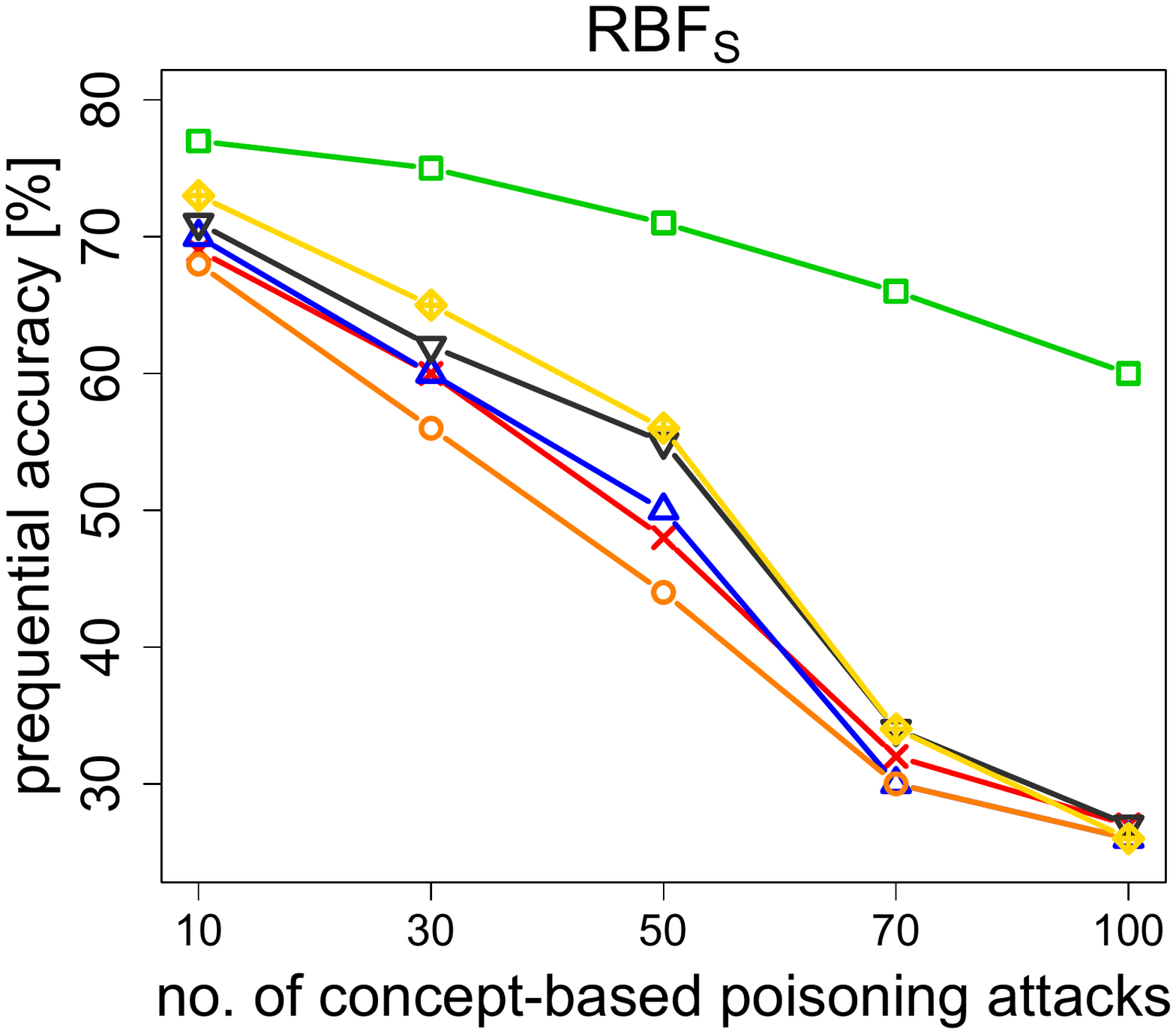}
			\includegraphics[width=0.3\linewidth,trim=2cm 2cm 2cm 2cm,clip]{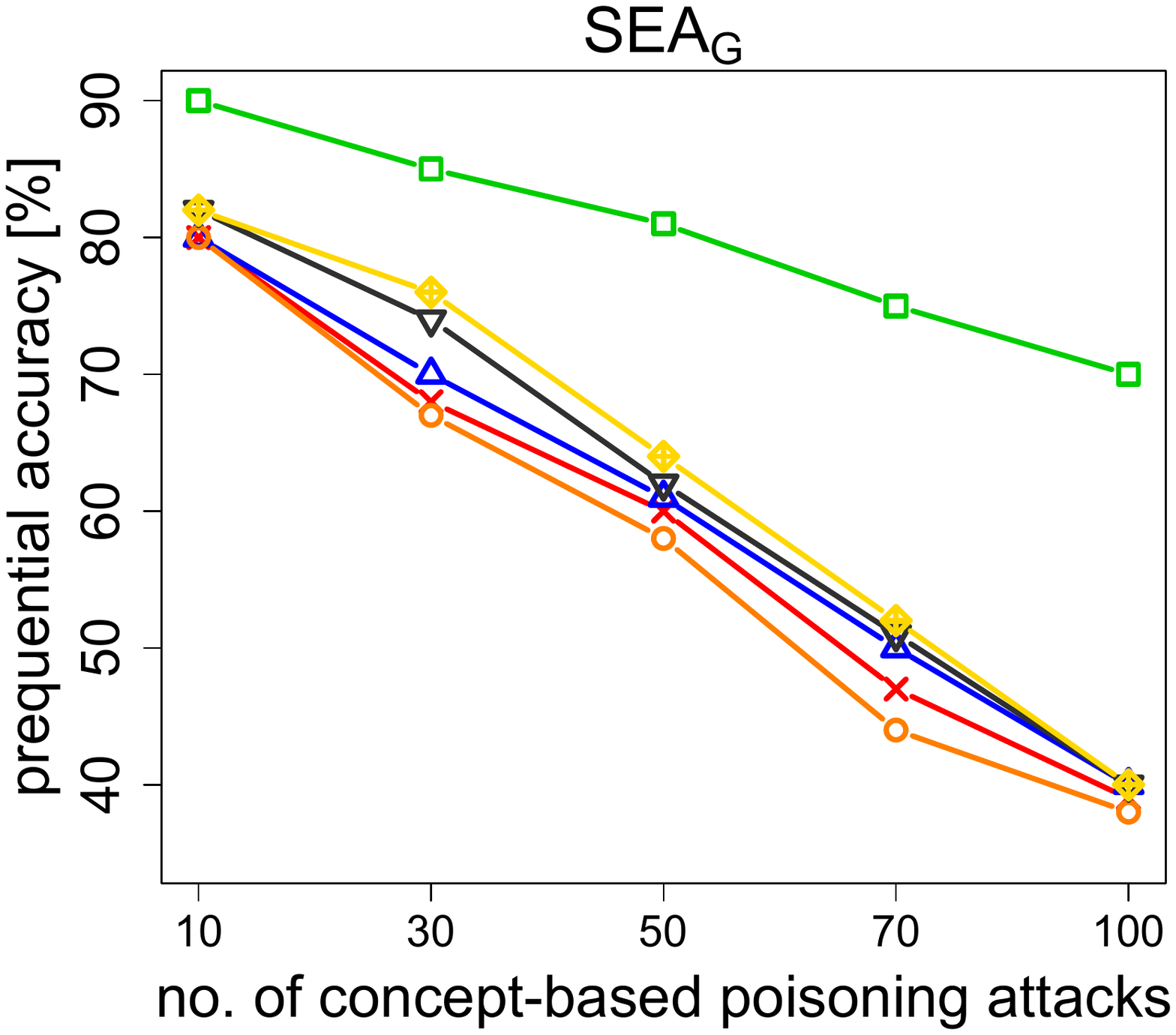}
			\includegraphics[width=0.3\linewidth,trim=2cm 2cm 2cm 2cm,clip]{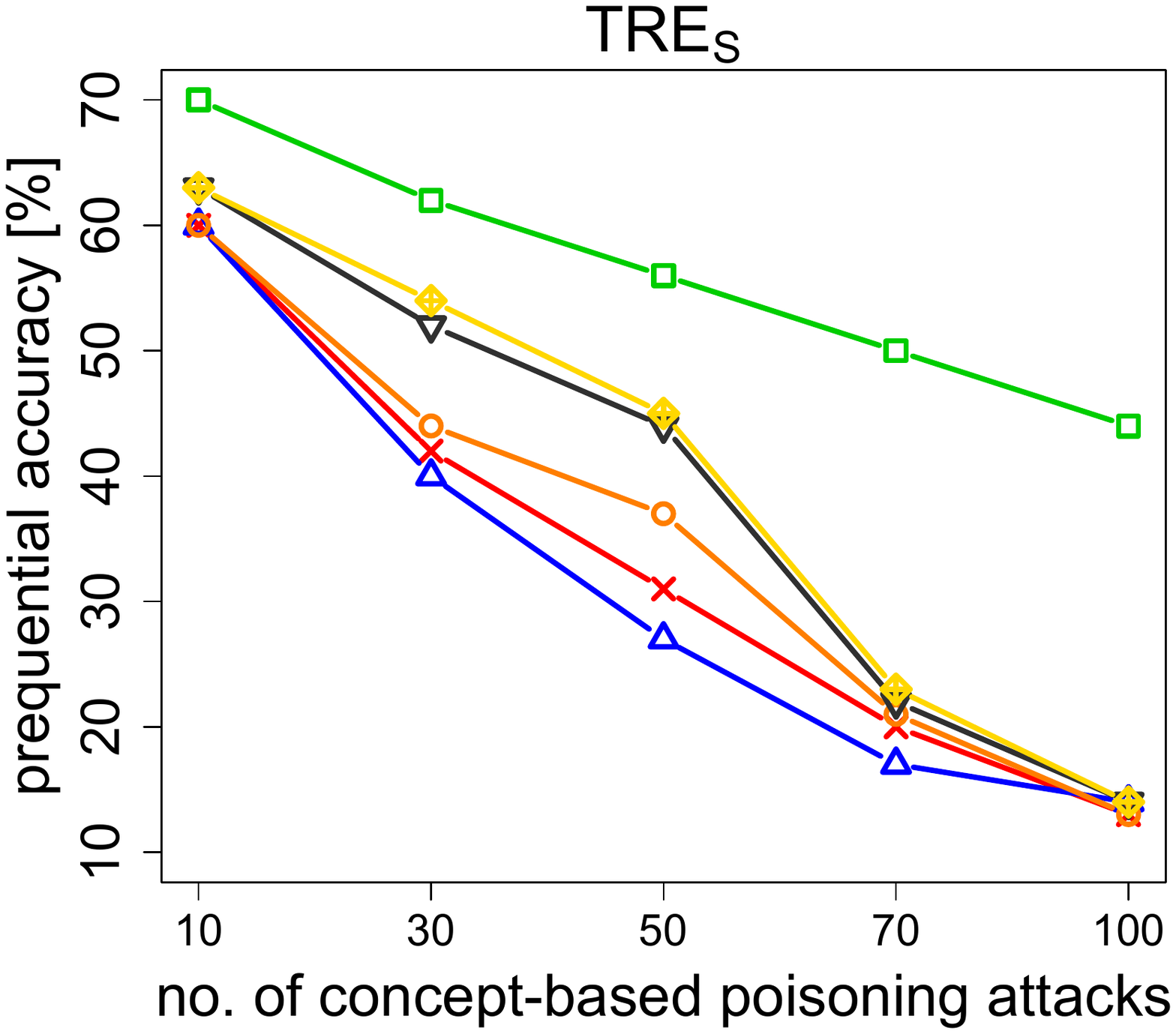}\vspace*{-1.5cm}
			\includegraphics[width=0.3\linewidth,trim=2cm 2cm 2cm 2cm,clip]{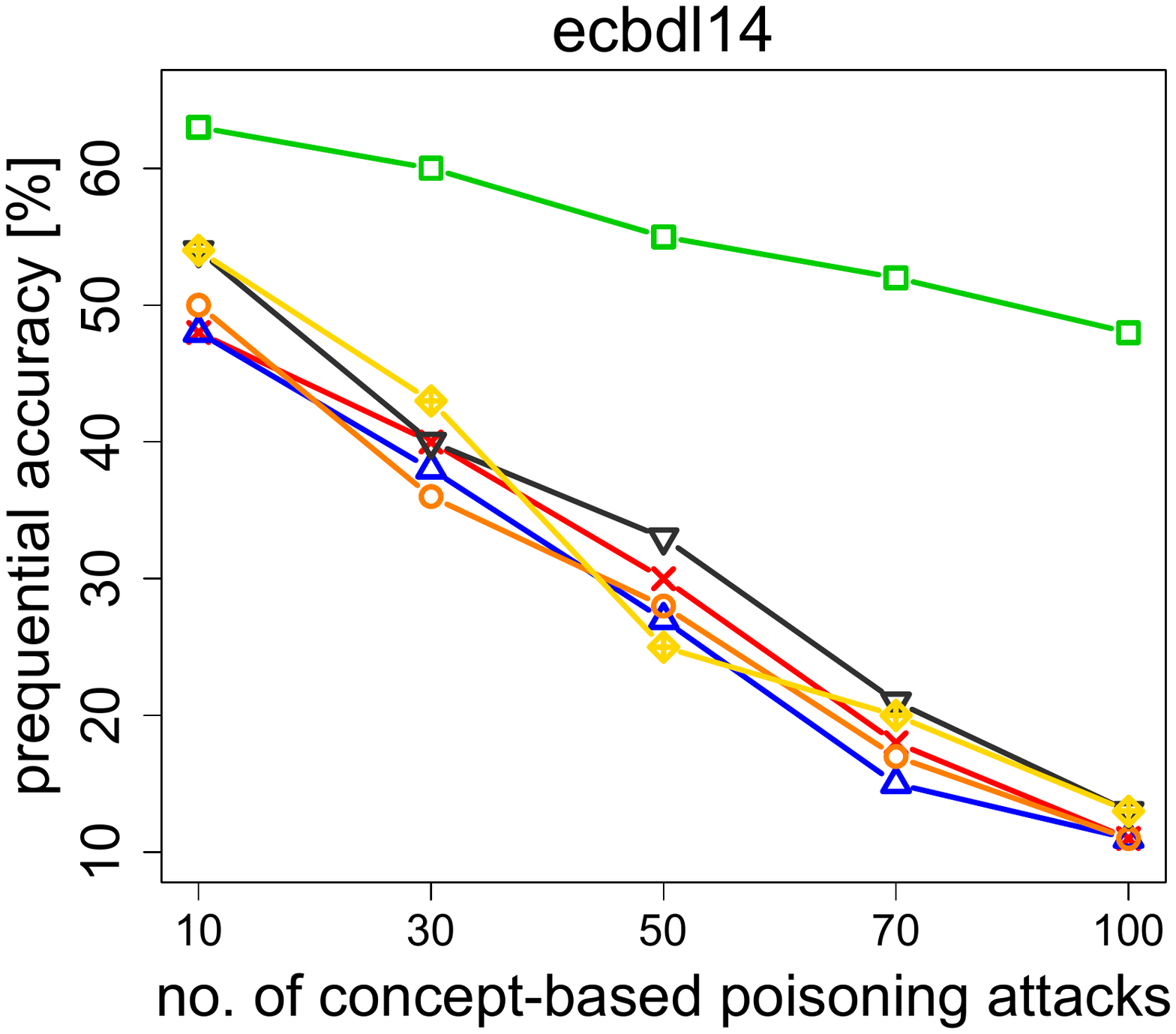}
			\includegraphics[width=0.3\linewidth,trim=2cm 2cm 2cm 2cm,clip]{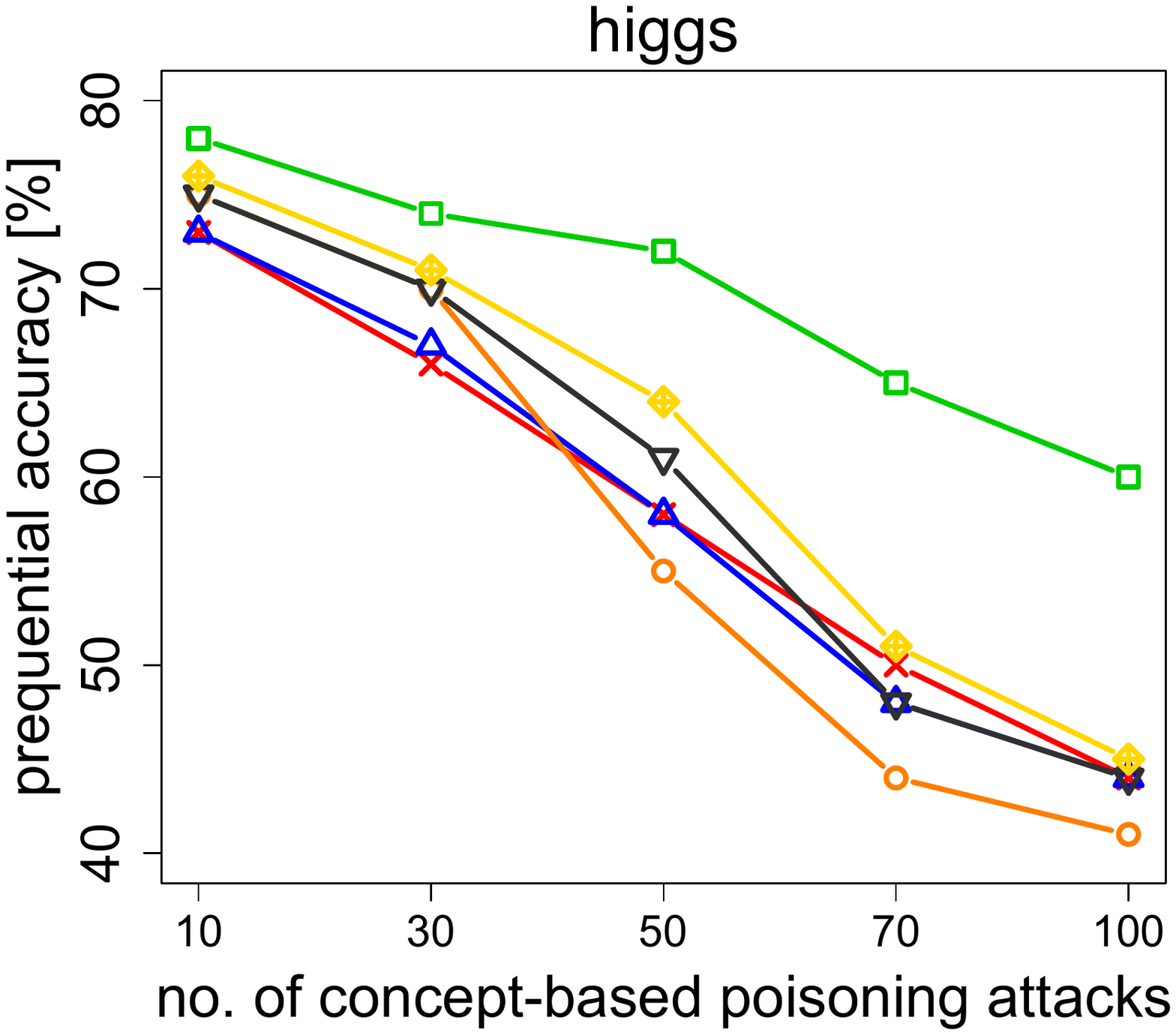}
			\includegraphics[width=0.3\linewidth,trim=2cm 2cm 2cm 2cm,clip]{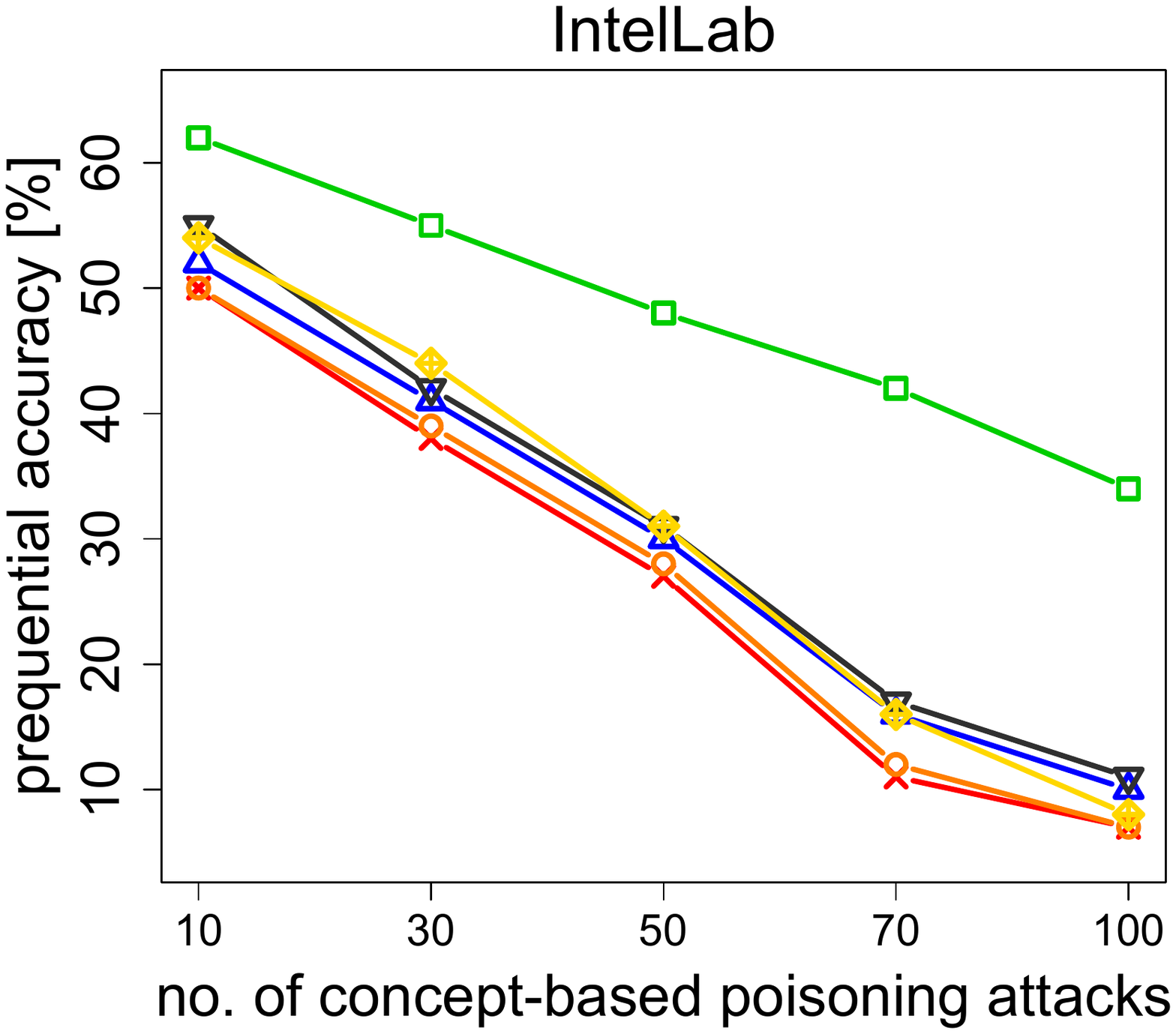}\vspace*{-1.5cm}
			\includegraphics[width=0.3\linewidth,trim=2cm 2cm 2cm 2cm,clip]{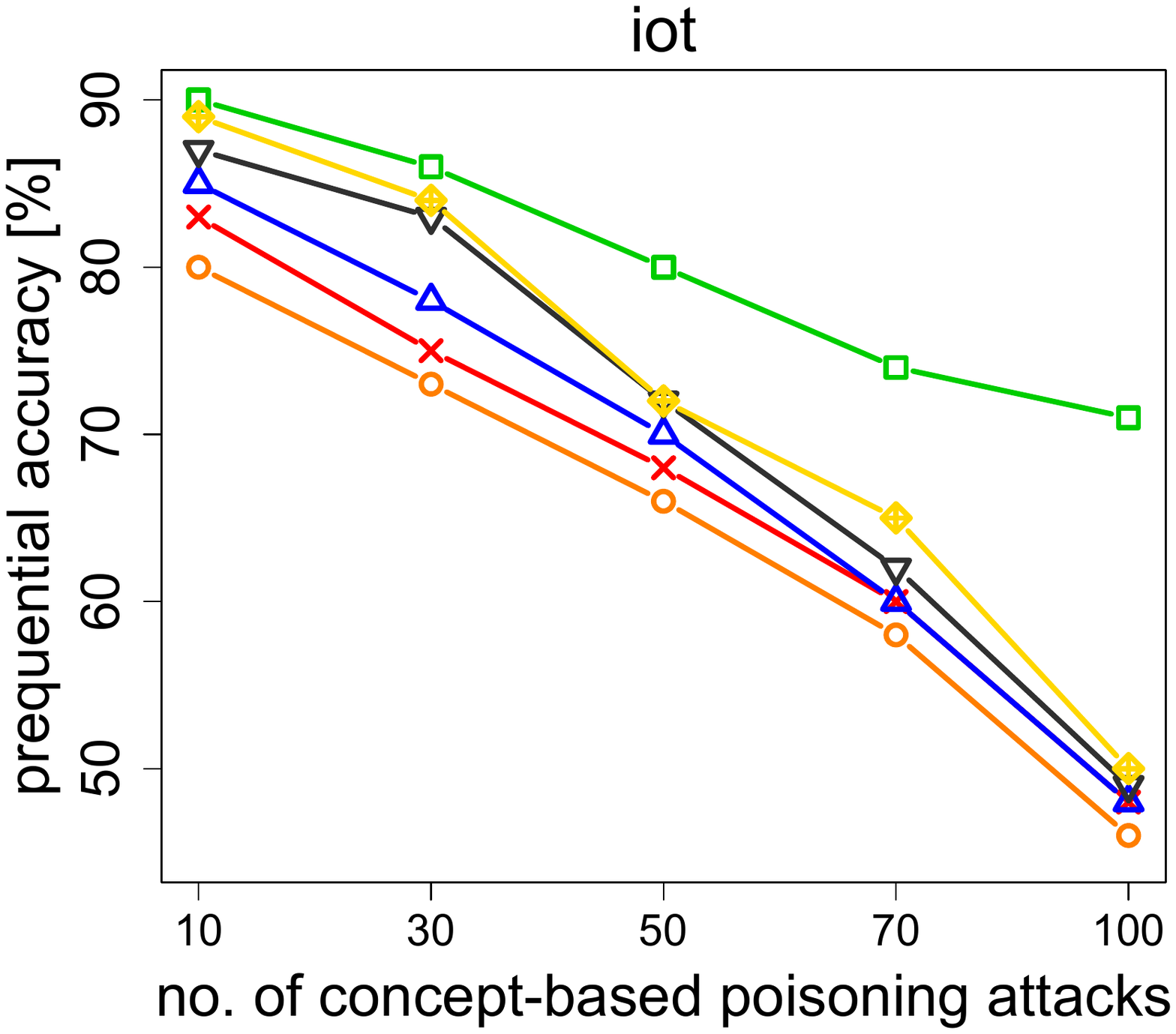}
			\includegraphics[width=0.3\linewidth,trim=2cm 2cm 2cm 2cm,clip]{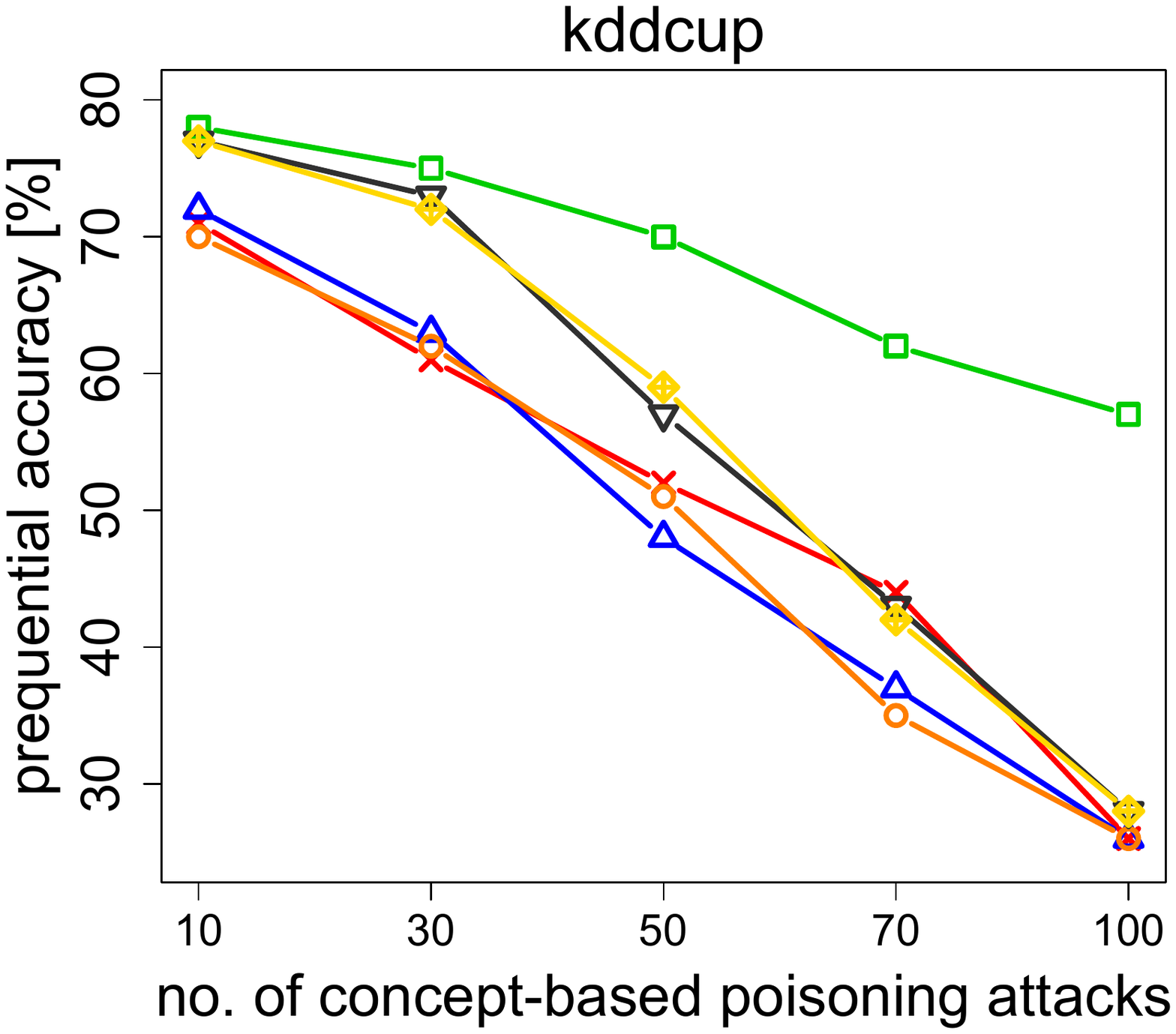}
			\includegraphics[width=0.3\linewidth,trim=2cm 2cm 2cm 2cm,clip]{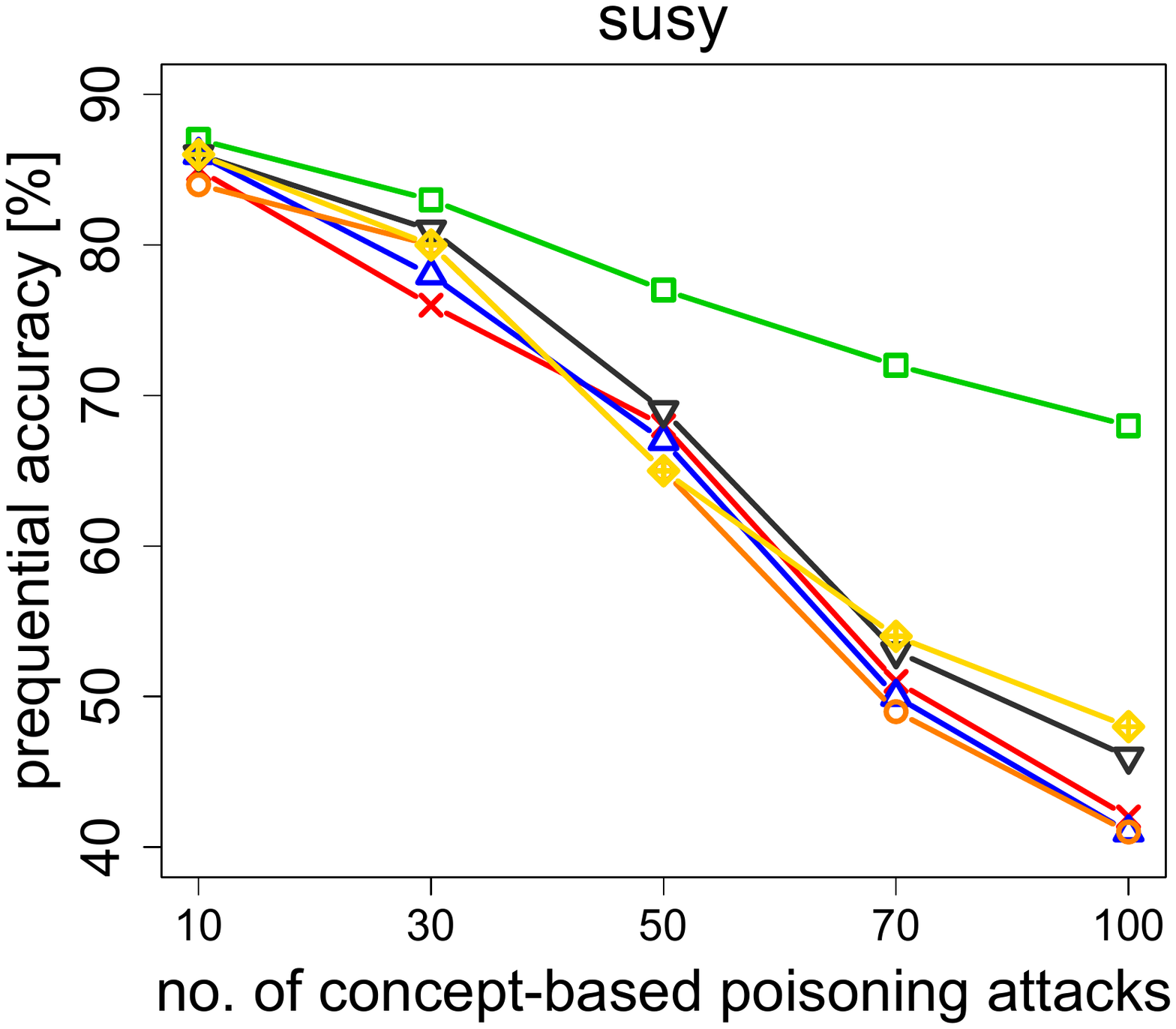}
			\caption{Relationship between prequential accuracy and number of injected adversarial concept drift via concept-based poisoning attacks.}
			\label{fig:exp2}
		\end{figure}

\begin{figure}[h]
	\centering
	\tiny
	\resizebox{\columnwidth}{!}{
		\begin{tikzpicture}
		\draw (1,0) -- (6,0);
		\foreach \x in {1,2,3,4,5,6} {
			\draw (\x, 0) -- ++(0,.1) node [below=0.15cm,scale=0.75] {\tiny \x};
			\ifthenelse{\x < 6}{\draw (\x+.5, 0) -- ++(0,.03);}{}
		}
		\coordinate (c0) at (1.10,0);
		\coordinate (c1) at (3.85,0);
		\coordinate (c2) at (4.50,0);
		\coordinate (c3) at (3.90,0);
		\coordinate (c4) at (4.85,0);
		\coordinate (c5) at (5.20,0);
		
		\node (l0) at (c0) [above right=.15cm and 0.1cm, align=center, scale=0.7] {\tiny RRBM-DD};
		\node (l1) at (c1) [above left=.15cm and 0.1cm, align=center, scale=0.7] {\tiny WSTD};
		\node (l2) at (c2) [above right=.40cm and 0.1cm, align=left, scale=0.7] {\tiny RDDM};
		\node (l3) at (c3) [above left=.40cm and 0.1cm, align=center, scale=0.7] {\tiny FHDDM};
		\node (l4) at (c4) [above right=.25cm and 0.1cm, align=center, scale=0.7] {\tiny ECDD};
		\node (l5) at (c5) [above right=.15cm and 0.1cm, align=left, scale=0.7] {\tiny EDDM};
		
		\fill[fill=gray,fill opacity=0.5] (1.10,-0.08) rectangle (2.55,0.08);
		
		\foreach \x in {0,...,5} {
			\draw (l\x) -| (c\x);
		};
		\end{tikzpicture}
	}
	\caption{The Bonferroni-Dunn test for comparison among drift detectors under concept-based poisoning attacks, based on prequential accuracy.}
	\label{fig:bon3}
\end{figure}
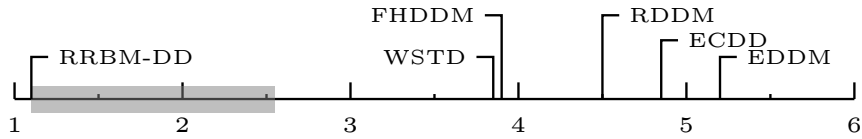

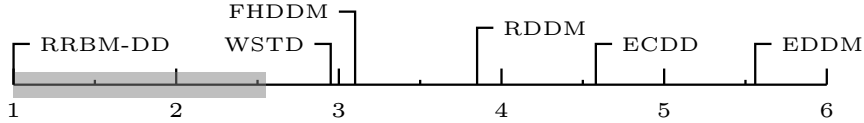
\begin{figure}[h]
	\centering
	\tiny
	\resizebox{\columnwidth}{!}{
		\begin{tikzpicture}
		\draw (1,0) -- (6,0);
		\foreach \x in {1,2,3,4,5,6} {
			\draw (\x, 0) -- ++(0,.1) node [below=0.15cm,scale=0.75] {\tiny \x};
			\ifthenelse{\x < 6}{\draw (\x+.5, 0) -- ++(0,.03);}{}
		}
		\coordinate (c0) at (1.00,0);
		\coordinate (c1) at (2.95,0);
		\coordinate (c2) at (3.85,0);
		\coordinate (c3) at (3.10,0);
		\coordinate (c4) at (4.58,0);
		\coordinate (c5) at (5.56,0);
		
		\node (l0) at (c0) [above right=.15cm and 0.1cm, align=center, scale=0.7] {\tiny RRBM-DD};
		\node (l1) at (c1) [above left=.15cm and 0.1cm, align=center, scale=0.7] {\tiny WSTD};
		\node (l2) at (c2) [above right=.25cm and 0.1cm, align=left, scale=0.7] {\tiny RDDM};
		\node (l3) at (c3) [above left=.35cm and 0.1cm, align=center, scale=0.7] {\tiny FHDDM};
		\node (l4) at (c4) [above right=.15cm and 0.1cm, align=center, scale=0.7] {\tiny ECDD};
		\node (l5) at (c5) [above right=.15cm and 0.1cm, align=left, scale=0.7] {\tiny EDDM};
		
		\fill[fill=gray,fill opacity=0.5] (1.00,-0.08) rectangle (2.55,0.08);
		
		\foreach \x in {0,...,5} {
			\draw (l\x) -| (c\x);
		};
		\end{tikzpicture}
	}
	\caption{The Bonferroni-Dunn test for comparison among drift detectors under concept-based poisoning attacks, based on RLR.}
	\label{fig:bon4}
\end{figure}

Experiment 2 was designed as a follow-up to Experiment 1. Here, we want to investigate a much more challenging scenario of concept-based poisoning attacks on data streams. For this purpose, we injected five different numbers of adversarial concepts into 12 benchmark data streams. Figure~\ref{fig:exp2} depicts the effects of varying levels of adversarial concept drift on the prequential accuracy of the underlying classifier, while Table~\ref{tab:rlr2} presents the RLR metric results. Additionally, Figures~\ref{fig:bon3} and \ref{fig:bon4} show the visualizations of Friedman ranking test with Bonferroni-Dunn post-hoc on both used metrics. 

			\begin{table}[h]
		\centering
		\caption{RLR for RRBM--DD and reference drift detectors under concept-based poisoning attacks.}
		\begin{tabular}{lcccccc}
			\toprule
			Stream & EDDM & ECDD & FHDDM & RDDM & WSTD & RRBM--DD \\ 
			\midrule
			HYP$_{I}$ & 0.44 & 0.45 & 0.50 & 0.52 & 0.51 & 0.78 \\
			LED$_S$ &  0.48 & 0.52 & 0.56 & 0.55 & 0.59 & 0.84 \\
			RBF$_G$ &  0.35 & 0.38 & 0.42 & 0.44 & 0.43 & 0.71 \\
			RBF$_S$ &  0.27 & 0.28 & 0.30 & 0.33 & 0.32 & 0.68 \\
			SEA$_G$ &  0.51 & 0.55 & 0.52 & 0.55 & 0.56 & 0.82 \\
			TRE$_S$ &  0.22 & 0.20 & 0.25 & 0.27 & 0.27 & 0.63 \\
			ecbdl14&  0.18 & 0.17 & 0.23 & 0.26 & 0.26 & 0.60 \\
			higgs & 0.52 & 0.56 & 0.57 & 0.61 & 0.63 & 0.88 \\
			IntelLab&  0.24 & 0.26 & 0.25 & 0.30 & 0.32 & 0.65 \\
			iot&  0.63 & 0.65 & 0.69 & 0.71 & 0.74 & 0.90 \\
			kddcup &  0.39 & 0.35 & 0.44 & 0.40 & 0.47 & 0.76 \\
			susy &  0.42 & 0.45 & 0.46 & 0.51 & 0.47 & 0.80 \\
			\midrule 
			avg. rank & 5.60 & 4.55 & 3.10 & 3.90 &2.95 & 1.00 \\
			\bottomrule
			\label{tab:rlr2}
		\end{tabular}
	\end{table}

Experiment 2 further confirms the observations made during the analysis of experiment 1. State-of-the-art drift detectors offer no robustness to adversarial attacks and can be very easily fooled or damaged by even a very small number of poisoning attacks. What is very important to notice is the difference in the impact of concept-based poisoning attacks versus their instance-based counterparts. Here we can see the catastrophic effects of injecting adversarial concepts into the stream, as every single reference drift detector converges at a point where it behaves worse than a random guess. This can be interpreted as adversarial data hijacking the updating process of the underlying classifier and forcing it to adapt to the malicious information. RRBM--DD once again offers excellent robustness, while maintaining a very good valid drift detection -- as evident from the high prequential accuracy of Adaptive Hoeffding Tree associated with it. Of course, concept-based poisoning attacks have a stronger effect on RRBM-DD, yet they are not capable of hindering its sensitivity and robustness. The first two experiments highlighted the excellent properties of RRBM-DD when dealing with adversarial concept drift, but it would be very beneficial to understand what exactly is the cause of such a performance. We will analyze this using an ablation study in experiment 4.

\smallskip
\noindent \textbf{RQ2 answer.} Yes, RRBM-DD can efficiently handle the injection of adversarial concepts, even when they fulfill smoothness and cluster assumptions. This proves that RRBM--DD is not simply insensitive to noisy data, but can handle truly adversarial scenarios with malicious party using crafted poisoning attacks on data.

	\subsection{Experiment 3: Evaluating robustness under class label sparsity}
	\label{sec:exp3}

\begin{figure}[h!]
			\centering
			\includegraphics[width=0.3\linewidth,trim=2cm 2cm 2cm 2cm,clip]{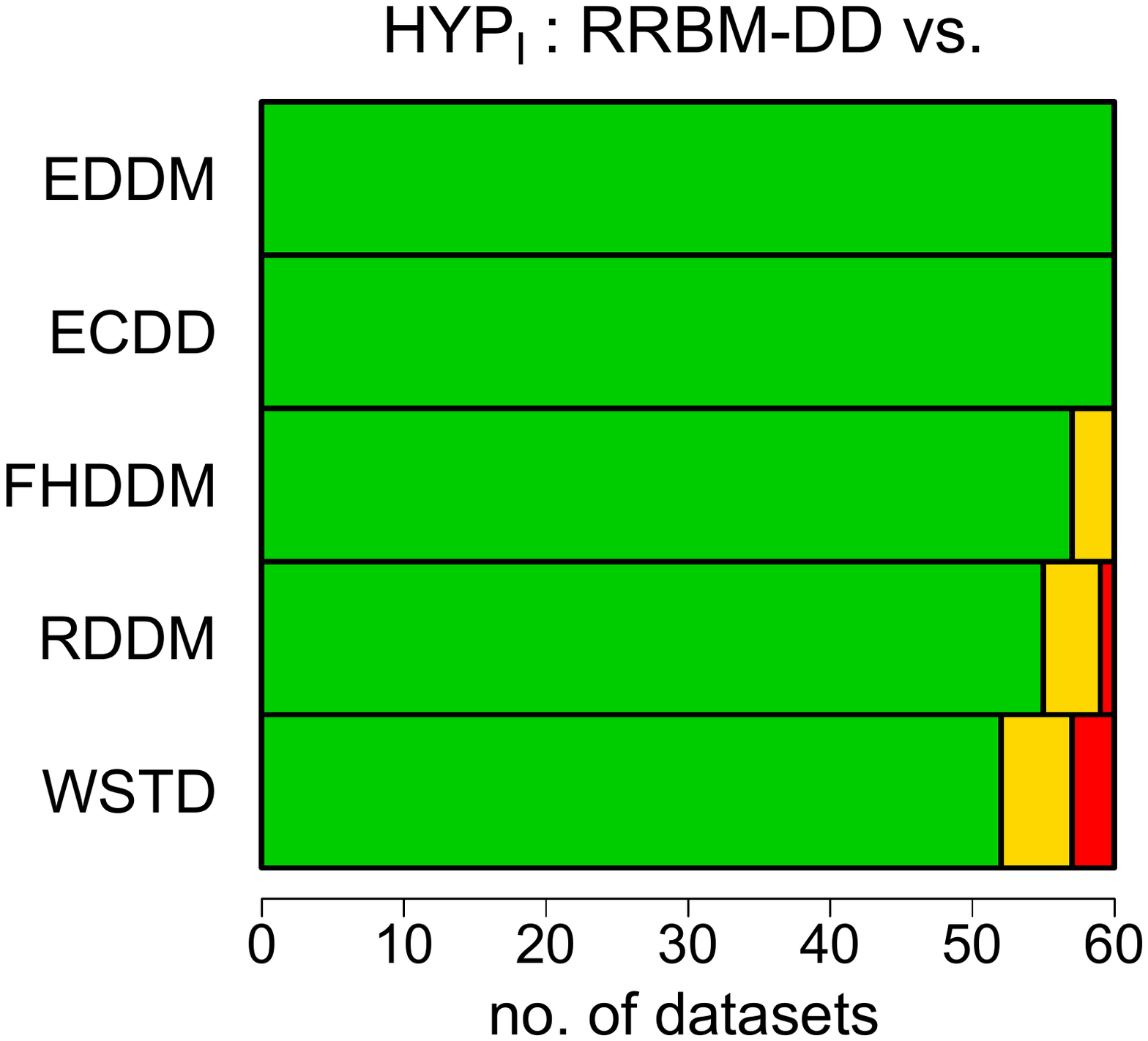}				    		   
			\includegraphics[width=0.3\linewidth,trim=2cm 2cm 2cm 2cm,clip]{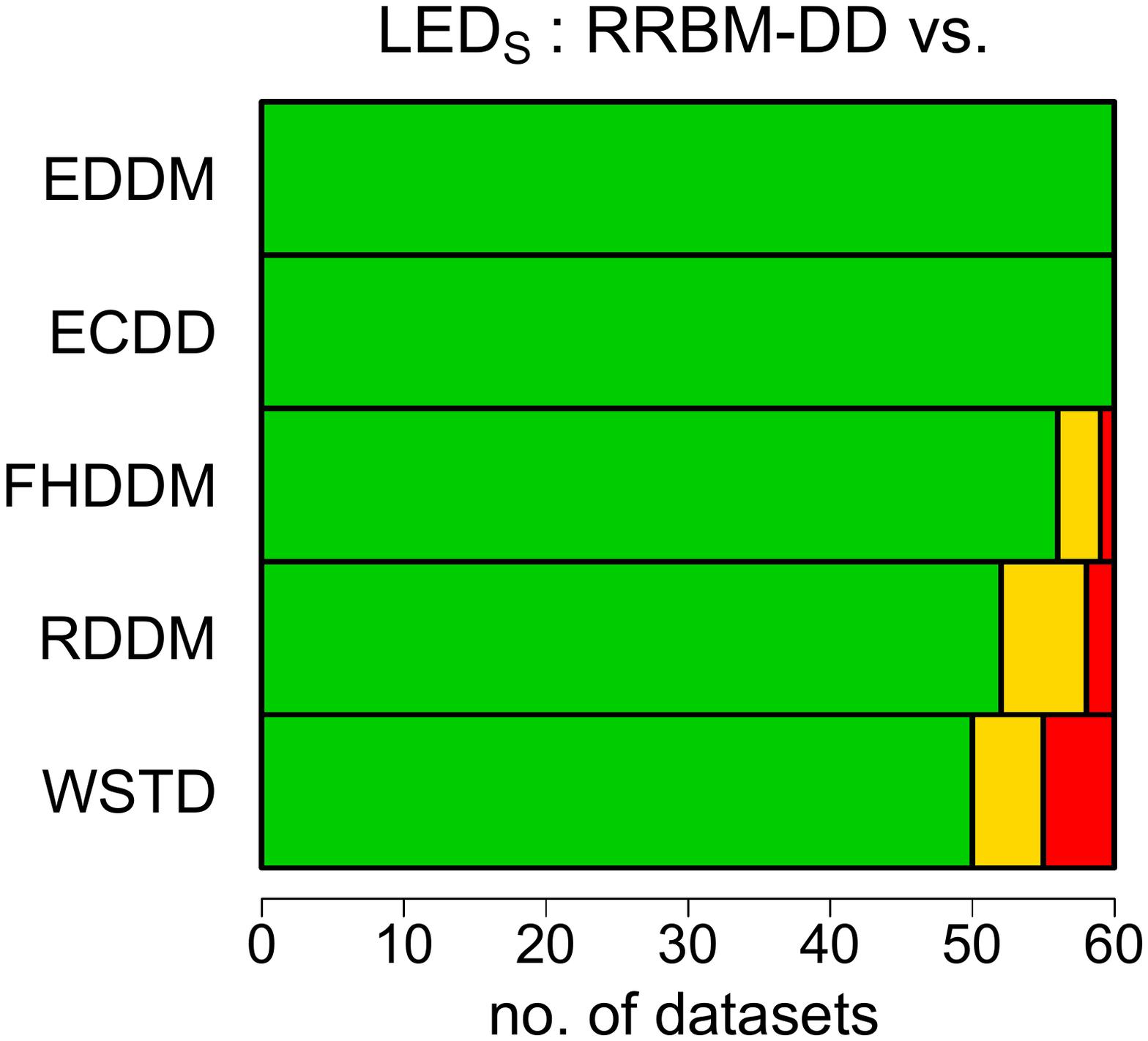}
			\includegraphics[width=0.3\linewidth,trim=2cm 2cm 2cm 2cm,clip]{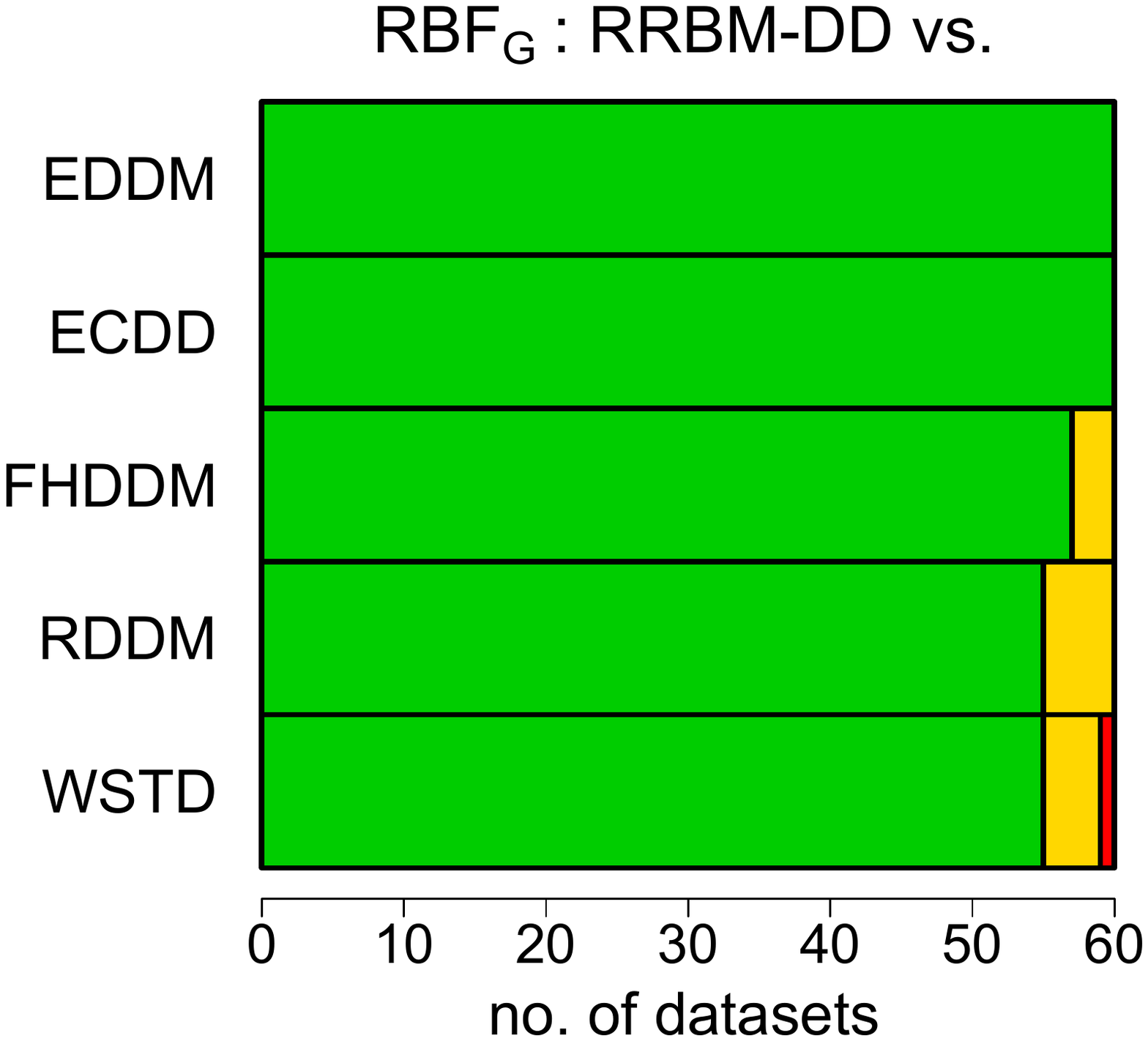}\vspace*{-1.5cm}
			\includegraphics[width=0.3\linewidth,trim=2cm 2cm 2cm 2cm,clip]{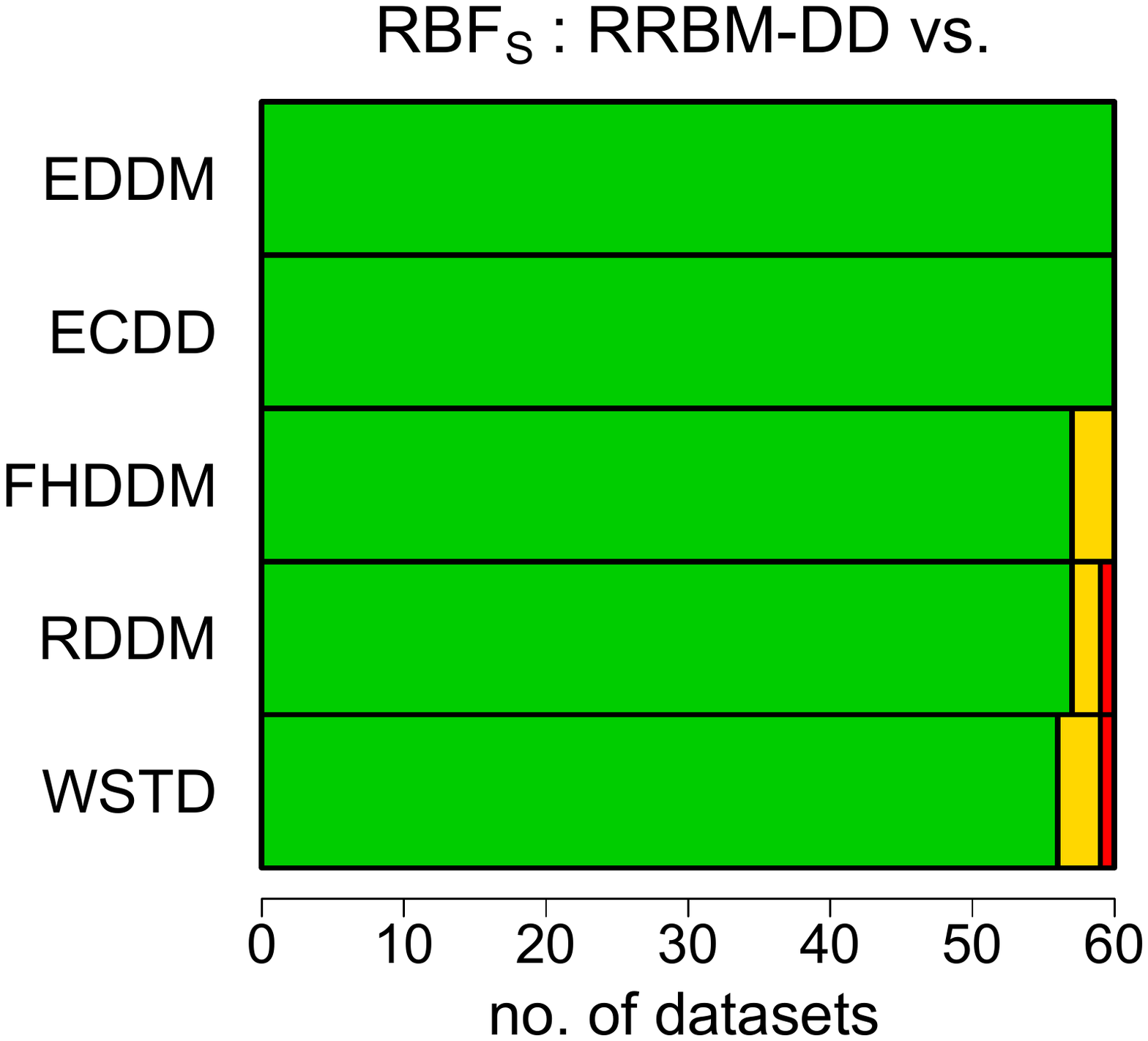}
			\includegraphics[width=0.3\linewidth,trim=2cm 2cm 2cm 2cm,clip]{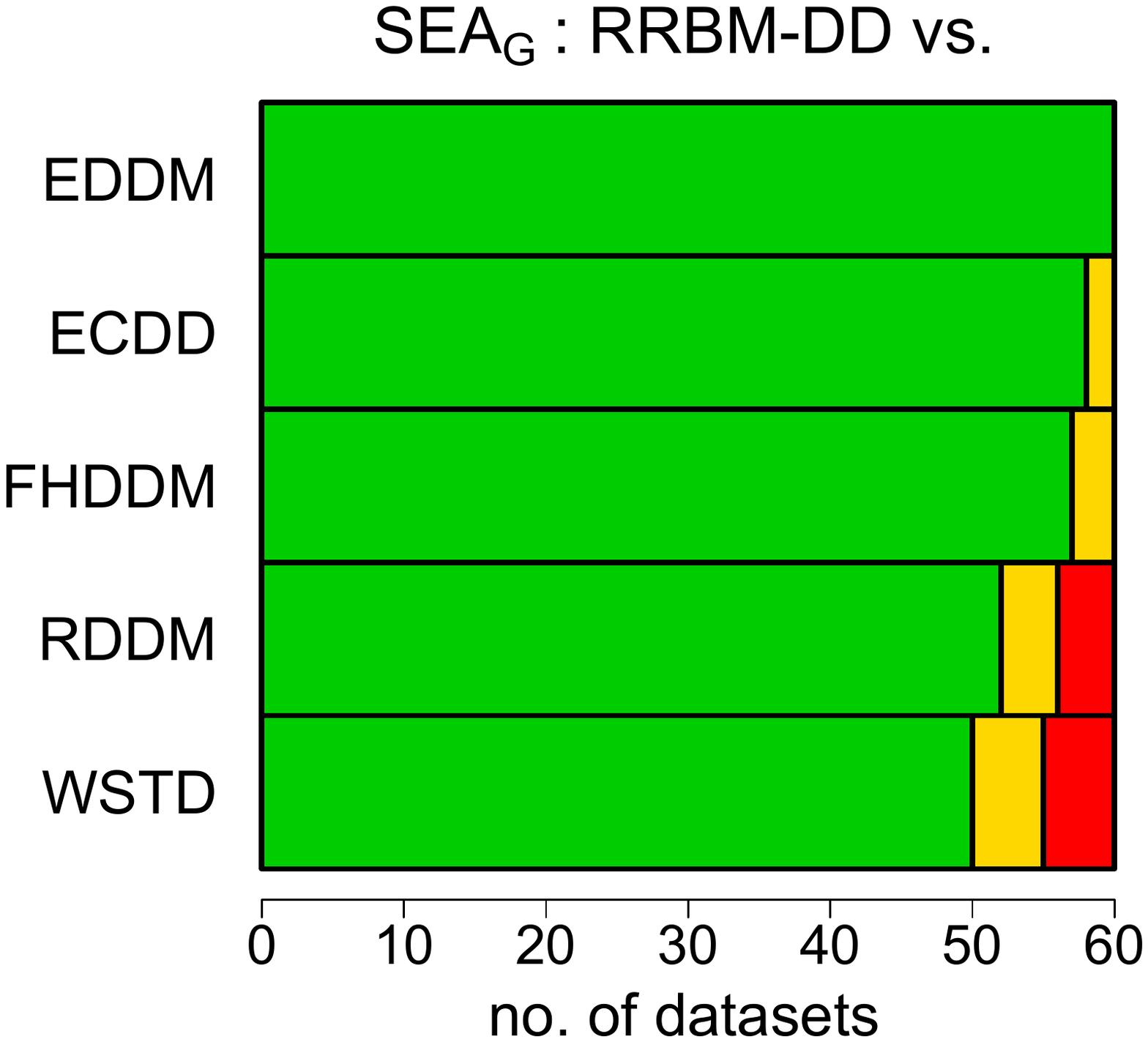}
			\includegraphics[width=0.3\linewidth,trim=2cm 2cm 2cm 2cm,clip]{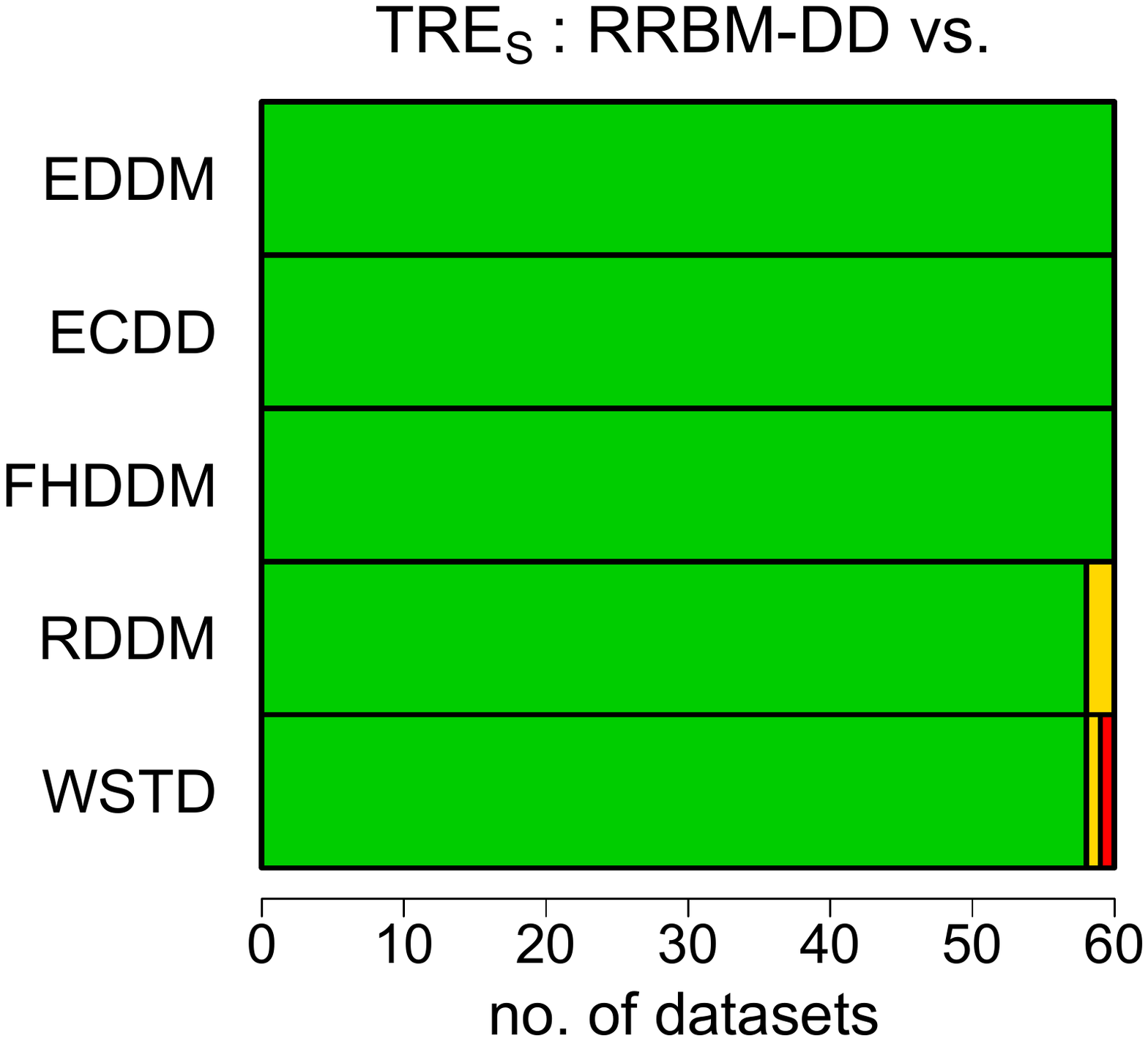}\vspace*{-1.5cm}
			\includegraphics[width=0.3\linewidth,trim=2cm 2cm 2cm 2cm,clip]{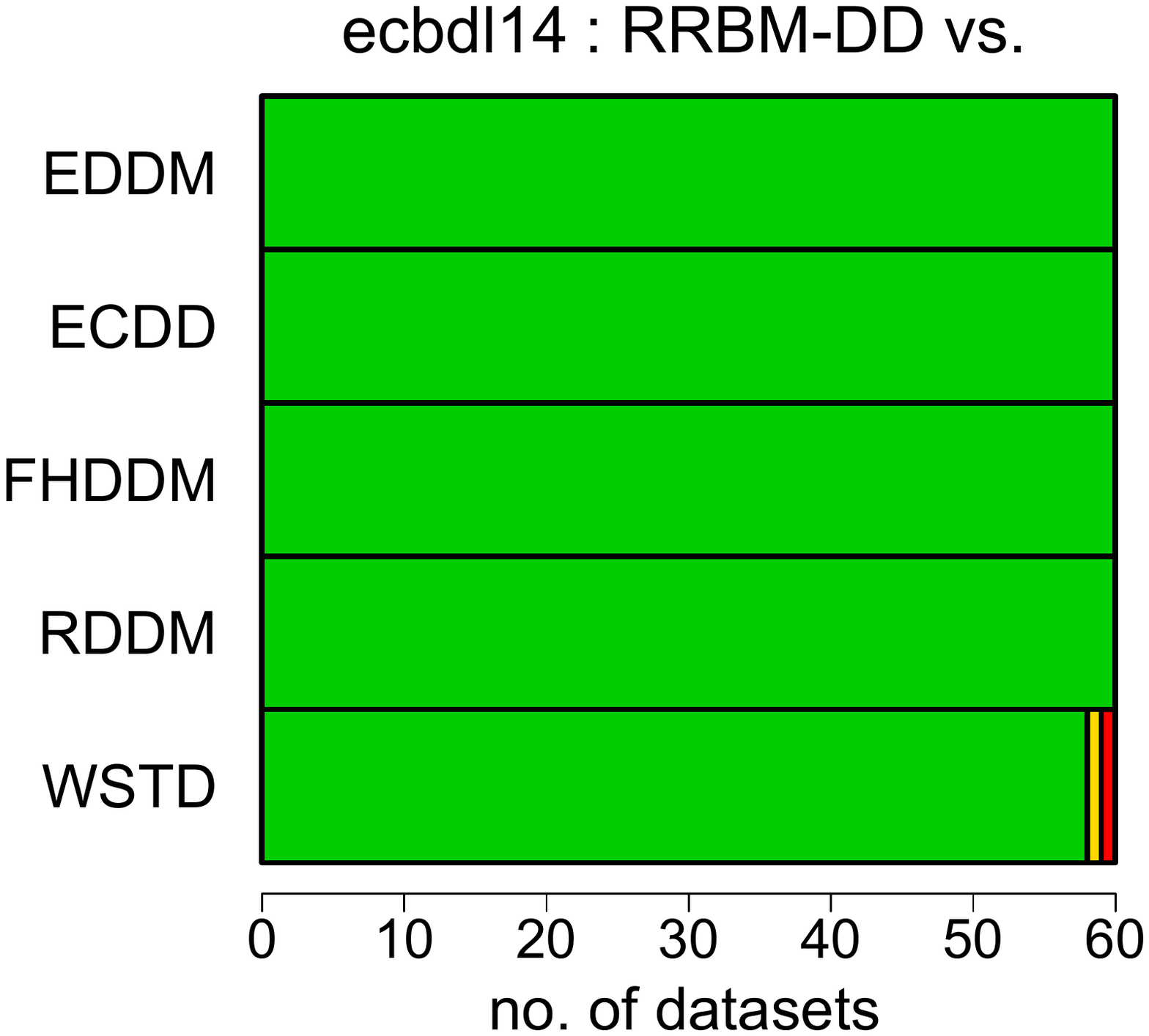}
			\includegraphics[width=0.3\linewidth,trim=2cm 2cm 2cm 2cm,clip]{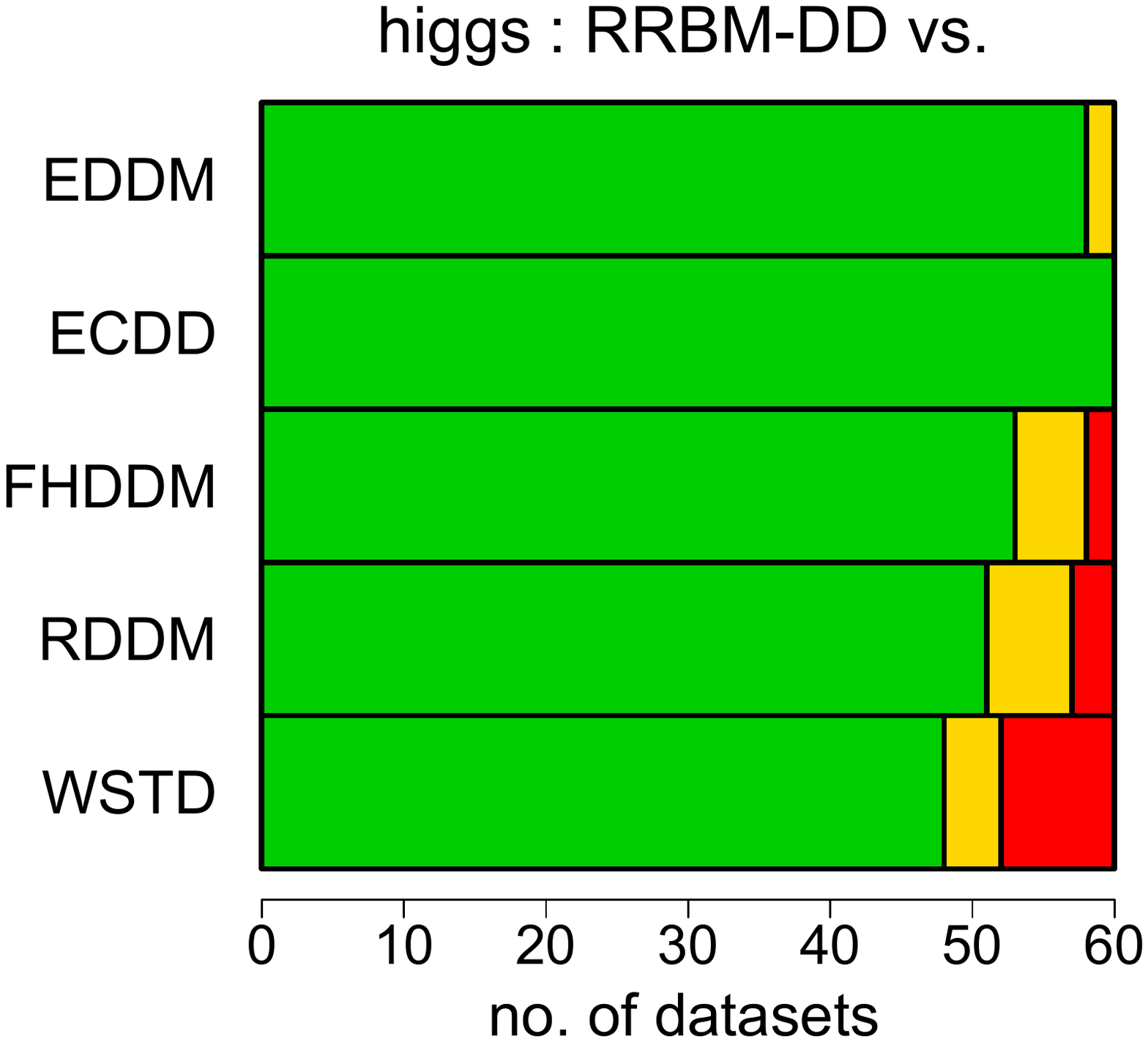}
			\includegraphics[width=0.3\linewidth,trim=2cm 2cm 2cm 2cm,clip]{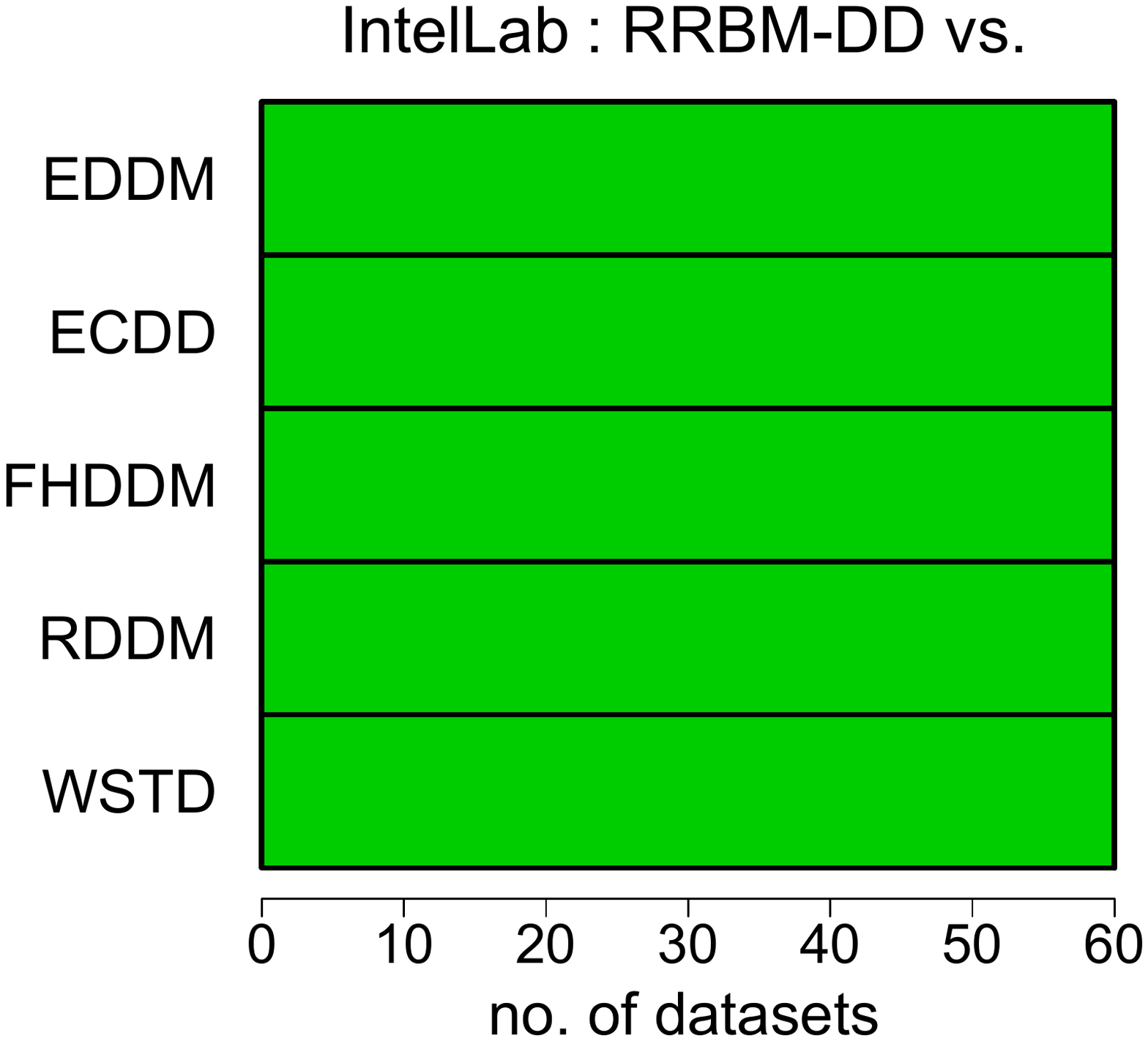}\vspace*{-1.5cm}
			\includegraphics[width=0.3\linewidth,trim=2cm 2cm 2cm 2cm,clip]{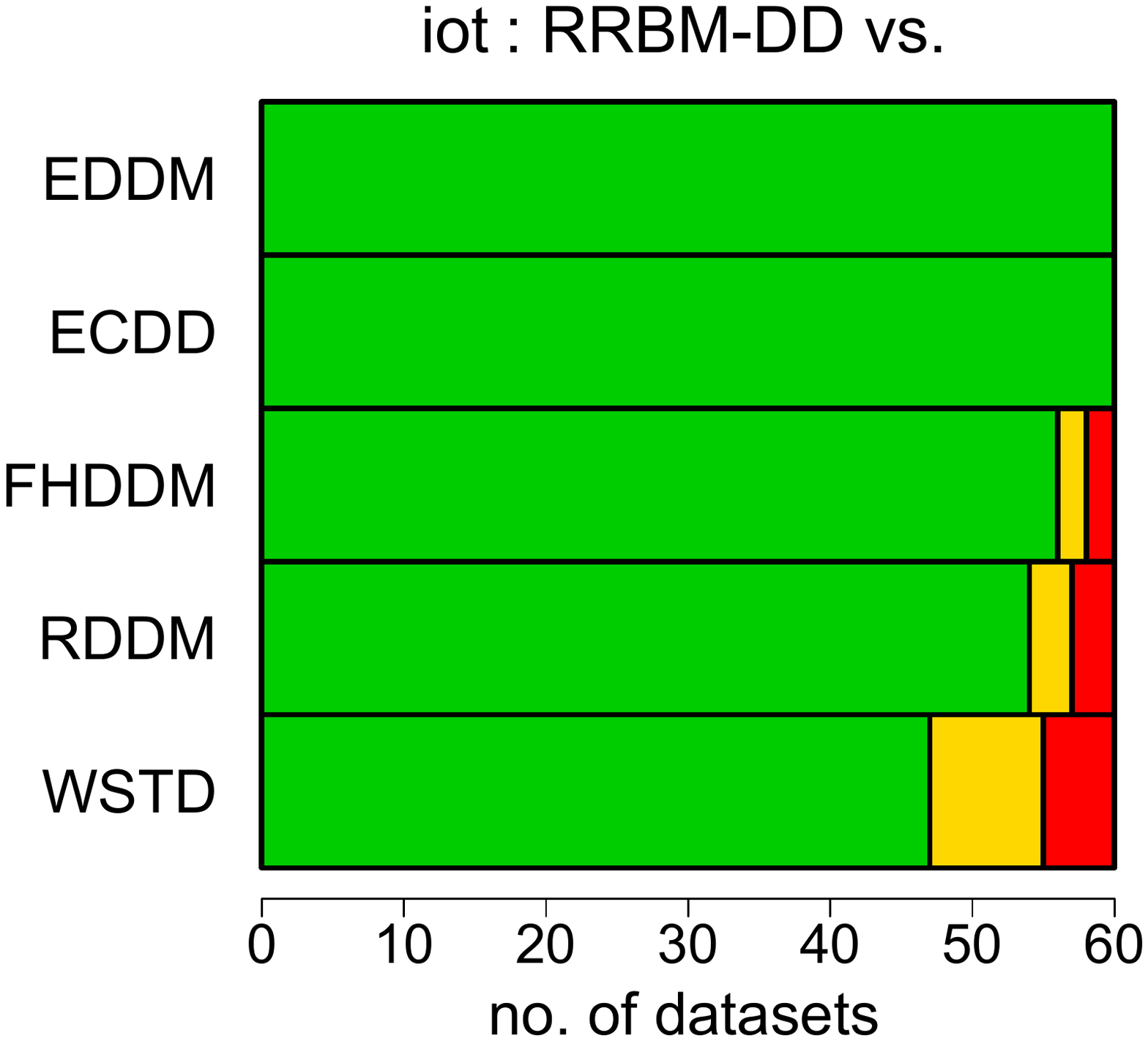}
			\includegraphics[width=0.3\linewidth,trim=2cm 2cm 2cm 2cm,clip]{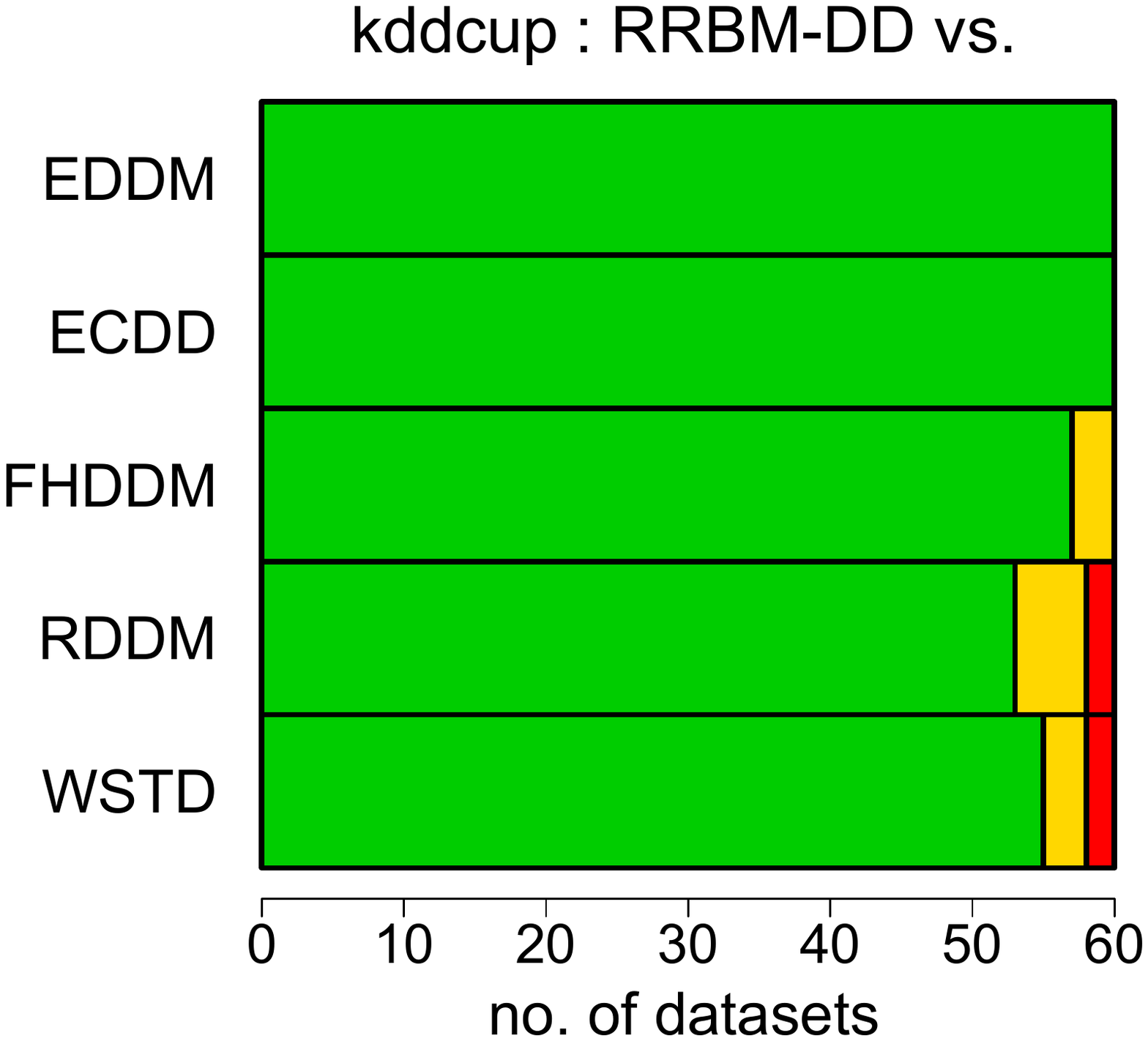}
			\includegraphics[width=0.3\linewidth,trim=2cm 2cm 2cm 2cm,clip]{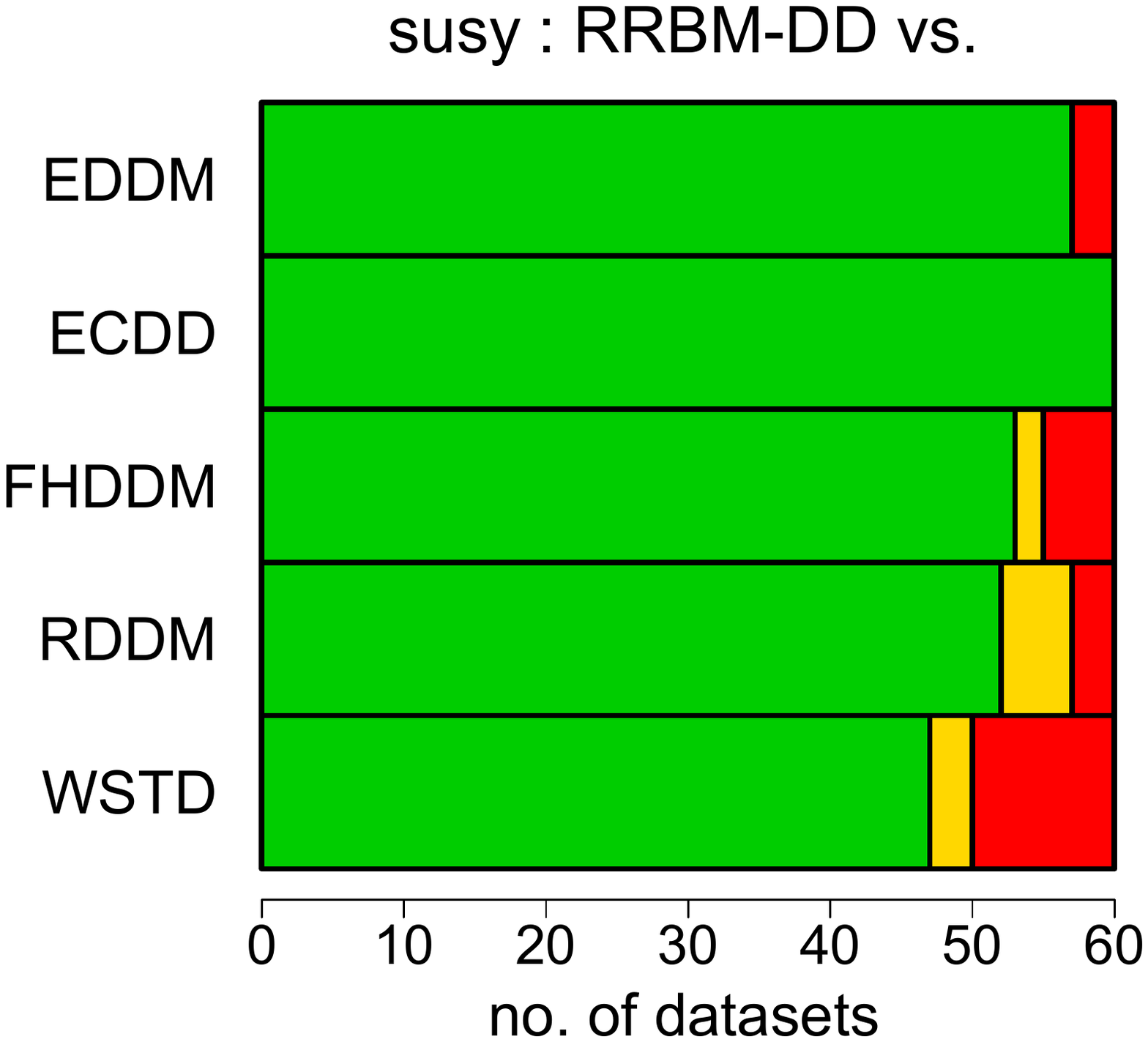}
			\caption{Comparison of RRBM--DD with reference drift detectors under instance-based poisoning attacks and sparsely labeled assumption (only 10\% labeled instances) with respect to the number of wins (green), ties (yellow), and losses (red), according to a pairwise F-test with statistical significance level $\alpha = 0.05$. Prequential accuracy used as a metric. 60 runs per data stream were obtained from 6 different labeling budgets, each repeated 10 times with random selection of instances to be labeled. }
			\label{fig:exp31}
		\end{figure}

\begin{figure}[h!]
			\centering
			\includegraphics[width=0.3\linewidth,trim=2cm 2cm 2cm 2cm,clip]{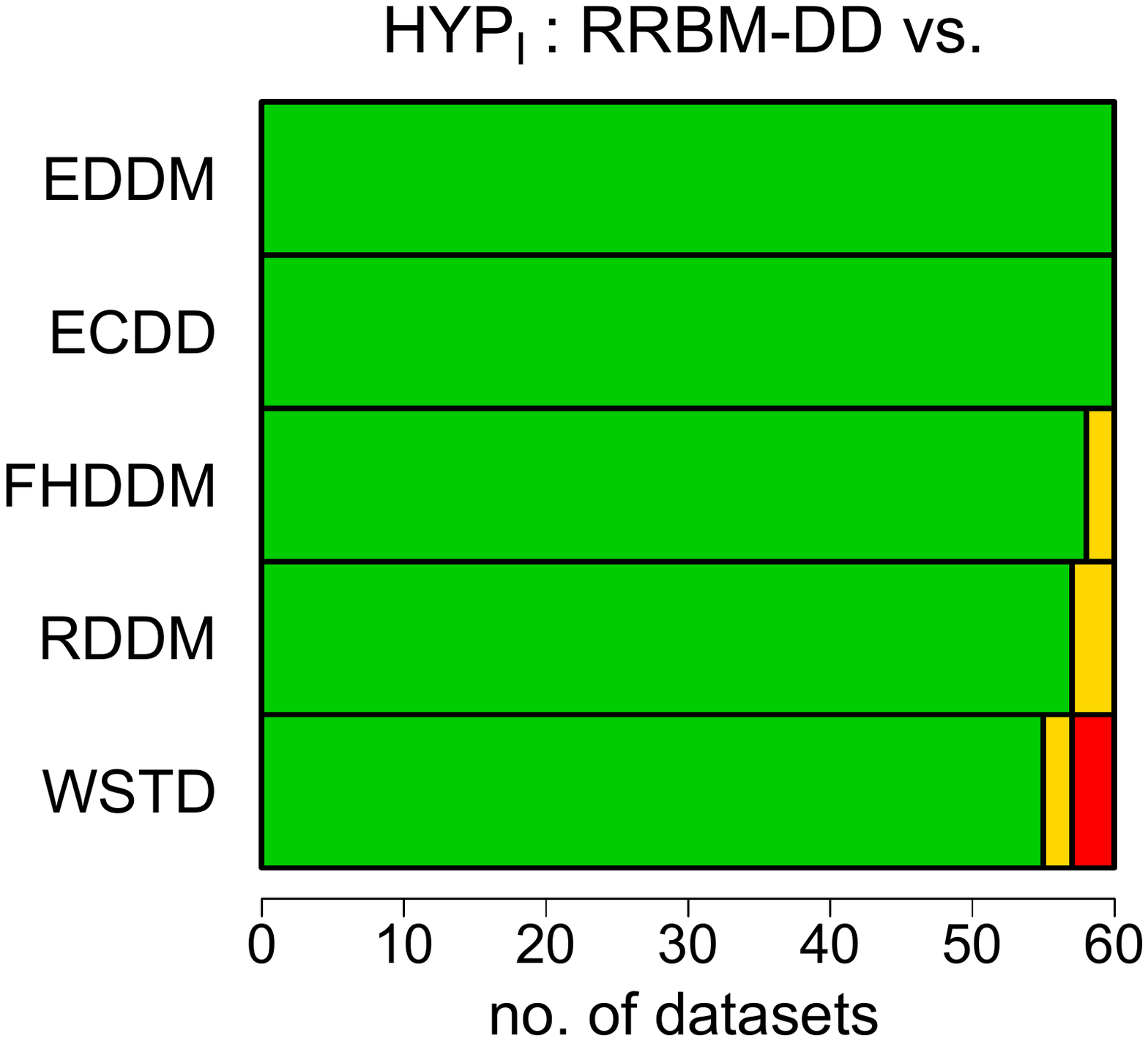}				    		   
			\includegraphics[width=0.3\linewidth,trim=2cm 2cm 2cm 2cm,clip]{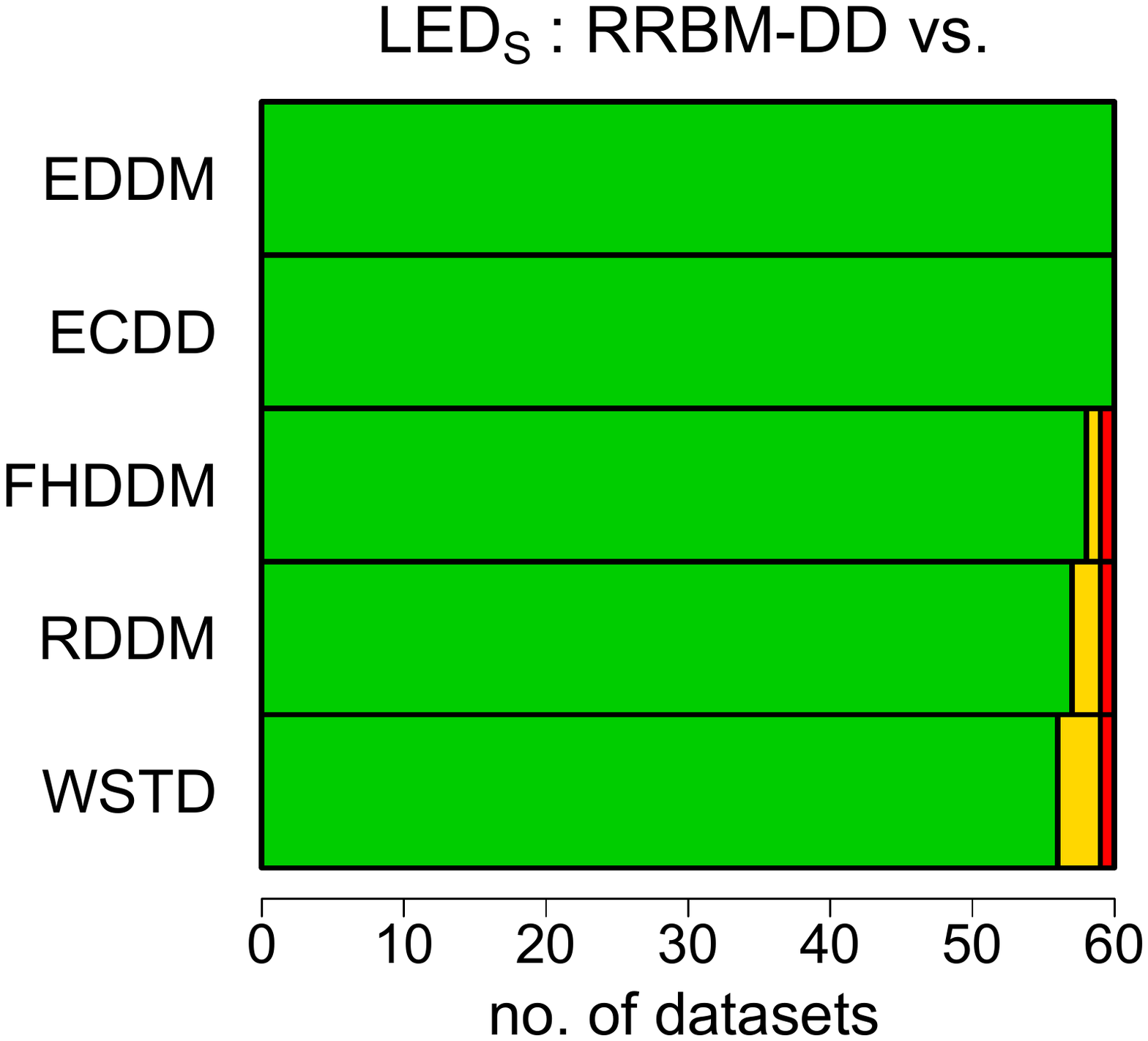}
			\includegraphics[width=0.3\linewidth,trim=2cm 2cm 2cm 2cm,clip]{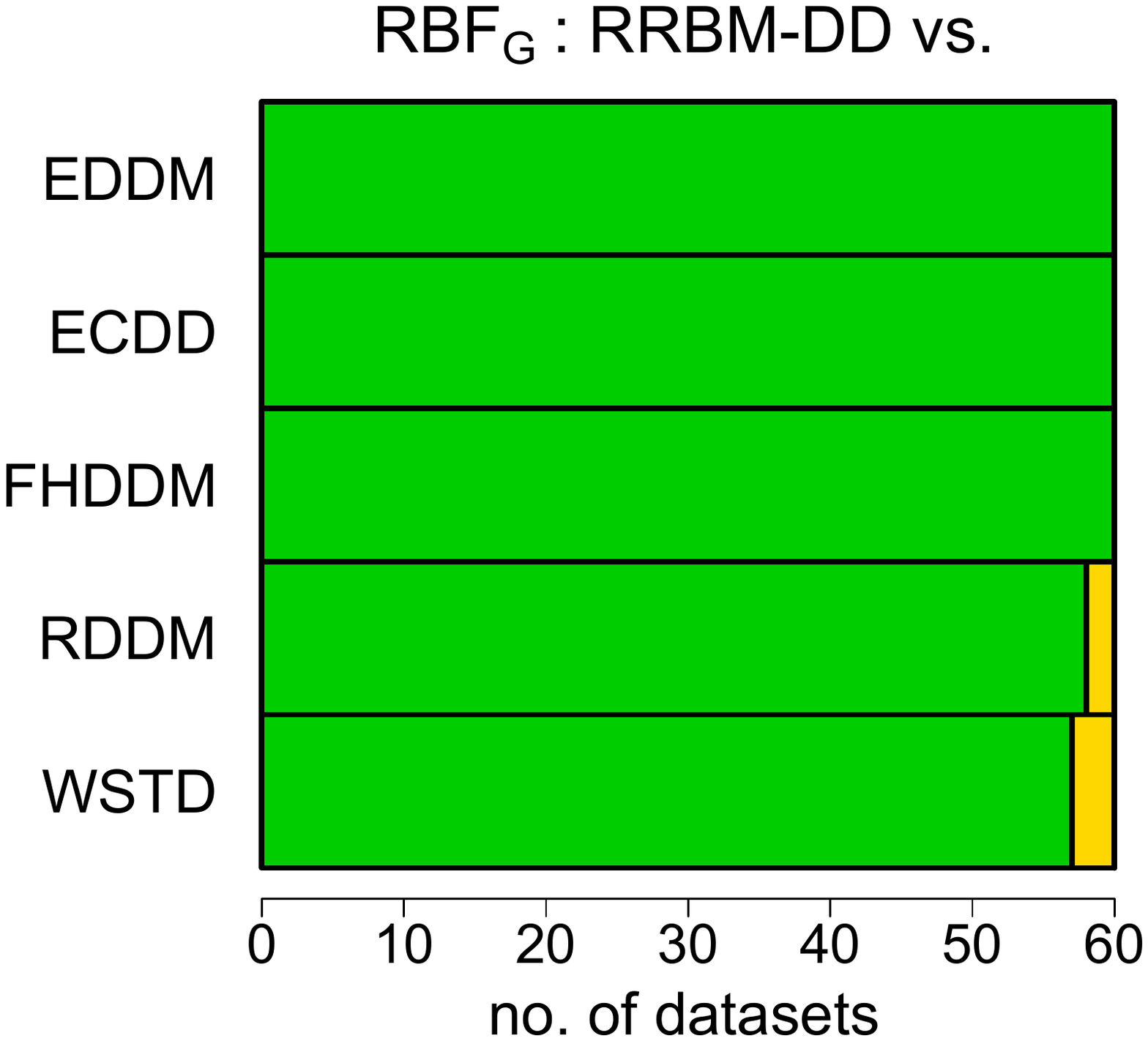}\vspace*{-1.5cm}
			\includegraphics[width=0.3\linewidth,trim=2cm 2cm 2cm 2cm,clip]{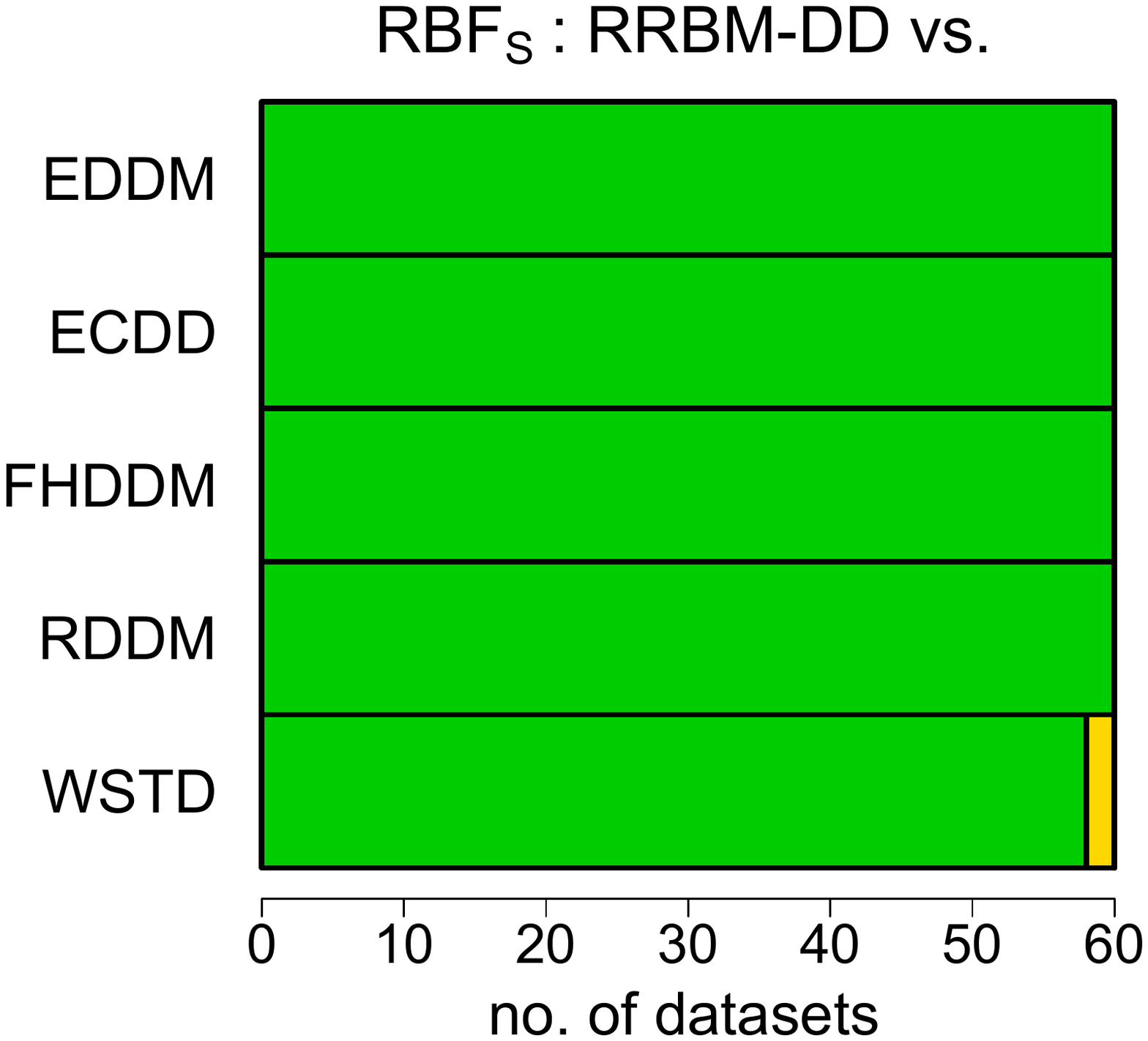}
			\includegraphics[width=0.3\linewidth,trim=2cm 2cm 2cm 2cm,clip]{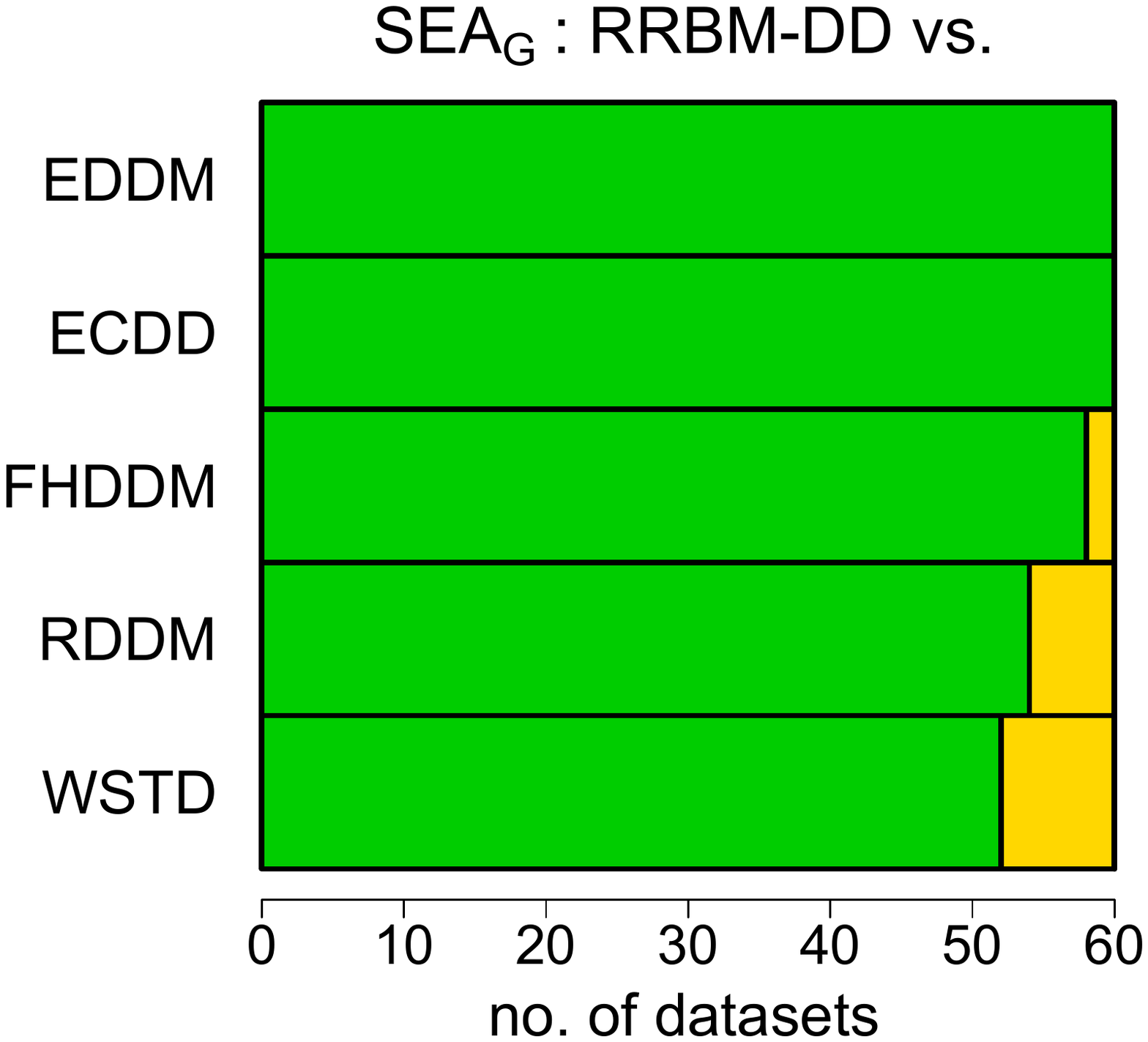}
			\includegraphics[width=0.3\linewidth,trim=2cm 2cm 2cm 2cm,clip]{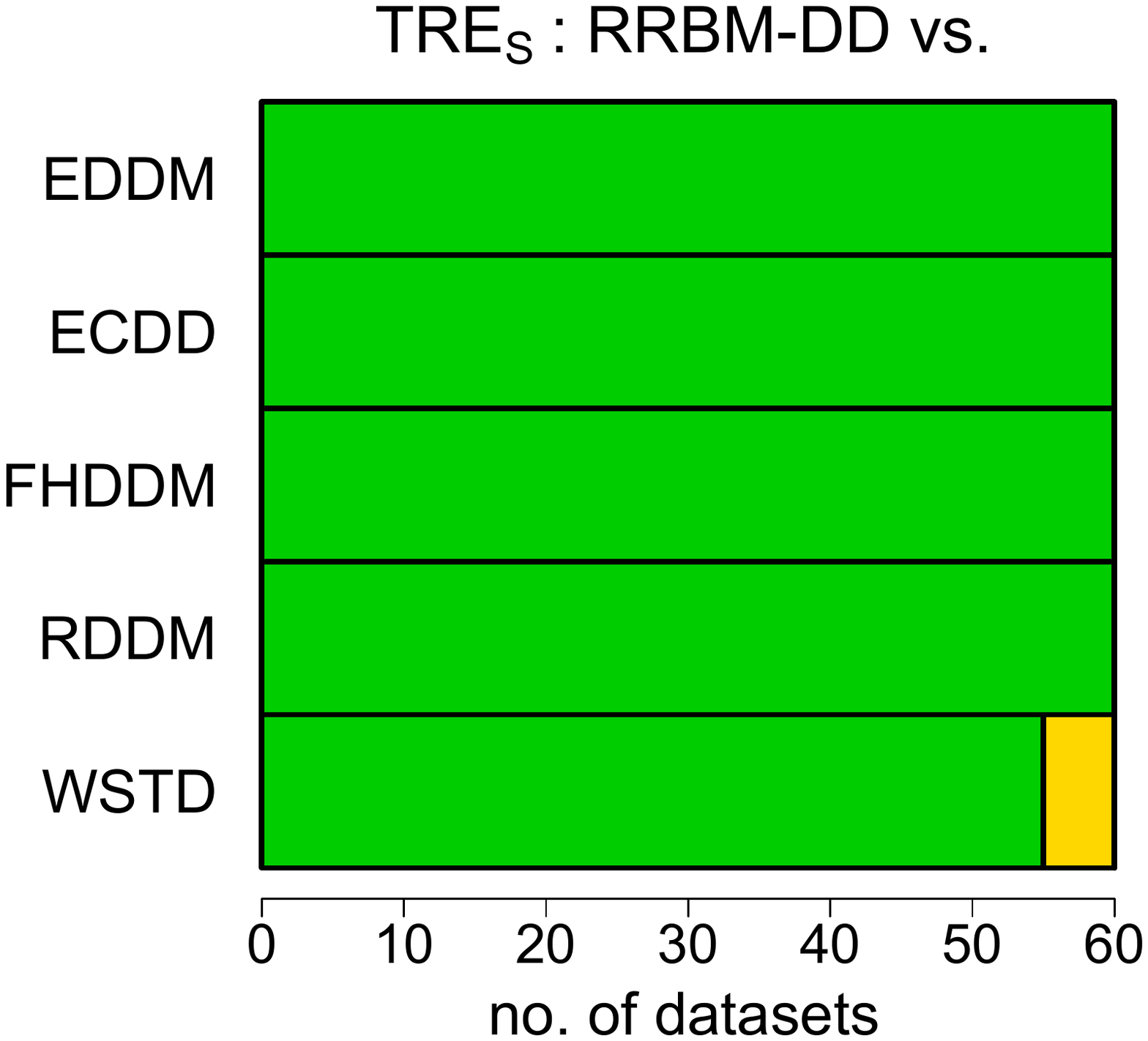}\vspace*{-1.5cm}
			\includegraphics[width=0.3\linewidth,trim=2cm 2cm 2cm 2cm,clip]{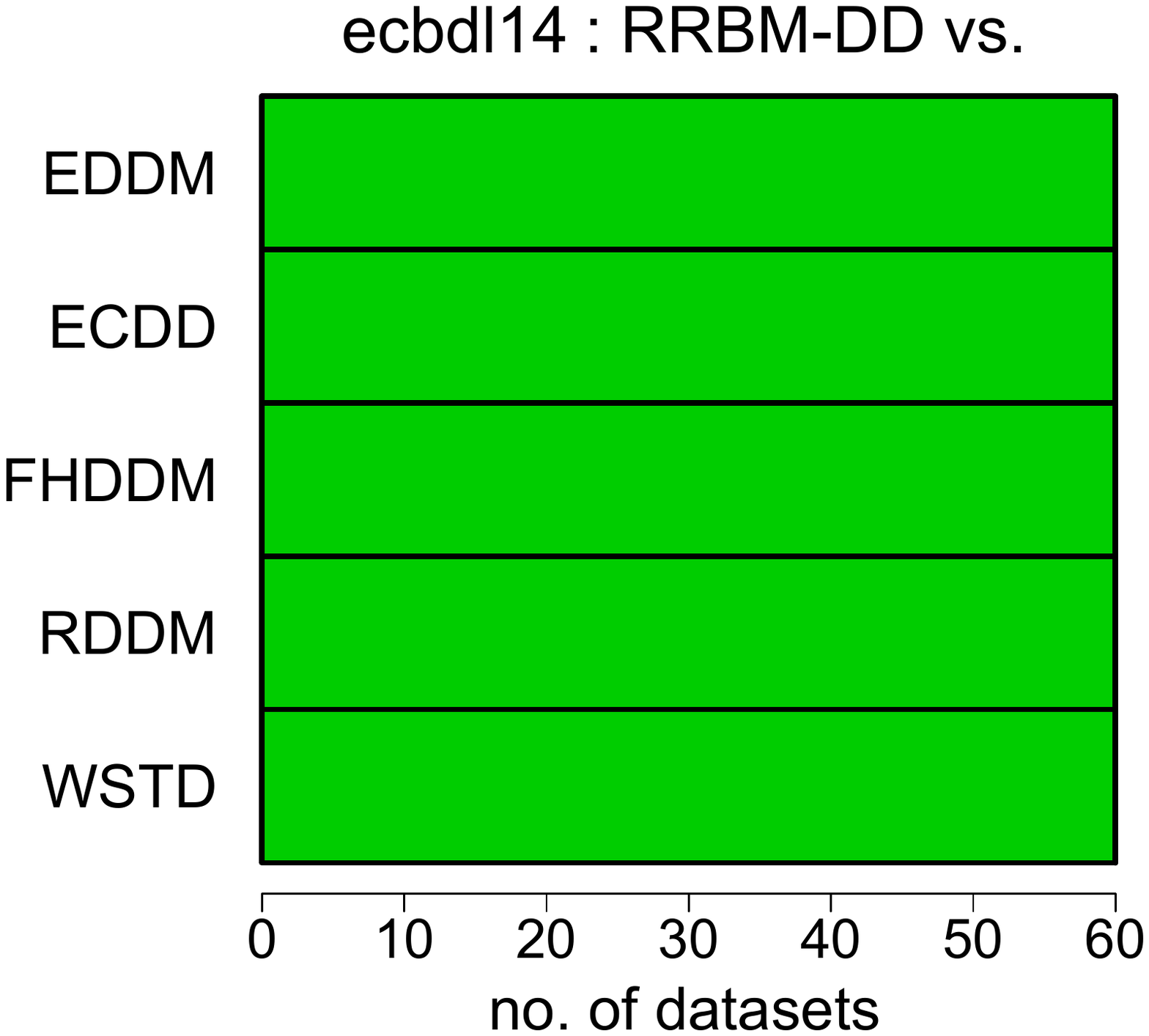}
			\includegraphics[width=0.3\linewidth,trim=2cm 2cm 2cm 2cm,clip]{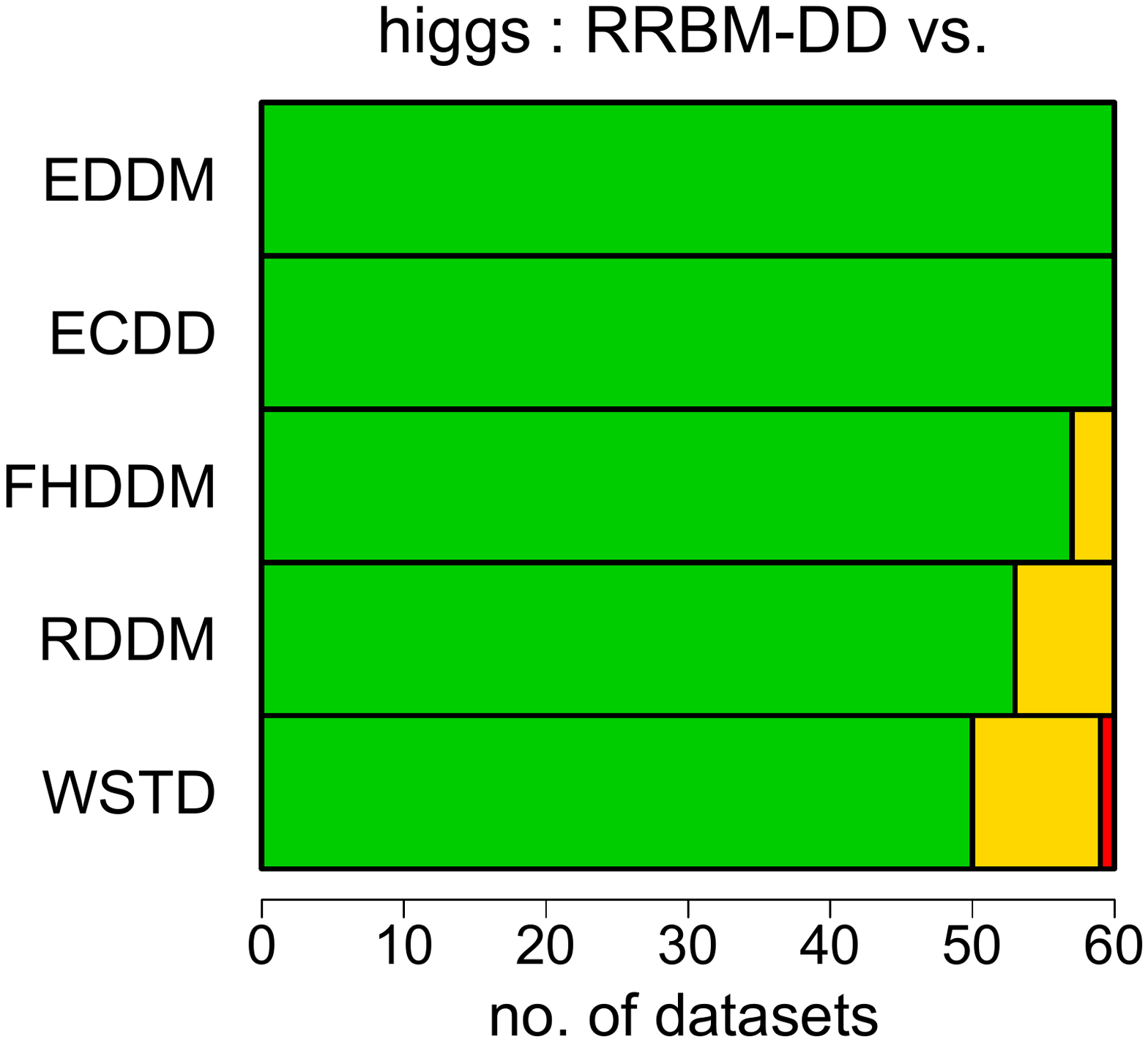}
			\includegraphics[width=0.3\linewidth,trim=2cm 2cm 2cm 2cm,clip]{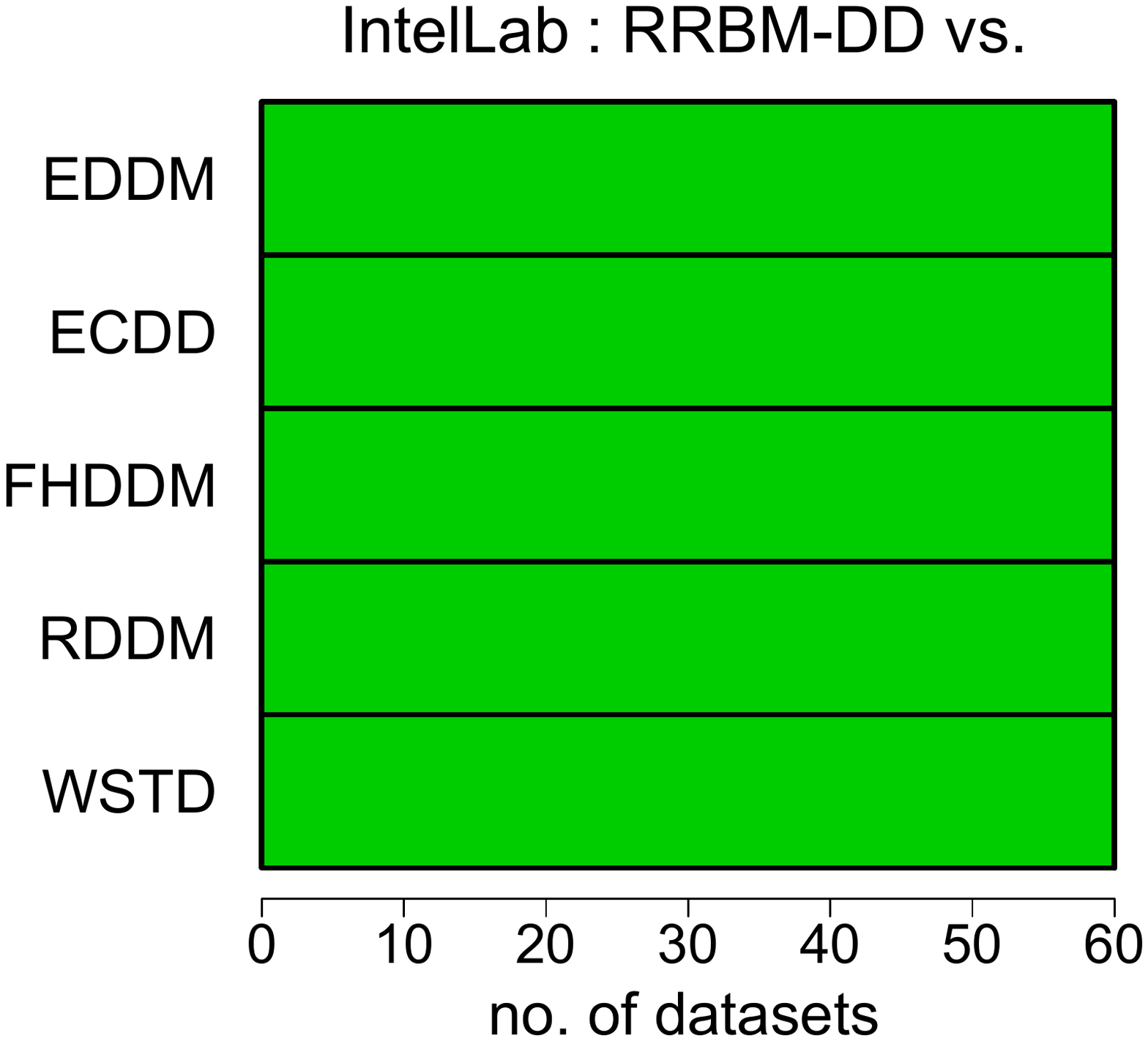}\vspace*{-1.5cm}
			\includegraphics[width=0.3\linewidth,trim=2cm 2cm 2cm 2cm,clip]{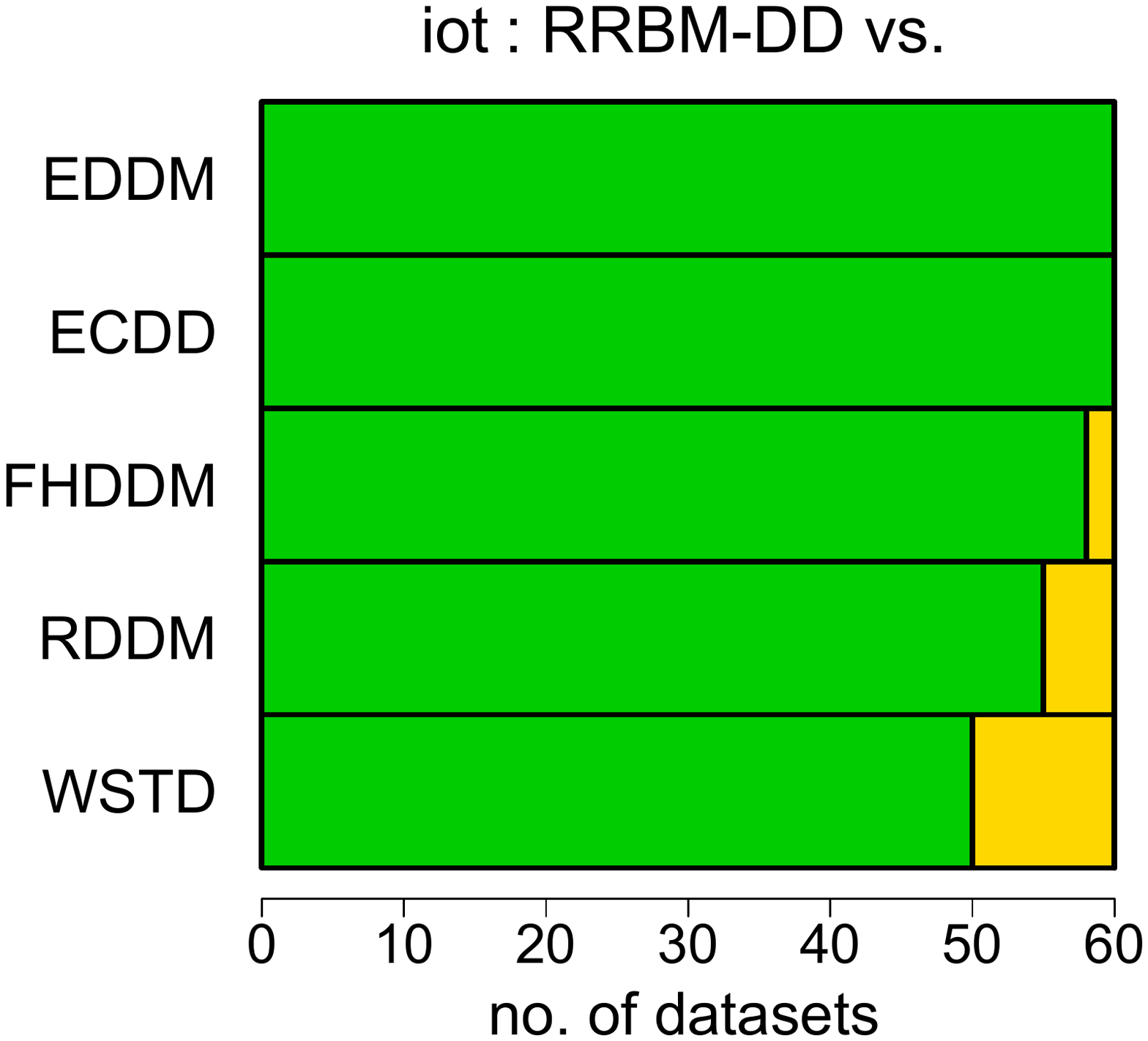}
			\includegraphics[width=0.3\linewidth,trim=2cm 2cm 2cm 2cm,clip]{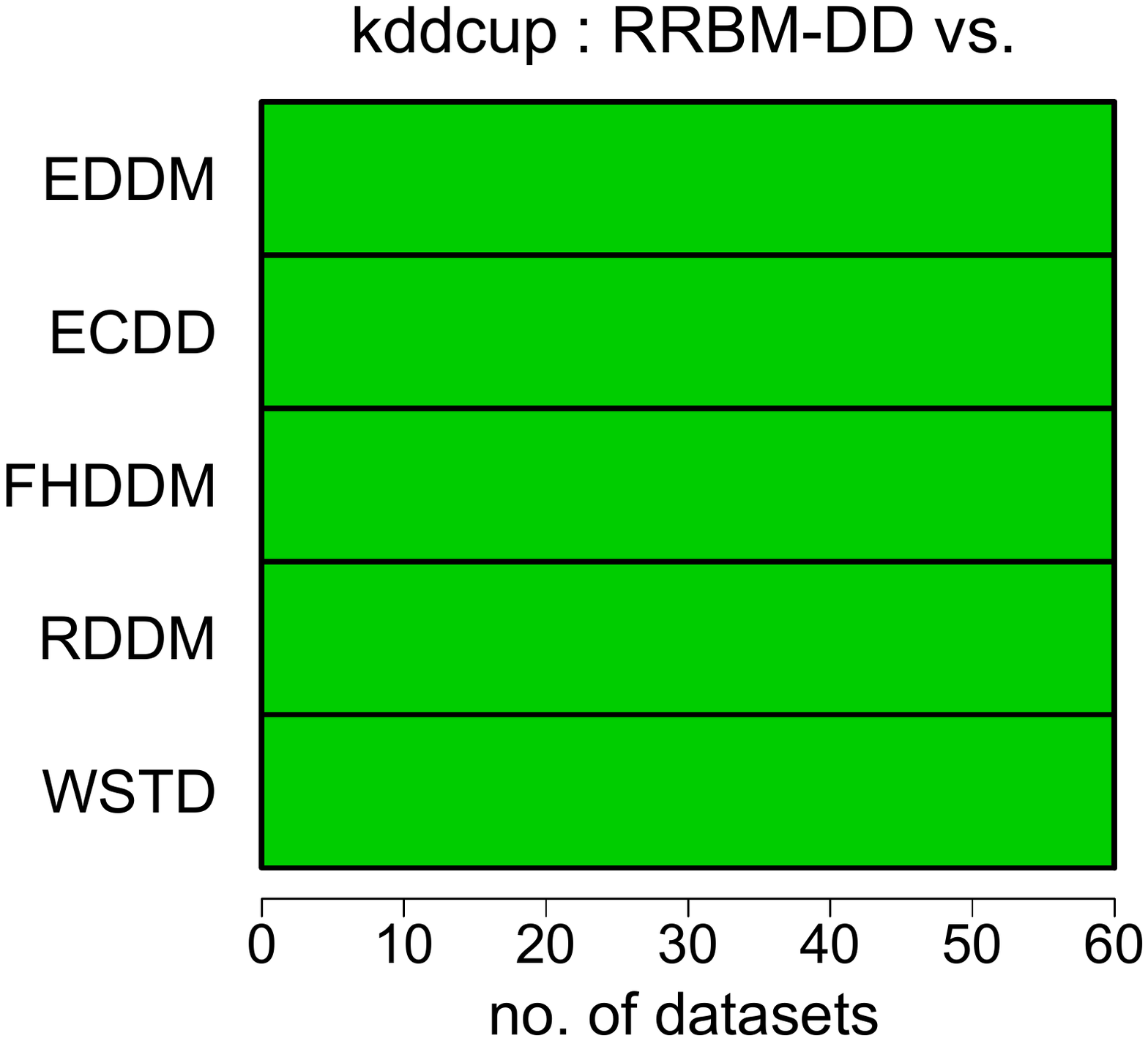}
			\includegraphics[width=0.3\linewidth,trim=2cm 2cm 2cm 2cm,clip]{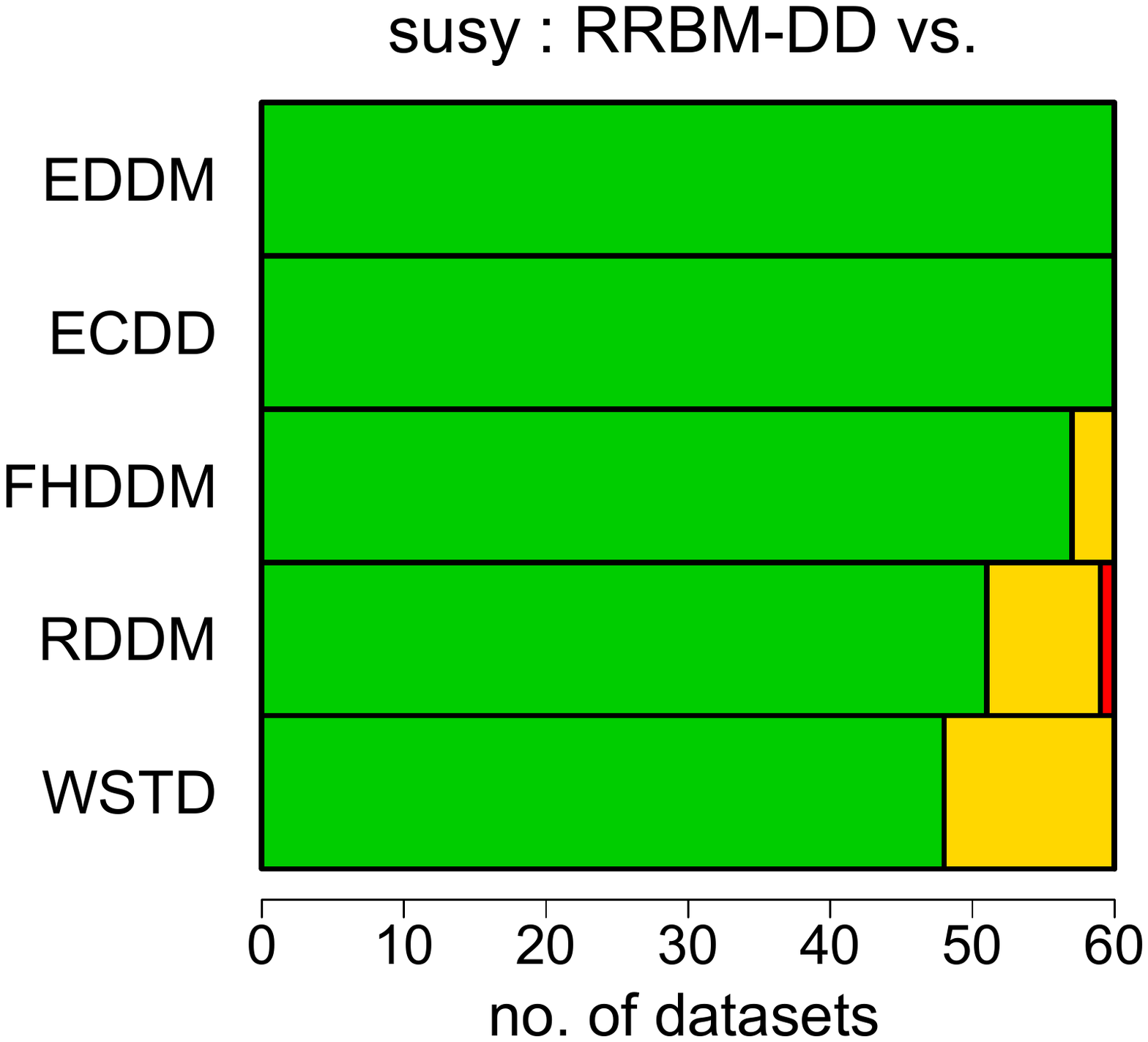}
			\caption{Comparison of RRBM--DD with reference drift detectors under concept-based poisoning attacks and sparsely labeled assumption with respect to the number of wins (green), ties (yellow), and losses (red), according to a pairwise F-test with statistical significance level $\alpha = 0.05$. Prequential accuracy used as a metric. 60 runs per data stream were obtained from 6 different labeling budgets, each repeated 10 times with random selection of instances to be labeled.}
			\label{fig:exp32}
		\end{figure}

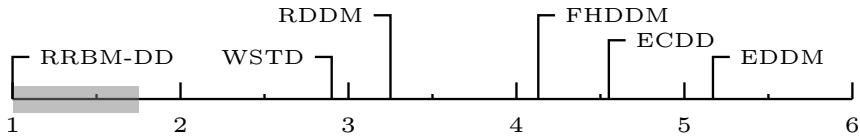
\begin{figure}[h]
	\centering
	\tiny
	\resizebox{\columnwidth}{!}{
		\begin{tikzpicture}
		\draw (1,0) -- (6,0);
		\foreach \x in {1,2,3,4,5,6} {
			\draw (\x, 0) -- ++(0,.1) node [below=0.15cm,scale=0.75] {\tiny \x};
			\ifthenelse{\x < 6}{\draw (\x+.5, 0) -- ++(0,.03);}{}
		}
		\coordinate (c0) at (1.00,0);
		\coordinate (c1) at (2.90,0);
		\coordinate (c2) at (3.25,0);
		\coordinate (c3) at (4.13,0);
		\coordinate (c4) at (4.55,0);
		\coordinate (c5) at (5.17,0);
		
		\node (l0) at (c0) [above right=.15cm and 0.1cm, align=center, scale=0.7] {\tiny RRBM-DD};
		\node (l1) at (c1) [above left=.15cm and 0.1cm, align=center, scale=0.7] {\tiny WSTD};
		\node (l2) at (c2) [above left=.40cm and 0.1cm, align=left, scale=0.7] {\tiny RDDM};
		\node (l3) at (c3) [above right=.40cm and 0.1cm, align=center, scale=0.7] {\tiny FHDDM};
		\node (l4) at (c4) [above right=.25cm and 0.1cm, align=center, scale=0.7] {\tiny ECDD};
		\node (l5) at (c5) [above right=.15cm and 0.1cm, align=left, scale=0.7] {\tiny EDDM};
		
		\fill[fill=gray,fill opacity=0.5] (1.00,-0.08) rectangle (1.75,0.08);
		
		\foreach \x in {0,...,5} {
			\draw (l\x) -| (c\x);
		};
		\end{tikzpicture}
	}
	\caption{The Bonferroni-Dunn test for comparison among drift detectors under instance-based poisoning attacks and sparsely labeled data, based on RLR and all examined labeling ratios.}
	\label{fig:bon5}
\end{figure}

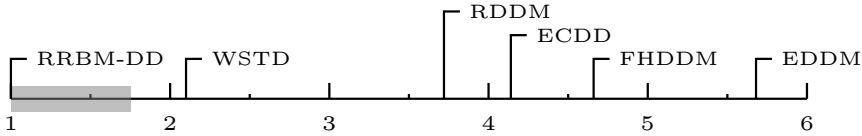
\begin{figure}[h]
	\centering
	\tiny
	\resizebox{\columnwidth}{!}{
		\begin{tikzpicture}
		\draw (1,0) -- (6,0);
		\foreach \x in {1,2,3,4,5,6} {
			\draw (\x, 0) -- ++(0,.1) node [below=0.15cm,scale=0.75] {\tiny \x};
			\ifthenelse{\x < 6}{\draw (\x+.5, 0) -- ++(0,.03);}{}
		}
		\coordinate (c0) at (1.00,0);
		\coordinate (c1) at (2.10,0);
		\coordinate (c2) at (3.72,0);
		\coordinate (c3) at (4.66,0);
		\coordinate (c4) at (4.14,0);
		\coordinate (c5) at (5.68,0);
		
		\node (l0) at (c0) [above right=.15cm and 0.1cm, align=center, scale=0.7] {\tiny RRBM-DD};
		\node (l1) at (c1) [above right=.15cm and 0.1cm, align=center, scale=0.7] {\tiny WSTD};
		\node (l2) at (c2) [above right=.45cm and 0.1cm, align=left, scale=0.7] {\tiny RDDM};
		\node (l3) at (c3) [above right=.15cm and 0.1cm, align=center, scale=0.7] {\tiny FHDDM};
		\node (l4) at (c4) [above right=.30cm and 0.1cm, align=center, scale=0.7] {\tiny ECDD};
		\node (l5) at (c5) [above right=.15cm and 0.1cm, align=left, scale=0.7] {\tiny EDDM};
		
		\fill[fill=gray,fill opacity=0.5] (1.00,-0.08) rectangle (1.75,0.08);
		
		\foreach \x in {0,...,5} {
			\draw (l\x) -| (c\x);
		};
		\end{tikzpicture}
	}
	\caption{The Bonferroni-Dunn test for comparison among drift detectors under concept-based poisoning attacks and sparsely labeled data, based on RLR and all examined labeling ratios.}
	\label{fig:bon6}
\end{figure}

The third experiment was designed to investigate the behavior of RRDM--DD under sparse access to class labels. As discussed in Sections~\ref{sec:int} and ~\ref{sec:apa} a common pitfall of many algorithms for learning from data streams lies in the unrealistic assumption that there is unlimited access to labeled instances. It is crucial to acknowledge that any algorithm for data stream mining, be it classifier or drift detector, must be flexible enough to work in scenarios where class labels are sparse. We used 12 benchmark data streams to select six different percentages of labeled instances. The label query procedure was repeated 10 times, in order to avoid impacting the results with selecting easy or difficult instances. This resulted in 60 runs for each out of 12 benchmark streams. Figures~\ref{fig:exp31} and \ref{fig:exp32} present the win-tie-loss plots for comparing the performance of RRBM--DD on sparsely labeled streams under instance-based and concept-based poisoning attacks. Tables~\ref{tab:rlr3} and ~\ref{tab:rlr4} present the RLR metrics for RRDM-DD and reference methods under both types of attacks, while Figures~\ref{fig:bon5} and~\ref{fig:bon6} show the visualizations of the Friedman ranking test with Bonferroni-Dunn post-hoc analysis.

			\begin{table}[h]
		\centering
		\caption{RLR for RRBM--DD and reference drift detectors under instance-based poisoning attacks and sparsely labeled data. Results averaged over all labeling budgets, presented with standard deviation.}
		\begin{tabular}{lcccccc}
			\toprule
			Stream & EDDM & ECDD & FHDDM & RDDM & WSTD & RRBM--DD \\ 
			\midrule
			HYP$_{I}$ & 0.47$\pm$0.11 & 0.49$\pm$0.10 & 0.61$\pm$0.12 & 0.55$\pm$0.16 & 0.57$\pm$0.13 & 0.80$\pm$0.08 \\
			LED$_S$ &  0.52$\pm$0.09 & 0.51$\pm$0.09 & 0.63$\pm$0.10 & 0.61$\pm$0.14 & 0.63$\pm$0.11 & 0.83$\pm$0.05 \\
			RBF$_G$ &  0.40$\pm$0.16 & 0.42$\pm$0.12 & 0.47$\pm$0.09 & 0.45$\pm$0.15 & 0.50$\pm$0.11 & 0.69$\pm$0.09 \\
			RBF$_S$ &  0.37$\pm$0.06 & 0.38$\pm$0.05 & 0.41$\pm$0.07 & 0.40$\pm$0.09 & 0.44$\pm$0.06 & 0.63$\pm$0.07 \\
			SEA$_G$ &  0.56$\pm$0.12 & 0.62$\pm$0.09 & 0.59$\pm$0.09 & 0.57$\pm$0.14 & 0.66$\pm$0.10 & 0.81$\pm$0.08 \\
			TRE$_S$ &  0.30$\pm$0.05 & 0.32$\pm$0.04 & 0.35$\pm$0.06 & 0.32$\pm$0.08 & 0.38$\pm$0.08 & 0.63$\pm$0.04 \\
			ecbdl14&  0.33$\pm$0.07 & 0.36$\pm$0.06 & 0.45$\pm$0.09 & 0.42$\pm$0.11 & 0.46$\pm$0.10 & 0.60$\pm$0.05 \\
			higgs & 0.58$\pm$0.14 & 0.62$\pm$0.10 & 0.65$\pm$0.12 & 0.64$\pm$0.16 & 0.70$\pm$0.11 & 0.82$\pm$0.07 \\
			IntelLab&  0.28$\pm$0.08 & 0.30$\pm$0.06 & 0.34$\pm$0.06 & 0.31$\pm$0.10 & 0.36$\pm$0.08 & 0.60$\pm$0.04 \\
			iot&  0.65$\pm$0.09 & 0.68$\pm$0.06 & 0.72$\pm$0.07 & 0.71$\pm$0.11 & 0.77$\pm$0.10 & 0.88$\pm$0.07 \\
			kddcup &  0.55$\pm$0.05 & 0.53$\pm$0.09 & 0.57$\pm$0.10 & 0.55$\pm$0.13 & 0.60$\pm$0.10 & 0.77$\pm$0.06 \\
			susy &  0.60$\pm$0.11 & 0.63$\pm$0.08 & 0.67$\pm$0.09 & 0.65$\pm$0.13 & 0.68$\pm$0.13 & 0.79$\pm$0.07 \\
			\midrule 
			avg. rank & 5.17 & 4.55 & 4.13 & 3.25 & 2.90 & 1.00 \\
			\bottomrule
			\label{tab:rlr3}
		\end{tabular}
	\end{table}

		\begin{table}[h]
	\centering
	\caption{RLR for RRBM--DD and reference drift detectors under concept-based poisoning attacks and sparsely labeled data. Results averaged over all labeling budgets, presented with standard deviation.}
	\begin{tabular}{lcccccc}
		\toprule
		Stream & EDDM & ECDD & FHDDM & RDDM & WSTD & RRBM--DD \\ 
		\midrule
			HYP$_{I}$ & 0.28$\pm$0.07 & 0.29$\pm$0.06 & 0.37$\pm$0.10 & 0.34$\pm$0.13 & 0.37$\pm$0.13 & 0.69$\pm$0.06 \\
			LED$_S$ &  0.25$\pm$0.05 & 0.32$\pm$0.05 & 0.44$\pm$0.08 & 0.41$\pm$0.10 & 0.46$\pm$0.08 & 0.75$\pm$0.07 \\
			RBF$_G$ &  0.23$\pm$0.06 & 0.26$\pm$0.07 & 0.33$\pm$0.07 & 0.29$\pm$0.12 & 0.34$\pm$0.14 & 0.62$\pm$0.06 \\
			RBF$_S$ &  0.18$\pm$0.04 & 0.19$\pm$0.06 & 0.20$\pm$0.04 & 0.18$\pm$0.05 & 0.22$\pm$0.04 & 0.59$\pm$0.04 \\
			SEA$_G$ &  0.40$\pm$0.12 & 0.42$\pm$0.13 & 0.44$\pm$0.10 & 0.41$\pm$0.15 & 0.45$\pm$0.12 & 0.74$\pm$0.08 \\
			TRE$_S$ &  0.09$\pm$0.03 & 0.11$\pm$0.04 & 0.14$\pm$0.04 & 0.13$\pm$0.04 & 0.17$\pm$0.04 & 0.51$\pm$0.05 \\
			ecbdl14&  0.08$\pm$0.02 & 0.07$\pm$0.02 & 0.11$\pm$0.02 & 0.09$\pm$0.03 & 0.13$\pm$0.02 & 0.48$\pm$0.05 \\
			higgs & 0.39$\pm$0.10 & 0.40$\pm$0.08 & 0.44$\pm$0.11 & 0.41$\pm$0.13 & 0.48$\pm$0.15 & 0.79$\pm$0.08 \\
			IntelLab&  0.12$\pm$0.04 & 0.14$\pm$0.04 & 0.16$\pm$0.05 & 0.13$\pm$0.05 & 0.18$\pm$0.04 & 0.58$\pm$0.09 \\
			iot&  0.51$\pm$0.13 & 0.52$\pm$0.16 & 0.56$\pm$0.12 & 0.55$\pm$0.16 & 0.60$\pm$0.13 & 0.81$\pm$0.11 \\
			kddcup &  0.23$\pm$0.03 & 0.20$\pm$0.05 & 0.24$\pm$0.05 & 0.19$\pm$0.05 & 0.28$\pm$0.03 & 0.68$\pm$0.05 \\
			susy &  0.31$\pm$0.07 & 0.34$\pm$0.07 & 0.36$\pm$0.09 & 0.37$\pm$0.10 & 0.39$\pm$0.10 & 0.70$\pm$0.07 \\
		\midrule 
		avg. rank & 5.38 & 4.14 & 4.66 & 3.72 & 2.10 & 1.00 \\
		\bottomrule
		\label{tab:rlr4}
	\end{tabular}
\end{table}

Sparse access to class labels significantly impacts state-of-the-art drift detectors. All of them were designed for fully supervised cases and use both feature values and class labels to compute statistics used in drift detection. By limiting the number of labeled instances they obtain, the estimators of used statistics become less and less reliable. Detectors are forced to make decisions based on a small, potentially non-representative sample and thus are much more prone to errors. This is already a very difficult problem, but becomes even more challenging when combined with the presence of adversarial concept drift. RRBM--DD, due to its trainable nature is capable of much better performance even under limited access to class labels. As we track the progress of the stream using regression-based trends, we can still compute them efficiently from a smaller sample size. Furthermore, used robust gradient was designed by its authors to work with smaller data streams \citep{Holland:2019}. 

\smallskip
\noindent \textbf{RQ3 answer.} Yes, RRBM-DD is capable of efficiently handling sparsely labeled data streams, without sacrificing its accuracy and robustness to adversarial concept drift.  

	\subsection{Experiment 4: Ablation study}
	\label{sec:exp4}

The fourth and final experiment was designed in the form of ablation study. Here, we want to switch off different components of RRBM--DD to gain understanding what is the source of its desirable performance and under what specific scenarios which component offers the most improvement to our drift detector. Tables~\ref{tab:rlr5} and~\ref{tab:rlr6} show the LRL metric performance averaged over fully and sparsely labeled data streams for four settings -- the proposed RRBM--DD model, its basic version RBM--DD without any robustness enhancements (discussed in Section 4.1), RBM--DD$_{RG}$ using only robust gradient calculation and RBM--DD$_{RE}$ using only robust energy function for estimating network state. 
	
			\begin{table}[h]
		\centering
		\caption{Ablation results for RRBM--DD according to RLR under instance-based poisoning attacks. Results averaged over fully and sparsely labeled benchmarks, presented with standard deviation.}
		\begin{tabular}{lcccc}
			\toprule
			Stream & RBM-DD & RBM--DD$_{RG}$ & RBM--DD$_{RE}$ & RRBM-DD \\ 
			\midrule
			HYP$_{I}$ & 0.58$\pm$0.10 & 0.77$\pm$0.09 & 0.64$\pm$0.09 & 0.82$\pm$0.09 \\
			LED$_S$ & 0.66$\pm$0.12 & 0.80$\pm$0.07 & 0.72$\pm$0.10 & 0.84$\pm$0.06 \\
			RBF$_G$ & 0.52$\pm$0.11 & 0.66$\pm$0.09 & 0.59$\pm$0.10 & 0.72$\pm$0.09 \\
			RBF$_S$ & 0.47$\pm$0.08 & 0.58$\pm$0.09 & 0.53$\pm$0.07 & 0.66$\pm$0.08 \\
			SEA$_G$ & 0.68$\pm$0.11 & 0.78$\pm$0.10 & 0.74$\pm$0.10 & 0.82$\pm$0.10 \\
			TRE$_S$ & 0.41$\pm$0.10 & 0.59$\pm$0.06 & 0.48$\pm$0.08 & 0.65$\pm$0.05 \\
			ecbdl14 & 0.48$\pm$0.11 & 0.58$\pm$0.07 & 0.53$\pm$0.09 & 0.61$\pm$0.07 \\
			higgs   & 0.73$\pm$0.11 & 0.82$\pm$0.09 & 0.77$\pm$0.10 & 0.85$\pm$0.08 \\
			IntelLab& 0.38$\pm$0.08 & 0.59$\pm$0.08 & 0.52$\pm$0.08 & 0.66$\pm$0.08 \\
			iot     & 0.77$\pm$0.10 & 0.86$\pm$0.05 & 0.83$\pm$0.08 & 0.90$\pm$0.05 \\
			kddcup  & 0.61$\pm$0.11 & 0.72$\pm$0.08 & 0.66$\pm$0.10 & 0.79$\pm$0.07 \\
			susy    & 0.70$\pm$0.12 & 0.80$\pm$0.09 & 0.73$\pm$0.09 & 0.83$\pm$0.09 \\
			\midrule 
			avg. rank & 4.00 & 2.05 & 2.95 & 1.00 \\
			\bottomrule
			\label{tab:rlr5}
		\end{tabular}
	\end{table}

			\begin{table}[h]
	\centering
	\caption{Ablation results for RRBM--DD according to RLR under concept-based poisoning attacks. Results averaged over fully and sparsely labeled benchmarks, presented with standard deviation.}
	\begin{tabular}{lcccc}
		\toprule
		Stream & RBM-DD & RBM--DD$_{RG}$ & RBM--DD$_{RE}$ & RRBM-DD \\ 
		\midrule
		HYP$_{I}$ & 0.38$\pm$0.11 & 0.44$\pm$0.10 & 0.65$\pm$0.06 & 0.72$\pm$0.07 \\
		LED$_S$ & 0.50$\pm$0.09 & 0.57$\pm$0.10 & 0.68$\pm$0.12 & 0.78$\pm$0.11 \\
		RBF$_G$ & 0.36$\pm$0.15 & 0.45$\pm$0.11 & 0.56$\pm$0.07 & 0.64$\pm$0.08 \\
		RBF$_S$ & 0.25$\pm$0.05 & 0.38$\pm$0.05 & 0.50$\pm$0.05 & 0.61$\pm$0.05 \\
		SEA$_G$ & 0.43$\pm$0.10 & 0.57$\pm$0.11 & 0.68$\pm$0.10 & 0.78$\pm$0.10 \\
		TRE$_S$ & 0.16$\pm$0.04 & 0.31$\pm$0.06 & 0.42$\pm$0.09 & 0.55$\pm$0.08 \\
		ecbdl14 & 0.14$\pm$0.04 & 0.29$\pm$0.05 & 0.40$\pm$0.08 & 0.51$\pm$0.07 \\
		higgs   & 0.51$\pm$0.13 & 0.65$\pm$0.11 & 0.76$\pm$0.10 & 0.83$\pm$0.11 \\
		IntelLab& 0.20$\pm$0.04 & 0.41$\pm$0.07 & 0.53$\pm$0.08 & 0.64$\pm$0.08 \\
		iot     & 0.64$\pm$0.13 & 0.69$\pm$0.12 & 0.74$\pm$0.10 & 0.83$\pm$0.11 \\
		kddcup  & 0.26$\pm$0.06 & 0.53$\pm$0.05 & 0.62$\pm$0.05 & 0.72$\pm$0.05 \\
		susy    & 0.37$\pm$0.11 & 0.50$\pm$0.09 & 0.63$\pm$0.07 & 0.75$\pm$0.08 \\
		\midrule 
		avg. rank & 4.00 & 2.90 & 2.10 & 1.00 \\
		\bottomrule
		\label{tab:rlr6}
	\end{tabular}
\end{table}

The basic RBM--DD performs only slightly better than state-of-the-art drift detectors. This shows that the excellent performance of RRBM--DD in the face of adversarial concept drift cannot be only contributed to its trainable nature or used neural model. One must notice that both gradient and energy function of RRBM--DD offer significant improvements on their own, but in specific conditions. Robust gradient offers a higher boost to robustness when dealing with instance-based poisoning attacks. This can be explained by the scaling approach used in it (see Eq.~\ref{eq:rgd2}--~\ref{eq:rgd4}), as it filters out instances that differ significantly from the core concept of each class. This makes it very robust to instance by instance attacks, as they will not affect significantly the training procedure of RRBM--DD. On the other hand, the robust energy function (see Eq.~\ref{eq:ref1}) offers much better performance when dealing with concept-based poisoning attacks. This can be contributed to the noise model that is embedded in said function. It allows to model not instances but entire noisy distributions, effectively cutting out adversarial concepts (i.e., disjuncts of instances) from strongly impacting the weight update of RRBM-DD. 

\smallskip
\noindent \textbf{RQ4 answer.} RRBM--DD benefits from interplay between both of its robustness enhancing components. They offer diverse, yet mutually complementary functionalities and their combination makes RRDM-DD robust to various types of adversarial attacks.
	
	\section{Conclusions and future works}
	\label{sec:con}
	
\noindent \textbf{Summary.} In this paper, we have presented a novel scenario of learning from data streams under adversarial concept drift. We proposed that two types of change may be present during the stream processing. Valid concept drift represents natural changes in the underlying data characteristics and must be correctly detected in order to allow a smooth adaptation of the model. Adversarial concept drift represents a poisoning attack from a malicious party that aims at hindering or disabling the classification system. We discussed the negative impacts of adversarial drift on classifiers and proposed a taxonomy of two types of attacks: instance-based and concept-based. We examined state-of-the-art concept drift detectors and concluded that none of them is capable of handling adversarial instances. To address this challenging task we proposed a novel approach that is robust to the presence of adversarial instances: Robust Restricted Boltzmann Machine Drift Detector. We presented how this fully trainable neural model can be used for efficient tracking of trends in the stream and accurate drift recognition. We enhanced our drift detector with the improved gradient calculation method and energy function with an embedded noise model. These modifications made our model robust to adversarial concept drift, while still being sensitive to valid drifts. Finally, we have introduced Relative Loss of Robustness, a novel measure for evaluating the performance of drift detectors and other streaming algorithms under adversarial concept drift. 

\smallskip
\noindent \textbf{Main conclusions.} The efficacy of RRBM--DD was evaluated on the basis of an extensive experimental study. The first two experiments compared RRBM--DD with five state-of-the-art drift detectors under instance-based and concept-based poisoning attacks. Obtained results clearly backed-up our observations that none of the existing drift detectors displays robustness to adversarial instances. RRBM--DD offered not only excellent detection rates of valid concept drifts, but also high robustness to all types and intensity levels of adversarial attacks. These characteristics were valid in both fully and sparsely labeled data streams, making RRBM-DD a highly suitable algorithm for a plethora of real-life applications. Finally, we have conducted an ablation study to understand the impact of RRBM-DD components on its robustness. Interestingly, both components were useful in specific tasks. The robust gradient algorithm was contributing most when dealing with instance-based poisoning attacks, while robust energy function offered significant gains for RRBM-DD when handling concept-based poisoning attacks.  

\smallskip
\noindent \textbf{Future works.} RRBM--DD opens several directions for future research. We plan to modify its cost functions in order to tackle the issue of simultaneous adversarial concept drift and dynamic class imbalance, as well as develop deep architectures based on RRBM--DD.
	
	\bibliographystyle{spbasic}       
	\bibliography{refs}
	
\end{document}